\documentclass[pdflatex,sn-mathphys-num]{sn-jnl}%

\usepackage{graphicx}%
\usepackage{multirow}%
\usepackage{amsmath,amssymb,amsfonts}%
\usepackage{amsthm}%
\usepackage{mathrsfs}%
\usepackage[title]{appendix}%
\usepackage{xcolor}%
\usepackage{textcomp}%
\usepackage{manyfoot}%
\usepackage{booktabs}%
\usepackage{algorithm}%
\usepackage{algorithmicx}%
\usepackage{algpseudocode}%
\usepackage{listings}%
\usepackage{todonotes}
\usepackage{tabularx} %
\usepackage{eso-pic} 
\usepackage[np, autolanguage]{numprint}

\usepackage[english, capitalise]{cleveref}
\usepackage[acronym,toc,shortcuts]{glossaries}

\newacronym{llm}{LLM}{Large Language Model}
\newacronym{ehr}{EHR}{Electronic Health Record}
\newacronym{loinc}{LOINC}{Logical Observation Identifiers Names and Codes}
\newacronym{omop}{OMOP}{Observational Medical Outcomes Partnership}
\newacronym{gbm}{GBM}{Gradient Boosted Machine}
\newacronym{auroc}{AUROC}{area under the receiver operating characteristic curve}
\newacronym{auprc}{AUPRC}{area under the precision-recall curve}
\newacronym{femr}{FEMR}{Framework for Electronic Medical Records}
\newacronym{ukb}{UKB}{UK Biobank}
\newacronym{lora}{LoRA}{Low-Rank Adaptation}
\newacronym{icl}{ICL}{in-context learning}
\newacronym{meme}{MEME}{Multiple Embedding Model for EHR}

\glsdisablehyper

\urldef\cpturl\url{https://gist.github.com/lieldulev/439793dc3c5a6613b661c33d71fdd185}
\urldef\icdurl\url{https://hcup-us.ahrq.gov/toolssoftware/procedureicd10/procedure_icd10_archive.jsp}
\urldef\cvxurl\url{https://www2a.cdc.gov/vaccines/iis/iisstandards/vaccines.asp?rpt=cvx}

\theoremstyle{thmstyleone}%

\theoremstyle{thmstyletwo}%

\theoremstyle{thmstylethree}%

\newcommand{\smalldash}{\scalebox{0.95}{-}}
\newcommand{\ci}[3]{#1\,_{#2\smalldash#3}}
\newcommand{\citext}[3]{\np{#1} (\np{#2}–\np{#3})}

\begin{document}

\title[Large Language Models are Powerful EHR Encoders]{Large Language Models are Powerful Electronic Health Record Encoders}

\author*[1,2,3]{\fnm{Stefan} \sur{Hegselmann}}\email{stefan.hegselmann@charite.de}
\equalcont{}

\author[1]{\fnm{Georg} \spfx{von} \sur{Arnim}} %
\equalcont{These authors contributed equally to this work.}

\author[1]{\fnm{Tillmann} \sur{Rheude}} %

\author[1]{\fnm{Noel} \sur{Kronenberg}} %

\author[4,5]{\fnm{David} \sur{Sontag}} %

\author[3]{\fnm{Gerhard} \sur{Hindricks}} %

\author[1,6]{\fnm{Roland} \sur{Eils}} %

\author[1]{\fnm{Benjamin} \sur{Wild}} %

\affil[1]{\orgname{Berlin Institute of Health at Charité -- Universitätsmedizin Berlin}, \orgdiv{Center of Digital Health}, \city{Berlin}, \country{Germany}}

\affil[2]{\orgname{Berlin Institute of Health at Charité -- Universitätsmedizin Berlin}, \orgdiv{BIH Biomedical Innovation Academy, BIH Charité Digital Clinician Scientist Program}, \city{Berlin}, \country{Germany}}

\affil[3]{\orgname{Deutsches Herzzentrum der Charité – Medical Heart Center of Charité and German Heart Institute Berlin}, \city{Berlin}, \country{Germany}}

\affil[4]{\orgdiv{Computer Science and Artificial Intelligence Laboratory (CSAIL)}, \orgname{Massachusetts Institute of Technology (MIT)}, \city{Cambridge}, \country{MA, USA}}

\affil[5]{\orgdiv{Layer Health, Inc.}, \country{MA, USA}}

\affil[6]{\orgdiv{Intelligent Medicine Institute}, \orgname{Fudan University}, \city{Shanghai}, \country{China}}

\abstract{\acfp{ehr} offer considerable potential for clinical prediction, but their complexity and heterogeneity challenge traditional machine learning.
Domain-specific \ac{ehr} foundation models trained on unlabeled \ac{ehr} data have shown improved predictive accuracy and generalization.
However, their development is constrained by limited data access and site-specific vocabularies.
We convert \ac{ehr} data into plain text by replacing medical codes with natural-language descriptions, enabling general-purpose \acfp{llm} to produce high-dimensional embeddings for downstream prediction tasks without access to private medical training data.
LLM-based embeddings perform on par with a specialized EHR foundation model, CLMBR-T-Base, across \np{15} clinical tasks from the EHRSHOT benchmark.
In an external validation using the UK Biobank, an LLM-based model shows statistically significant improvements for some tasks, which we attribute to higher vocabulary coverage and slightly better generalization.
Overall, we reveal a trade-off between the computational efficiency of specialized EHR models and the portability and data independence of LLM-based embeddings.}


\maketitle

\section{Introduction}
\label{sec:introduction}

\acfp{ehr} are now widely used in modern healthcare, providing comprehensive, longitudinal views of a patient's health status \cite{dash_big_2019}.
Machine learning methods can leverage this rich data for risk stratification and to support clinical decision-making \cite{ahsan_retrieving_2024, rajkomar_machine_2019, xu_quantitative_2021}.
In recent years, researchers have explored a variety of prediction tasks based on \acp{ehr}, including hospital readmission \cite{golas_machine_2018, rajkomar_scalable_2018}, length of hospital stay \cite{rajkomar_scalable_2018}, sepsis onset detection \cite{lauritsen_early_2020, moor_predicting_2023}, mortality prediction \cite{rajkomar_scalable_2018, thorsen-meyer_dynamic_2020}, discharge diagnoses \cite{rajkomar_scalable_2018}, and heart failure outcomes \cite{desai_comparison_2020}.
The overarching goal is to harness \ac{ehr} data using machine learning to improve clinical outcomes and reduce healthcare costs.

However, machine learning on \ac{ehr} data poses significant challenges due to its inherent complexity.
\ac{ehr} data is characterized by variable-length sequences of patient visits, irregular sampling intervals, missing entries, heterogeneous and noisy information, and a wide range of hierarchical medical concepts \cite{kim_evolving_2019}.
As a result, deep learning models often achieve only modest improvements over traditional methods such as logistic regression or tree-based models for \ac{ehr} prediction tasks \cite{rajkomar_scalable_2018, rasmy_med-bert_2021, steinberg_language_2021}.
To address these challenges, recent approaches have employed large-scale foundation models pretrained on unlabeled \ac{ehr} data using unsupervised learning \cite{bommasani_opportunities_2022}.
Many of these models adopt strategies from natural language processing, such as masked word prediction as in BERT \cite{devlin_bert_2019} or autoregressive next-word prediction as in GPT \cite{radford_language_2019}.
Treating \ac{ehr} data as sequences of medical codes enables analogous methods such as masked code prediction \cite{odgaard_core-behrt_2024, pang_cehr-bert_2021, rasmy_med-bert_2021, li_behrt_2020} or next-code prediction \cite{steinberg_language_2021, waxler_generative_2025, shmatko_learning_2025}.
However, code-based \ac{ehr} foundation models face two fundamental obstacles to interoperability and generalization: site-specific coding practices and fixed vocabularies learned during pretraining.
For example, CLMBR-T-Base \cite{steinberg_language_2021} supports only \np{26249} unique codes from its training corpus, and when applied to the \ac{ukb} with \np{50702} unique medical codes, only \np{7969} (\np{16}\%) could be mapped, leaving \np{84}\% of codes unseen by the model.
Achieving interoperability would require pretraining on diverse \ac{ehr} datasets from many institutions, which is difficult due to the sensitivity of healthcare data.
This motivates models that operate on natural-language descriptions of clinical codes, which avoid fixed vocabularies and transfer more readily across institutions.

\begin{figure}[t!]
    \centering
    \vspace{-0.25cm}
    \includegraphics[width=\linewidth]{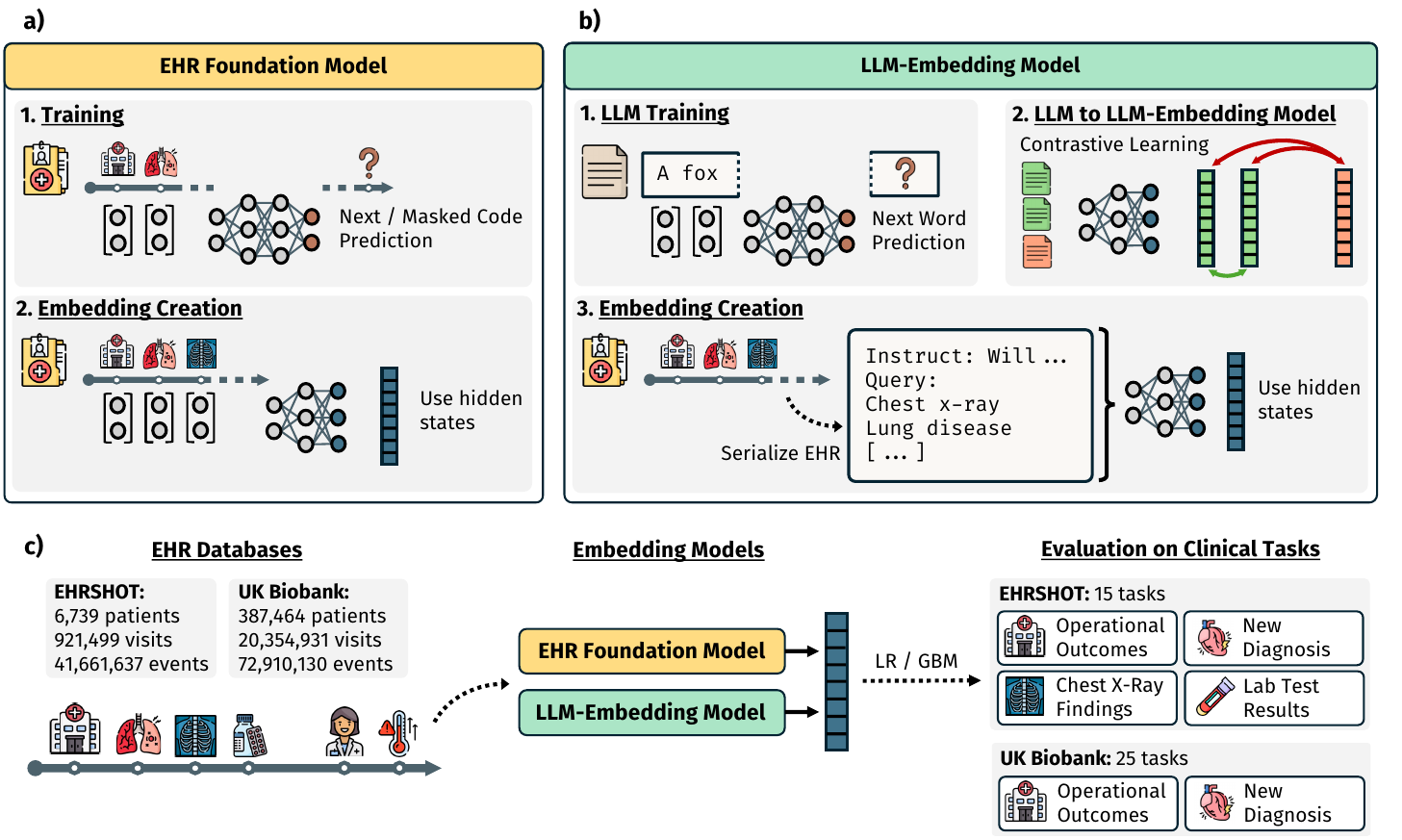}
    \vspace{-0.32cm}
    \caption{\textbf{Study Overview.} (a) \ac{ehr} foundation models are pretrained on unlabeled \ac{ehr} data. Common unsupervised learning tasks are masked-code or next-code prediction. To obtain a representation for an \ac{ehr}, we use the hidden states of the pretrained model. (b) \acp{llm} are pretrained on vast amounts of text data. To obtain an \ac{llm} embedding model, architectural changes are applied, and contrastive learning is used to improve representational performance. To obtain an \ac{ehr} embedding, the data is first serialized as text and then processed by the \ac{llm} embedding model. Again, we use the hidden states for the embedding. (c) We use the EHRSHOT benchmark and the \acf{ukb} cohort for our experiments. Medical events of each patient are converted into numerical embeddings using an \ac{ehr} foundation model and an \ac{llm} embedding model, respectively. A logistic regression (LR) model is trained, validated, and tested for each clinical prediction task. We also test a \acf{gbm} prediction model for the count-based baseline. Images from Flaticon.com.}
    \label{fig:overview}
\end{figure}

\acfp{llm} benefit from pretraining on vast general-purpose text corpora and a broad range of natural-language tasks \cite{penedo_refinedweb_2023, raffel_exploring_2020}.
This extensive pretraining enables strong language comprehension and allows them to capture domain-agnostic patterns that can be adapted for healthcare applications.
Consequently, \acp{llm} have demonstrated strong performance in extracting medical concepts \cite{agrawal_large_2022}, summarizing medical texts \cite{van_veen_adapted_2024}, and predicting medical outcomes \cite{hegselmann_tabllm_2023}, even in low-resource settings.
Recent work extends \acp{llm} to structured \ac{ehr} by serializing records into text and either using model generations for prediction \cite{shoham_cpllm_2024, zhu_prompting_2024, cui_llms-based_2024, acharya_clinical_2024, chen_clinicalbench_2024, makarov_large_2025} or extracting fixed-dimensional embeddings for downstream classifiers \cite{gao_when_2024, lee_clinical_2025, contreras_dellirium_2024, kirchler_large_2026}.
While both paradigms can be competitive with common baselines, many prior studies use short context windows and evaluate on private or emergency-department cohorts, limiting longitudinal coverage and external validity \cite{gao_when_2024, contreras_dellirium_2024, lee_clinical_2025} (\cref{subsec:methods/existing_methods}).
Moreover, most modern \acp{llm}, such as GPT \cite{brown_language_2020} and Qwen \cite{yang_qwen2_2024, yang_qwen3_2025}, use decoder-only transformer architectures trained with left-to-right objectives, which are not optimized for representation learning.
To address this limitation, recent work converts decoder-only \acp{llm} into effective \ac{llm} embedding models via contrastive learning or related techniques \cite{behnamghader_llm2vec_2024, lee_nv-embed_2024, muennighoff_generative_2024, li_towards_2023, zhang_qwen3_2025}.
Additionally, these state-of-the-art models offer large context windows, making them well suited for handling long inputs such as serialized \ac{ehr} data.

In this study, we present a systematic evaluation of modern general-purpose \ac{llm} embedding models as encoders of longitudinal \ac{ehr} data for clinical prediction \cite{wornow_ehrshot_2023} (\cref{fig:overview}).
To this end, we convert structured \ac{ehr} records into a list of plain-text descriptions of medical codes available at prediction time.
Using a state-of-the-art \ac{llm} embedding model, Qwen3-Embedding-8B (Qwen3-Emb-8B) \cite{yang_qwen3_2025, zhang_qwen3_2025} with a context size of \np{8192} tokens, we generate high-dimensional \ac{ehr} embeddings that serve as inputs to logistic regression classifiers across 15 clinical tasks from the EHRSHOT benchmark.
Rather than proposing a new modeling architecture, we assess the representation capabilities of these models under a standardized and reproducible evaluation protocol.
We intentionally use a simple embedding-plus-classifier pipeline to enable fair comparison with prior work and to isolate representation quality from downstream modeling choices.
We analyze performance in the few-shot setting to evaluate generalization, apply paired statistical tests to assess task-level differences between the \ac{llm} embedding model and competing methods, and conduct extensive ablation studies to identify the factors that drive its effectiveness.
Finally, we perform an external validation on the \ac{ukb} for predicting mortality, hospitalization, and the onset of \np{23} diseases \cite{steinfeldt_medical_2025} to assess generalization across datasets and coding systems.

\section{Results}
\label{sec:results}

\subsection{Experimental Setup}

Our primary analyses used the EHRSHOT benchmark, which contains \acp{ehr} from \np{6739} adult patients treated at Stanford Health Care and Lucile Packard Children’s Hospital between 1990 and 2023.
The dataset includes \np{921499} visits and more than \np{41.6} million clinical events, and defines a standardized evaluation across \np{15} clinical prediction tasks from \np{4} task categories, with predefined splits and public code \cite{wornow_ehrshot_2023}.
\Cref{tab:cohort_ehrshot_ukb_overview} summarizes cohort statistics, and task details are shown in \cref{tab:ehrshot_prediction_tasks_overview}.
Additional information on task definitions and preprocessing is provided in \cref{subsec:methods/ehrshot_database_and_prediction_tasks}.

\begin{table}[]
    \caption{\textbf{Cohort Overview.} Summary statistics for EHRSHOT and UK Biobank, including the number of patients, visits, events, and patient characteristics.}
    \label{tab:cohort_ehrshot_ukb_overview}
    \centering
    \footnotesize
    \vspace{-0.3cm}
    \setlength{\tabcolsep}{9.0pt} 
        \begin{tabular}{@{}p{3cm} c@{\hspace{15pt}}c@{}} \toprule
        \textbf{Attribute} & \textbf{EHRSHOT} & \textbf{UK Biobank} \\ \midrule
        \textbf{Num Patients} &  6,739      & 387,464 \\
        \textbf{Num Visits}   & 921,499    & 20,354,931 \\ %
        \textbf{Num Events}   & 41,661,637 & 72,910,130 \\
        \textbf{Num Female}   & 3,441      & 214,565 \\
        \textbf{Num Male}     & 3,298      & 172,899 \\
        \textbf{Age, mean $\pm$ SD}      & 59.3 $\pm$ 17.9        & 56.78 $\pm$ 8.11 \\
        \textbf{American Indian}    & 25         & 0 \\
        \textbf{Asian}              & 1,043      & 8,659 \\
        \textbf{Black}              & 298        & 5,715 \\
        \textbf{Pacific Islander}   & 74         & 0 \\
        \textbf{Unknown}            & 1,563      & 7,202 \\
        \textbf{White}              & 3,736      & 365,888 \\
        \textbf{Hispanic}           & 1,038      & - \\
        \textbf{Non-Hispanic}       & 5,701      & - \\  \bottomrule
    \end{tabular}
\end{table}

\begin{table}[]
    \caption{\textbf{EHRSHOT Prediction Tasks Overview.} The EHRSHOT benchmark defines 15 clinical prediction tasks spanning four task groups. The number of examples per task differs based on the prevalence and frequency of clinical events. Canonical splits for training, validation, and testing are defined to ensure reproducible experiments \cite{wornow_ehrshot_2023}.}
    \label{tab:ehrshot_prediction_tasks_overview}
    \centering
    \footnotesize
    \begin{tabular}{@{}p{2.85cm} p{2cm} p{2cm} p{2cm} p{2.15cm}@{}} \toprule  %
    \textbf{Attribute} & \textbf{\begin{tabular}[c]{@{}l@{}}Train Labels   \\ (Positive)\end{tabular}} & \textbf{\begin{tabular}[c]{@{}l@{}}Valid Labels   \\ (Positive)\end{tabular}} & \textbf{\begin{tabular}[c]{@{}l@{}}Test Labels   \\ (Positive)\end{tabular}} & \textbf{\begin{tabular}[c]{@{}l@{}}Total Labels   \\ (Positive)\end{tabular}} \\ \midrule
    \multicolumn{5}{l}{\textbf{Operational Outcomes}} \\ \midrule
    Long Length of Stay    & 2,569 (681)               & 2,231 (534)               & 2,195 (552)              & 6,995 (1,767)              \\
    30-day Readmission     & 2,609 (370)               & 2,207 (281)               & 2,189 (260)              & 7,005 (911)               \\
    ICU Transfer           & 2,402 (113)               & 2,052 (92)                & 2,037 (85)               & 6,491 (290)               \\ \midrule
    \multicolumn{5}{l}{\textbf{Anticipating Lab Test Results}} \\ \midrule
    Thrombocytopenia       & 68,776 (9,774)             & 54,504 (6,962)             & 56,338 (7,960)            & 179,618 (24,696)          \\
    Hyperkalemia           & 76,349 (1,215)             & 60,168 (886)              & 63,653 (948)             & 200,170 (3,049)            \\
    Hypoglycemia           & 122,108 (1,065)            & 95,488 (858)              & 100,568 (783)            & 318,164 (2,706)            \\
    Hyponatremia           & 81,336 (20,181)            & 64,473 (14,674)            & 67,028 (16,003)           & 212,837 (50,858)          \\
    Anemia                 & 70,501 (9,544)             & 56,224 (7,445)             & 58,155 (7,636)            & 184,880 (24,625)          \\ \midrule
    \multicolumn{5}{l}{\textbf{Assignment of New Diagnoses}}  \\ \midrule
    Hypertension           & 1,260 (184)               & 1,250 (177)               & 1,261 (160)              & 3,771 (521)               \\
    Hyperlipidemia         & 1,684 (205)               & 1,441 (189)               & 1,317 (172)              & 4,442 (566)               \\
    Pancreatic Cancer      & 2,576 (155)               & 2,215 (53)                & 2,220 (56)               & 7,011 (264)               \\
    Celiac                 & 2,623 (62)                & 2,284 (11)                & 2,222 (21)               & 7,129 (94)                \\
    Lupus                  & 2,570 (104)               & 2,226 (33)                & 2,243 (20)               & 7,039 (157)               \\
    Acute MI               & 2,534 (175)               & 2,177 (146)               & 2,127 (144)              & 6,838 (465)               \\ \midrule
    \multicolumn{5}{l}{\textbf{Anticipating Chest X-ray Findings}} \\ \midrule   
    Chest X-Ray   Findings & 7,481 (4,771)              & 9,366 (6,032)              & 9,428 (6,400)             & 26,275 (17,203)  \\      
    \bottomrule
    \end{tabular}
\end{table}

For external validation, we used the \ac{ukb}, a population-based cohort of \np{502489} UK participants \cite{sudlow_uk_2015, bycroft_uk_2018}.
This setting allowed us to assess generalization across healthcare systems, particularly because CLMBR-T-Base was trained on data from the same hospital system as EHRSHOT.
We followed the EHRSHOT setup as closely as possible and evaluated one-year risk of hospitalization, mortality, and onset of \np{23} diseases \cite{steinfeldt_medical_2025} (\cref{subsec:methods/external_validation_ukb}).
The processed \ac{ukb} subset used in our study comprised \np{387464} patients, approximately \np{20.4} million visits, and more than \np{72} million clinical events (\cref{tab:cohort_ehrshot_ukb_overview}).
\cref{tab:ukb_prediction_tasks_overview} reports tasks and label distributions for the \ac{ukb}.

\begin{figure}[]
    \centering
    \includegraphics[width=0.5\textwidth]{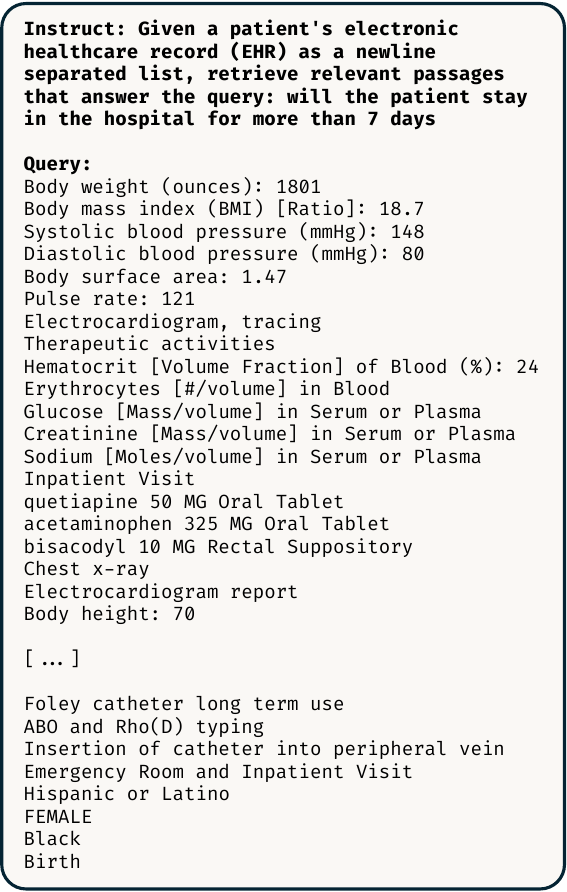}
    \caption{\textbf{Example EHR Text Serialization with Instruction.} The \ac{ehr} data are serialized into plain text to enable the use of LLM embedding models. The default serialization consists of a newline-separated list of the most recent occurrence of each medical code. No additional preprocessing or filtering is applied. Each code is represented by its text description, with an optional unit (given in brackets), and an optional value. Floating-point values are rounded to two decimal places.}
    \label{fig:example_ehr_text_serialization}
\end{figure}

To apply \acp{llm} to structured \ac{ehr} data, we serialized each patient record into plain text with a maximum context length of \np{8192} tokens.
Our default serialization was a simple newline-separated list of medical code descriptions, including units and values when available, with minimal preprocessing.
To remain within the token budget, we retained only the most recent occurrence of each medical code, which performed better than using the first occurrence or adding basic date and time information.
An example serialization is shown in \cref{fig:example_ehr_text_serialization}, and further details are provided in \cref{subsec:methods/ehr_text_serialization}.
For the \ac{ukb}, we used the same list-based serialization but omitted units and values because they were not available.

We evaluated three instruction-tuned \ac{llm} embedding models: Qwen3-Embedding-8B (Qwen3-Emb-8B) \cite{yang_qwen3_2025, zhang_qwen3_2025}, GTE-Qwen2-7B-Instruct (Qwen2-Emb-7B) \cite{yang_qwen2_2024, li_towards_2023}, and LLM2Vec-Llama-3.1-8B-Instruct (Llama3.1-LLM2Vec-8B) \cite{grattafiori_llama_2024, behnamghader_llm2vec_2024}.
We focused primarily on Qwen3-Emb-8B because of its recency and stronger long-context performance.
All embedding models received task-specific prompts (\cref{tab:instructions_for_llm_embedding_models}).
As an in-domain baseline, we included CLMBR-T-Base, a \np{141}-million-parameter autoregressive foundation model trained on \np{2.57} million de-identified \acp{ehr} from Stanford Medicine \cite{steinberg_language_2021, wornow_ehrshot_2023}.
The EHRSHOT validation and test splits were fully separated from CLMBR-T-Base pretraining (Fig. 1 in \cite{wornow_ehrshot_2023}).
We also extended the comparison to encoder-only biomedical language models using mean pooling and chunk-wise concatenation of \np{512}-token segments, and a simplified variant of the \ac{meme} method \cite{lee_clinical_2025}.

For each embedding model, we computed patient-level embeddings and trained a logistic regression classifier on the training split, with hyperparameters selected on the validation set.
Following the EHRSHOT protocol, we used the same logistic-regression head for all embedding-based representations to isolate embedding quality from downstream model complexity and reduce overfitting risk in few-shot settings.
As a baseline, we trained a \ac{gbm} on count-based representations of medical concepts.
The count baselines used ontology expansion from EHRSHOT \cite{wornow_ehrshot_2023} and were further extended with string values, numeric values, and time binning.

\subsection{General-Purpose LLM Embeddings Rival Domain-Specific EHR Models}

\begin{table}[t!]
    \caption{\textbf{Performance for All Examples on EHRSHOT.} Mean \acf{auroc} performance and approximate 95\% confidence intervals across tasks of selected models for four task groups. The macro-averaged performance across all task groups is given in the right-most column. The LLM embedding model Qwen3-Emb-8B with a context size of \np{8192} tokens and a logistic regression (LR) classification head performs on par with the EHR foundation model CLMBR-T-Base. Combining the embeddings of the LLM embedding model and CLMBR-T-Base by concatenation leads to an increase in performance. Additional model variants can be found in Table S4.
    }
    \label{tab:ehrshot_performance_on_all_examples}
    \centering
    \footnotesize
    \setlength{\tabcolsep}{2.6pt} 
    \begin{tabular}{>{\raggedright\arraybackslash}p{3.1cm} 
                >{\raggedright\arraybackslash}p{1.8cm} 
                >{\raggedright\arraybackslash}p{1.8cm} 
                >{\raggedright\arraybackslash}p{1.8cm} 
                >{\raggedright\arraybackslash}p{1.8cm} 
                >{\raggedright\arraybackslash}p{1.8cm}@{}}
    \toprule
\textbf{Model}                           & \textbf{Operational Outcomes} & \textbf{Anticipating Lab Test Results} & \textbf{Assignment of New Diagnosis} & \textbf{Anticipating Chest X-ray Findings} & \textbf{Macro Avg. Across Task Groups} \\ \midrule
\multicolumn{6}{l}{\textbf{Baselines} \cite{wornow_ehrshot_2023}} \\ \midrule
CLMBR-T-Base                             & $\ci{0.824}{.803}{.845}$ & $\ci{0.832}{.824}{.840}$ & $\ci{0.707}{.667}{.746}$ & $\ci{0.713}{.702}{.724}$ & $\ci{0.769}{.746}{.792}$ \\
Count-based + GBM                       & $\ci{0.824}{.804}{.844}$ & $\ci{0.841}{.833}{.849}$ & $\ci{0.758}{.724}{.793}$ & $\ci{0.686}{.674}{.699}$ & $\ci{0.777}{.756}{.799}$ \\ \midrule
\multicolumn{6}{l}{\textbf{LLM Embedding Models}} \\ \midrule
Qwen3-Emb-8B                             & $\ci{0.797}{.773}{.820}$ & $\ci{0.842}{.835}{.850}$ & $\ci{0.714}{.672}{.757}$ & $\ci{0.722}{.711}{.733}$ & $\ci{0.769}{.744}{.794}$ \\
Qwen3-Emb-4B                             & $\ci{0.787}{.764}{.810}$ & $\ci{0.824}{.816}{.831}$ & $\ci{0.718}{.667}{.768}$ & $\ci{0.708}{.696}{.719}$ & $\ci{0.759}{.730}{.787}$ \\
Qwen3-Emb-0.6B                           & $\ci{0.778}{.753}{.803}$ & $\ci{0.742}{.732}{.753}$ & $\ci{0.684}{.631}{.737}$ & $\ci{0.705}{.694}{.716}$ & $\ci{0.727}{.697}{.758}$ \\ \midrule
\multicolumn{6}{l}{\textbf{LLM Embedding Model + EHR Foundation Model} \cite{wornow_ehrshot_2023}} \\ \midrule
Qwen3-Emb-8B + CLMBR-T-Base              & $\ci{0.821}{.800}{.842}$ & $\ci{0.864}{.858}{.871}$ & $\ci{0.736}{.695}{.777}$ & $\ci{0.731}{.721}{.742}$ & $\ci{0.788}{.764}{.812}$ \\ \midrule
\multicolumn{6}{l}{\textbf{Multiple Embedding Model for EHR (MEME)} \cite{lee_clinical_2025} \textbf{with Linear Head} } \\ \midrule
Qwen3-Emb-8B MEME                        & $\ci{0.814}{.793}{.834}$ & $\ci{0.845}{.837}{.852}$ & $\ci{0.728}{.673}{.784}$ & $\ci{0.717}{.705}{.728}$ & $\ci{0.776}{.746}{.806}$ \\
BioClinicalBERT MEME                             & $\ci{0.756}{.733}{.778}$ & $\ci{0.699}{.686}{.713}$ & $\ci{0.704}{.651}{.758}$ & $\ci{0.648}{.635}{.661}$ & $\ci{0.702}{.671}{.732}$ \\ \midrule
\multicolumn{6}{l}{\textbf{Encoder Language Models with Chunked Inputs}} \\ \midrule
BioClinicalBERT                             & $\ci{0.738}{.712}{.763}$ & $\ci{0.698}{.685}{.711}$ & $\ci{0.707}{.668}{.746}$ & $\ci{0.679}{.666}{.691}$ & $\ci{0.705}{.680}{.730}$ \\
MedBERT                                  & $\ci{0.742}{.718}{.767}$ & $\ci{0.694}{.683}{.706}$ & $\ci{0.663}{.614}{.713}$ & $\ci{0.683}{.671}{.696}$ & $\ci{0.696}{.667}{.725}$ \\
\bottomrule
\end{tabular}
\end{table}

\begin{table}[]
    \caption{\textbf{Win--Tie--Loss Summary of Qwen3-Emb-8B Compared to Competing Models.} Entries report the number of tasks where Qwen3-Emb-8B significantly outperforms (W), ties with (T), or significantly underperforms (L) the comparator model in the 8-shot, 64-shot, and all-data settings. Higher wins or losses are in bold. Full statistical results are in Table S5, Table S15, and Table S17.}
    \label{tab:qwen_win_tie_loss_shots}
    \centering
    \footnotesize
    \setlength{\tabcolsep}{2.6pt} 
    \begin{tabular}{>{\raggedright\arraybackslash}p{3.4cm}
                >{\raggedright\arraybackslash}p{1.1cm}
                >{\raggedright\arraybackslash}p{1.1cm}
                >{\raggedright\arraybackslash}p{1.1cm}@{}}
    \toprule
    \textbf{Qwen3-Emb-8B+LR vs.} & \textbf{8-shot} & \textbf{64-shot} & \textbf{All} \\ 
    \midrule
    \multicolumn{4}{l}{\textbf{EHRSHOT}} \\ \midrule
    CLMBR+LR & \textbf{4}/10/1 & \textbf{3}/10/2 & \textbf{3}/10/2 \\
    BioClinicalBERT+LR & \textbf{6}/9/0 & \textbf{7}/8/0 & \textbf{8}/7/0 \\
    Count-based+GBM & \textbf{8}/7/0 & 1/12/\textbf{2} & 2/10/\textbf{3} \\
    \midrule
    \multicolumn{4}{l}{\textbf{UKB}} \\ \midrule
    CLMBR+LR & \textbf{2}/23/0 & \textbf{4}/16/0 & \textbf{6}/19/0 \\
    BioClinicalBERT+LR & \textbf{6}/19/0 & \textbf{2}/18/0 & \textbf{8}/17/0 \\
    Count-based+GBM & \textbf{2}/22/1 & \textbf{1}/19/0 & \textbf{8}/14/3 \\
    \midrule
    \multicolumn{4}{l}{\textbf{UKB; Qwen3-Emb-8B restricted to CLMBR codes}} \\ \midrule
    CLMBR+LR & \textbf{5}/18/2 & \textbf{3}/17/0 & \textbf{2}/23/0 \\
   \bottomrule
    \end{tabular}
\end{table}

Using all available training and validation examples, Qwen3-Emb-8B matched the in-domain \ac{ehr} foundation model CLMBR-T-Base on EHRSHOT, with an overall macro-\ac{auroc} of \citext{0.769}{0.744}{0.794} versus \citext{0.769}{0.746}{0.792} (\cref{tab:ehrshot_performance_on_all_examples}).
Qwen3-Emb-8B performed slightly better in three of four task categories, namely lab prediction, assignment of new diagnoses, and chest X-ray prediction.
Task-level statistical testing showed that Qwen3-Emb-8B significantly outperformed CLMBR-T-Base on thrombocytopenia, hyponatremia, and hyperkalemia, whereas CLMBR-T-Base performed significantly better on anemia and hypoglycemia (\cref{tab:qwen_win_tie_loss_shots} and \cref{tab:ehrshot_significance_qwen}).
No significant differences were observed for the remaining tasks, indicating that most task-level differences were small within statistical uncertainty.
Concatenating Qwen3-Emb-8B and CLMBR-T-Base embeddings improved performance to \citext{0.788}{0.764}{0.812}, suggesting that the two models capture complementary information (\cref{tab:ehrshot_performance_on_all_examples}).
Smaller Qwen3 variants performed worse, with \citext{0.759}{0.730}{0.787} for Qwen3-Emb-4B and \citext{0.727}{0.697}{0.758} for Qwen3-Emb-0.6B.

The \ac{gbm}-based count baseline with ontology expansion, string and numeric values, and time binning substantially improved over the original EHRSHOT count baseline that used ontology expansion alone \cite{wornow_ehrshot_2023} (\cref{tab:ehrshot_count_models}).
Using all data, the count model achieved \citext{0.777}{0.756}{0.799}, slightly exceeding both Qwen3-Emb-8B and CLMBR-T-Base (\cref{tab:ehrshot_performance_on_all_examples}).
It significantly outperformed Qwen3-Emb-8B on thrombocytopenia, hyponatremia, and anemia, whereas Qwen3-Emb-8B only outperformed it on hypoglycemia and chest X-ray prediction (\cref{tab:qwen_win_tie_loss_shots} and \cref{tab:ehrshot_significance_qwen}).
These results underscore the strength of count-based models when many labeled examples are available and the marginal gains of pretrained representations.
We evaluated \np{14} encoder-based configurations, of which BioClinicalBERT with mean embeddings over 512-token chunks performed best at \citext{0.705}{0.680}{0.730} (\cref{tab:ehrshot_performance_on_all_examples} and \cref{tab:ehrshot_performance_on_all_examples_full}).
Qwen3-Emb-8B significantly outperformed BioClinicalBERT on eight of \np{15} tasks, indicating a clear advantage over encoder-only baselines under long-context list-based serialization (\cref{tab:qwen_win_tie_loss_shots}).
Using \ac{meme} with a logistic-regression head yielded slight gains for Qwen3-Emb-8B but no improvement for BioClinicalBERT (\cref{tab:ehrshot_performance_on_all_examples}).

\begin{figure}[t!]
    \centering
    \includegraphics[width=.85\textwidth]{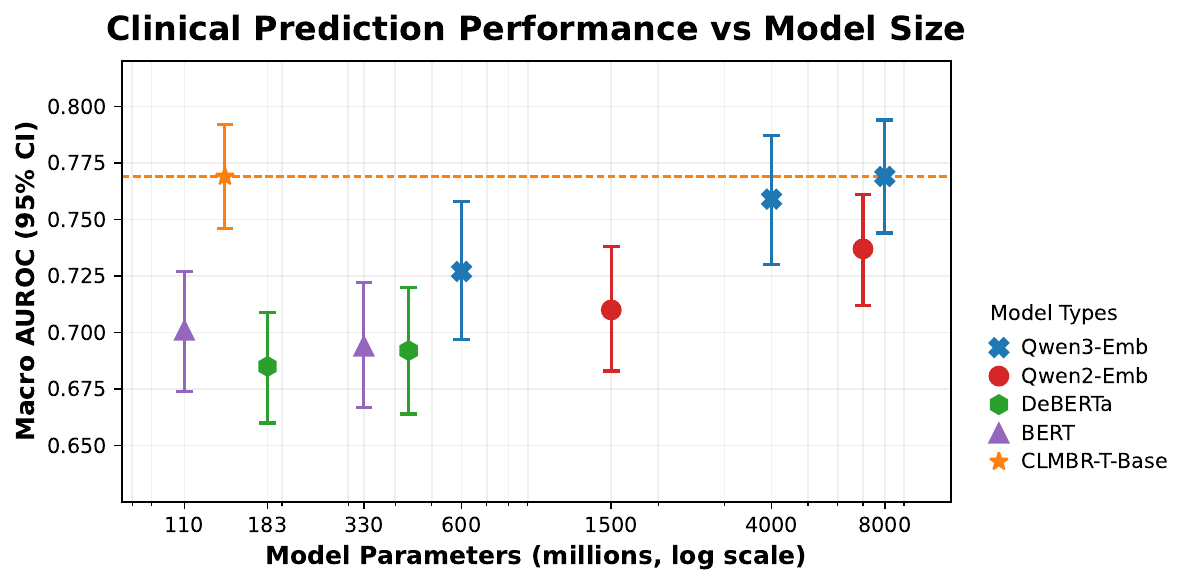}
    \caption{\textbf{Scaling Behavior on EHRSHOT.} Number of model parameters (x-axis) and macro-averaged \acf{auroc} performance with approximate 95\% confidence intervals across all four task groups (y-axis). We include only models with varying sizes. The performance results of Qwen3- and Qwen2-based LLM embedding models suggest scaling behavior with model size. Encoder-only models based on the BERT architecture do not show this trend. The specialized EHR foundation model, CLMBR-T-Base, is the most parameter-efficient model. Full results in \cref{tab:ehrshot_performance_on_all_examples_full}.}
    \label{fig:scaling_behavior_of_models}
\end{figure}

Scaling analyses showed only modest gains with increasing size for Qwen-based and DeBERTa models, whereas larger BERT variants performed worse (\cref{fig:scaling_behavior_of_models}).
CLMBR-T-Base remained the most parameter-efficient model relative to predictive performance.
Runtime analysis on EHRSHOT further emphasized this efficiency difference  (\cref{tab:ehrshot_encoding_time}).
CLMBR-T-Base required 6:04 minutes to encode all examples of the EHRSHOT benchmark, whereas Qwen3-Emb-8B required 21:48:56 hours and encoder models required 2:21:35 to 7:00:21 hours because of chunking (\cref{subsec:methods/llm_embedding_models_and_baselines}).
Thus, LLM-based methods achieved similar predictive performance at the cost of a substantially larger memory footprint and higher computational cost.

\begin{table}[ht]
    \caption{\textbf{Performance for All Examples on UKB.} Mean \acf{auroc} performance and approximate 95\% confidence intervals across tasks for three task groups.
    The LLM embedding model Qwen3-Emb-8B with a logistic regression (LR) classification head outperforms the EHR foundation model CLMBR-T-Base and the count-based baseline using a \acf{gbm} head. The assignment of new diagnoses prediction is based on the mean across all 23 provided diseases. Additional model variants can be found in Table S14.}
    \label{tab:ukb_performance_on_all_examples}
    \centering
    \footnotesize
    \setlength{\tabcolsep}{2.6pt} 
    \begin{tabular}{>{\raggedright\arraybackslash}p{3.2cm} 
                >{\raggedright\arraybackslash}p{2.2cm} 
                >{\raggedright\arraybackslash}p{2.4cm} 
                >{\raggedright\arraybackslash}p{2.2cm} 
                >{\raggedright\arraybackslash}p{2.2cm} @{}}
    \toprule
\textbf{Model}                           & \textbf{Mortality prediction} & \textbf{Operational Outcomes (Hospitalization)} & \textbf{Assignment of New Diagnoses}  & \textbf{Macro Avg. Across Task Groups}  \\ \midrule
\multicolumn{5}{l}{\textbf{Baselines} \cite{wornow_ehrshot_2023}} \\ \midrule
CLMBR-T-Base                & $\ci{0.801}{.772}{.830}$  & $\ci{0.689}{.685}{.693}$ & $\ci{0.719}{.708}{.731}$  & $\ci{0.736}{.726}{.747}$ \\
Count-based + GBM           & $\ci{0.780}{.749}{.810}$  & $\ci{0.709}{.705}{.712}$ & $\ci{0.634}{.622}{.646}$  & $\ci{0.708}{.697}{.719}$ \\ \midrule
\multicolumn{5}{l}{\textbf{LLM Embedding Model}} \\ \midrule
Qwen3-Emb-8B                & $\ci{0.811}{.781}{.840}$  & $\ci{0.698}{.694}{.702}$ & $\ci{0.743}{.731}{.755}$  & $\ci{0.751}{.740}{.761}$\\ \midrule
\multicolumn{5}{l}{\textbf{Sensitivity Analysis Restricted to CLMBR Codes}} \\ \midrule
Qwen3-Emb-8B CLMBR    & $\ci{0.806}{.775}{.836}$  & $\ci{0.687}{.683}{.691}$ & $\ci{0.736}{.723}{.749}$  & $\ci{0.743}{.732}{.754}$ \\ \midrule
\multicolumn{5}{l}{\textbf{Encoder Language Models with Chunked Inputs}} \\ \midrule
BioClinicalBERT               & $\ci{0.778}{.746}{.811}$  & $\ci{0.675}{.671}{.678}$ & $\ci{0.705}{.693}{.717}$  & $\ci{0.719}{.708}{.731}$ \\
\bottomrule
\end{tabular}
\end{table}

In the \ac{ukb}, Qwen3-Emb-8B achieved slightly higher overall performance than CLMBR-T-Base, with \citext{0.751}{0.740}{0.761} compared to \citext{0.736}{0.726}{0.747} (\cref{tab:ukb_performance_on_all_examples}).
Statistical testing showed significant improvements on six of \np{25} tasks and no significant differences otherwise (\cref{tab:qwen_win_tie_loss_shots}).
Because only \np{16}\% of UKB codes mapped to the CLMBR-T-Base vocabulary, we performed a sensitivity analysis restricting Qwen3-Emb-8B to the same codes (\cref{tab:ukb_performance_on_all_examples} and \cref{fig:sensitivity_analysis_auroc_ukb}).
Under this restriction, performance decreased to \citext{0.743}{0.732}{0.754}, placing it between full Qwen3-Emb-8B and CLMBR-T-Base.
Qwen3-Emb-8B restricted to CLMBR-T-Base-mappable codes significantly outperformed CLMBR-T-Base on two instead of six tasks (\cref{tab:qwen_win_tie_loss_shots}).
This indicates that the gains on \ac{ukb} can be explained by both broader vocabulary coverage and slightly improved generalization.
Qwen3-Emb-8B also significantly outperformed BioClinicalBERT on eight \ac{ukb} tasks (\cref{tab:qwen_win_tie_loss_shots}).
The count-based model performed slightly worse overall in the \ac{ukb}, likely reflecting the difficulty of \ac{gbm} learning on highly imbalanced tasks (\cref{tab:ukb_prediction_tasks_overview}).

\subsection{LLM Embeddings Achieve Strong Performance in Low-Data Regimes}
\label{res:performance_fewshot}

\begin{figure}[t!]
    \centering
    \includegraphics[width=\linewidth]{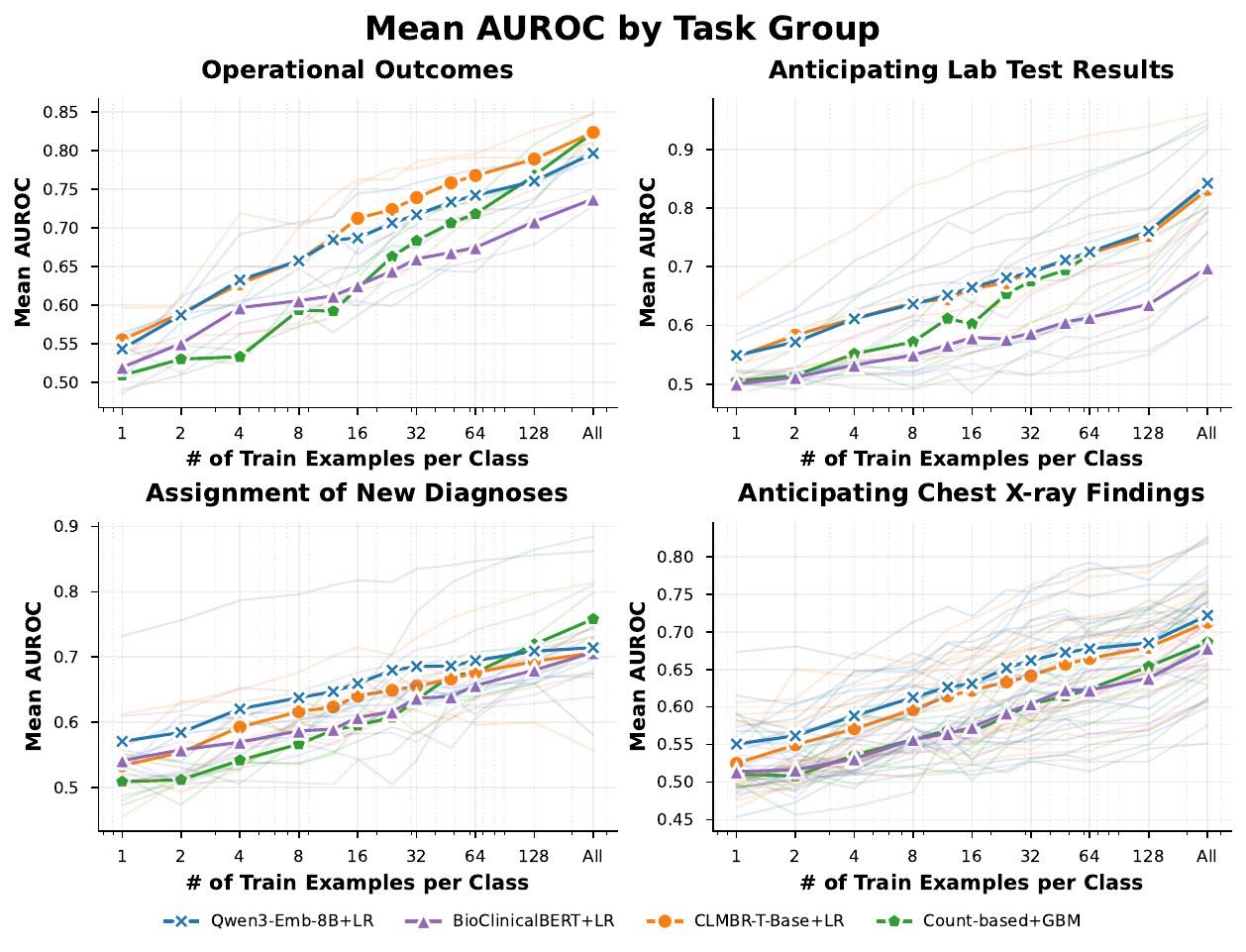}
    \vspace{-0.5cm}
    \caption{\textbf{Few-Shot Performance on EHRSHOT.} Mean \acf{auroc} performance across subtasks for four task groups (bold). Blurred lines show averaged \ac{auroc} values across five bootstrapped runs using different seeds \cite{wornow_ehrshot_2023}. The LLM embedding model performs similarly to the \ac{ehr} foundation model, CLMBR-T-Base, and shows the largest performance gains over the count-based model at intermediate numbers of training examples. The LLM embedding model consistently outperforms the biomedical embedding model BioClinicalBERT.}
    \label{fig:ehrshot_performance_in_few_shot_setting}
\end{figure}

We evaluated performance under limited supervision using the few-shot protocol from EHRSHOT \cite{wornow_ehrshot_2023}, including statistical comparisons for 8-shot and 64-shot settings.
Across EHRSHOT task groups, Qwen3-Emb-8B maintained slightly higher performance than CLMBR-T-Base for new diagnoses and chest X-ray tasks across most shot settings (\cref{fig:ehrshot_performance_in_few_shot_setting}).
CLMBR-T-Base showed higher performance only for operational outcomes, beginning at 16 shots.
Statistical testing confirmed that Qwen3-Emb-8B significantly outperformed CLMBR-T-Base on four versus one tasks in the 8-shot setting and on three versus two tasks in the 64-shot setting, while most tasks remained indistinguishable (\cref{tab:qwen_win_tie_loss_shots}).
The count baseline performed worse in the smallest-shot settings, but often matched or exceeded Qwen3-Emb-8B and CLMBR-T-Base by 64 or 128 shots, consistent with its strength in higher-data regimes (\cref{fig:ehrshot_performance_in_few_shot_setting}).
These findings indicate that the advantage of LLM embeddings is most apparent with limited labeled data.
Qwen3-Emb-8B also consistently outperformed BioClinicalBERT, which was likewise reflected in the task-level statistical testing (\cref{tab:qwen_win_tie_loss_shots}).
Additional metrics are reported in the supplement, including \ac{auprc} and Brier score (\cref{fig:auprc_performance_in_few_shot_settings}, \cref{fig:brier_performance_in_few_shot_settings}) and task-level \ac{auroc}, \ac{auprc}, and Brier scores (\cref{fig:task_specific_auroc_performance}, \cref{fig:task_specific_auprc_performance}, and \cref{fig:task_specific_brier_performance}).

\begin{figure}[t!]
    \centering
    \includegraphics[width=\linewidth]{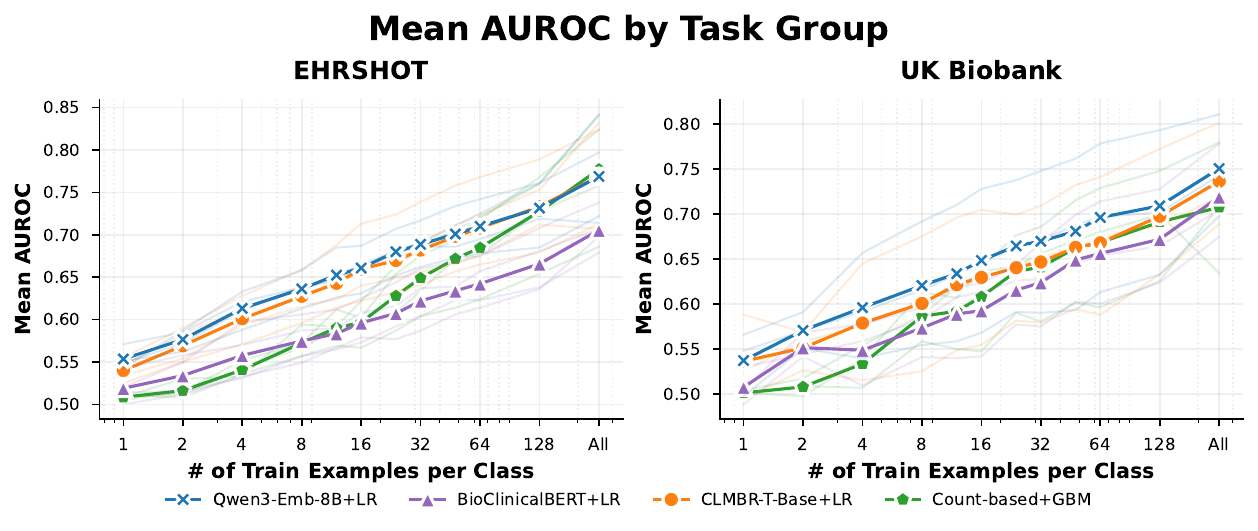}
    \vspace{-0.5cm}
    \caption{\textbf{Few-Shot Performance on EHRSHOT and UKB.} Macro-averaged \acf{auroc} performance across all subtasks of EHRSHOT (left) and UK Biobank (right). Blurred lines show averaged \ac{auroc} values for the different task groups. On EHRSHOT, the LLM embedding model performs on par with the \ac{ehr} foundation model, CLMBR-T-Base. On the UK Biobank, the LLM embedding model slightly outperforms the \ac{ehr} foundation model across all shot sizes.} 
    \label{fig:performance_EHRShot_UKB_AUROC}
\end{figure}

In the \ac{ukb}, Qwen3-Emb-8B again showed slightly higher average performance than CLMBR-T-Base across shot sizes (\cref{fig:performance_EHRShot_UKB_AUROC}).
Statistical analysis showed significant improvements over CLMBR-T-Base on two tasks in the 8-shot setting and four tasks in the 64-shot setting, while most tasks showed no significant difference (\cref{tab:qwen_win_tie_loss_shots}).
Restricting Qwen3-Emb-8B to CLMBR-T-Base-mappable codes again yielded intermediate performance between full Qwen3-Emb-8B and CLMBR-T-Base (\cref{tab:qwen_win_tie_loss_shots} and \cref{fig:sensitivity_analysis_auroc_ukb}), supporting a mixed contribution of vocabulary coverage and representation quality.
Qwen3-Emb-8B also outperformed the count baseline in few-shot UKB settings, although by smaller margins than in EHRSHOT.
Only two significant improvements over the count model were observed at 8 shots and one at 64 shots, indicating that most few-shot performance differences were modest in this cohort (\cref{tab:qwen_win_tie_loss_shots}).
Qwen3-Emb-8B also remained consistently stronger than BioClinicalBERT across few-shot settings.

\begin{table}[ht]
    \caption{\textbf{Performance for All Examples on EHRSHOT for Different Serializations.} Mean \acf{auroc} performance and approximate 95\% confidence intervals for the list serialization used in this work and three alternatives using the first occurrence of each code and adding timestamps to each code. We also tested a handcrafted Markdown EHR serialization and JSON, XML, and YAML data formats derived from it, showing slightly lower performance than the simple list serialization. Additional serialization results can be found in Table S7.}
    \label{tab:ehrshot_performance_formats}
    \centering
    \footnotesize
    \setlength{\tabcolsep}{2.6pt} 
    \begin{tabular}{>{\raggedright\arraybackslash}p{3.05cm} 
                >{\raggedright\arraybackslash}p{1.8cm} 
                >{\raggedright\arraybackslash}p{1.8cm} 
                >{\raggedright\arraybackslash}p{1.8cm} 
                >{\raggedright\arraybackslash}p{1.8cm} 
                >{\raggedright\arraybackslash}p{1.8cm}@{}}
    \toprule
\textbf{Model}                           & \textbf{Operational Outcomes} & \textbf{Anticipating Lab Test Results} & \textbf{Assignment of New Diagnosis} & \textbf{Anticipating Chest X-ray Findings} & \textbf{Macro Avg. Across Task Groups} \\ \midrule
\multicolumn{6}{l}{\textbf{EHR List Serializations for Qwen3-Emb-8B}} \\ \midrule
List codes recent (ours)                              & $\ci{0.797}{.773}{.820}$ & $\ci{0.842}{.835}{.850}$ & $\ci{0.714}{.672}{.757}$ & $\ci{0.722}{.711}{.733}$ & $\ci{0.769}{.744}{.794}$ \\
List codes first                          & $\ci{0.761}{.736}{.785}$ & $\ci{0.715}{.701}{.728}$ & $\ci{0.731}{.688}{.773}$ & $\ci{0.676}{.663}{.690}$ & $\ci{0.721}{.694}{.747}$ \\
List codes recent + time                       & $\ci{0.795}{.772}{.818}$ & $\ci{0.844}{.837}{.851}$ & $\ci{0.692}{.637}{.748}$ & $\ci{0.727}{.716}{.738}$ & $\ci{0.765}{.734}{.795}$ \\
List codes first + time                   & $\ci{0.746}{.720}{.772}$ & $\ci{0.703}{.689}{.716}$ & $\ci{0.718}{.674}{.761}$ & $\ci{0.645}{.629}{.660}$ & $\ci{0.703}{.675}{.730}$ \\  \midrule
\multicolumn{6}{l}{\textbf{EHR Alternative Serialization Formats for Qwen3-Emb-8B}} \\ \midrule
Markdown                                    & $\ci{0.773}{.749}{.797}$ & $\ci{0.859}{.852}{.866}$ & $\ci{0.725}{.683}{.767}$ & $\ci{0.694}{.681}{.707}$ & $\ci{0.763}{.737}{.788}$ \\
JSON                                        & $\ci{0.773}{.749}{.796}$ & $\ci{0.858}{.851}{.865}$ & $\ci{0.736}{.692}{.780}$ & $\ci{0.690}{.677}{.704}$ & $\ci{0.764}{.738}{.790}$ \\
XML                                         & $\ci{0.771}{.747}{.795}$ & $\ci{0.862}{.855}{.868}$ & $\ci{0.726}{.681}{.771}$ & $\ci{0.676}{.663}{.690}$ & $\ci{0.759}{.732}{.785}$ \\
YAML                                        & $\ci{0.773}{.749}{.796}$ & $\ci{0.863}{.856}{.870}$ & $\ci{0.723}{.677}{.769}$ & $\ci{0.684}{.670}{.698}$ & $\ci{0.761}{.734}{.787}$ \\ 
\bottomrule
\end{tabular}
\end{table}

\subsection{Effect of the EHR Serialization on LLM-Based Embedding Performance}

To assess the effect of the EHR text serialization, we compared our default list format, which retains the most recent occurrence of each medical code, with variants using the first occurrence or adding date and time information (\cref{tab:ehrshot_performance_formats}).
Using first rather than most recent occurrences substantially reduced performance, especially for lab-test prediction and chest X-ray findings, indicating that these tasks depend strongly on recent information.
Adding date and time information did not improve performance and instead caused a small decrease for recent-code serializations and a larger decrease for first-code serializations.
This suggests that current \ac{llm} embedding models do not effectively exploit explicit temporal markers in this simple long-context list format.
We also evaluated a handcrafted Markdown serialization designed to emphasize clinically relevant structure (\cref{subsec:methods/ehr_alternative_text_serialization}).
Overall, Markdown did not improve performance for Qwen3-Emb-8B relative to the simpler list-based serialization (\cref{tab:ehrshot_performance_formats}).
For lab-test prediction, however, Markdown with preprocessed laboratory information yielded small AUROC gains of 0.006 to 0.021 for Qwen3-Emb-8B, Qwen2-Emb-7B, and Llama3.1-LLM2Vec-8B (\cref{tab:ehrshot_performance_formats_full}).
These gains are consistent with stronger text cues in the Markdown lab-value representation.
Notably, Qwen3-Emb-8B remained clearly stronger than Qwen2-Emb-7B and Llama3.1-LLM2Vec-8B on the default list serialization (\cref{tab:ehrshot_performance_on_all_examples_full}), suggesting that it effectively extracts relevant signals from the raw data without relying strongly on preprocessing.
Replacing Markdown with JSON, XML, or YAML led to only minor differences (\cref{tab:ehrshot_performance_formats}).
XML showed the largest decrease, likely because of its greater formatting overhead.

\begin{figure}[t!]
    \centering
    \includegraphics[width=\linewidth]{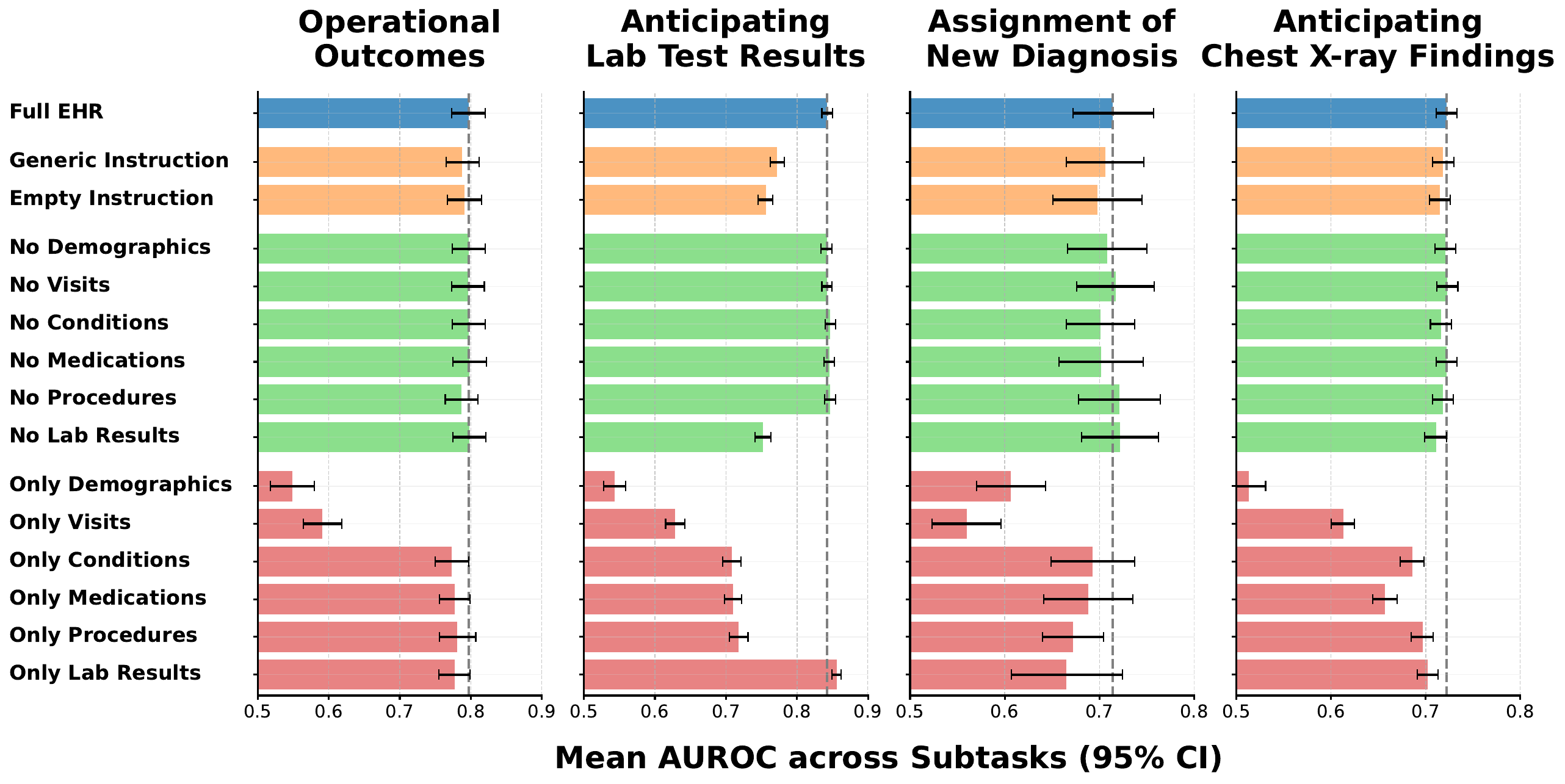}
    \vspace{-0.3cm}
    \caption{\textbf{Effects of EHR Serialization Content on EHRSHOT.} Mean \acf{auroc} performance with approximate 95\% confidence intervals for Qwen3-Emb-8B. The default list serialization (Full EHR) appears at the top, followed by runs with a generic and an empty instruction (orange). We then evaluate the serialization by removing specific code categories (green) and by retaining only individual categories (red). Category definitions are given in \cref{tab:code_categories_ablation}, and full results are reported in \cref{tab:ehrshot_ehr_serialization_ablation_experiments_for_llm_embeddings_models}.}
\label{fig:ehrshot_ehr_serialization_ablation_experiments_for_llm_embeddings_models}
\end{figure}

We further quantified the contribution of individual serialization components using ablations with Qwen3-Emb-8B (\cref{fig:ehrshot_ehr_serialization_ablation_experiments_for_llm_embeddings_models}).
Replacing task-specific instructions with a generic prompt slightly reduced performance, and removing instructions entirely caused a further decrease.
The largest drop occurred for lab-test prediction, suggesting that task-aligned instructions help the model focus on relevant spans.
Removing individual code categories for demographics, visits, conditions, medications, procedures, and labs had limited impact overall except for lab results, which were critical for lab prediction tasks (\cref{fig:ehrshot_ehr_serialization_ablation_experiments_for_llm_embeddings_models}).
This pattern indicates substantial redundancy across \ac{ehr} information sources.
Using only a single modality showed that demographics and visits alone, accounting for up to 1.5\% of all recorded events, were weak predictors.
Using only conditions, medications, procedures, or lab results yielded similar overall performance across task groups, with lab results being especially informative for lab-test prediction.
However, no single modality matched the full-\ac{ehr} representation, indicating that the modalities provide complementary information.

\subsection{Effect of Context Length and Temporal Scope on Embedding Effectiveness}

\begin{figure}[t!]
    \centering
    \includegraphics[width=\linewidth]{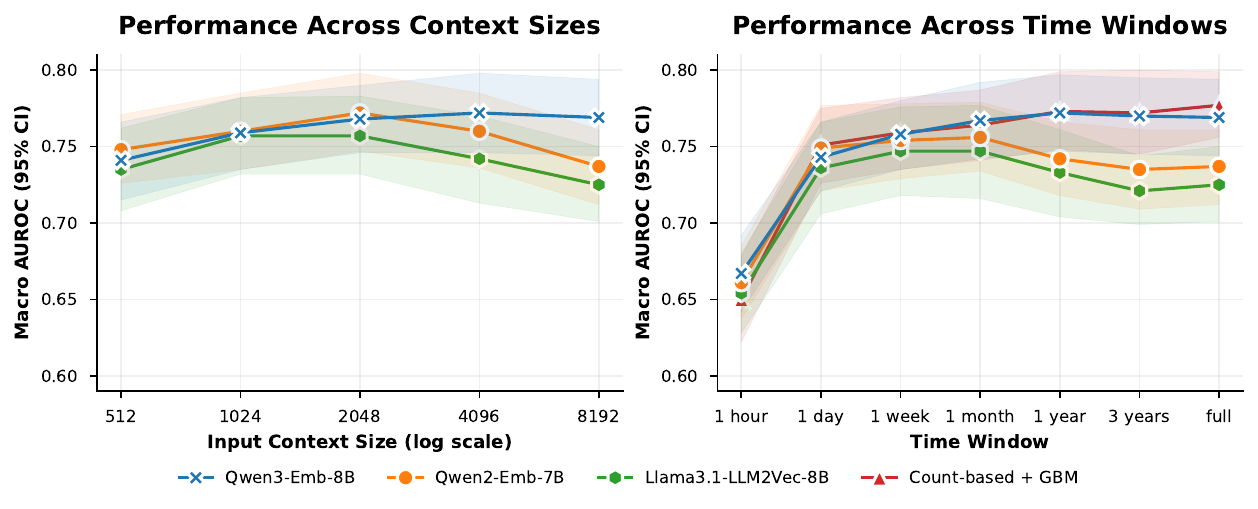}
    \caption{\textbf{Performance Across Context Size and Time Windows on EHRSHOT.} Macro-averaged \acf{auroc} performance with approximate 95\% confidence intervals (y-axis) across all task groups for different LLM embedding models and the count-based baseline, shown for different context sizes (left) and different time windows before prediction time (right). The LLM embedding models for the time-window experiments use a context size of \np{8192} tokens. All results are given in \cref{tab:ehrshot_performance_of_llm_embedding_models_across_context_sizes} and \cref{tab:ehrshot_performance_of_llm_embedding_models_across_time_windows}.}
\label{fig:performance_of_llm_embedding_models_across_time_windows_and_context_sizes}
\end{figure}

To assess sensitivity to input length and recency, we performed ablation studies on context size and temporal window using EHRSHOT (\cref{fig:performance_of_llm_embedding_models_across_time_windows_and_context_sizes}).
The models differed markedly in their optimal context sizes, with Qwen3-Emb-8B performing best at \np{4096} tokens, Qwen2-Emb-7B at \np{2048} tokens, and Llama3.1-LLM2Vec-8B at \np{1024} to \np{2048} tokens.
For Qwen2-Emb-7B and Llama3.1-LLM2Vec-8B, performance dropped substantially at \np{8192} tokens.
These results show that only Qwen3-Emb-8B handled unstructured long-context \ac{ehr} input robustly for the simple list serialization, indicating that model scale alone is insufficient for strong long-context performance.
Temporal-window analyses showed a similar pattern, in part because larger windows also increase effective input length.
Qwen3-Emb-8B performed best with a one-year window, whereas Qwen2-Emb-7B and Llama3.1-LLM2Vec-8B performed best with shorter windows of one month and one week, respectively (\cref{fig:performance_of_llm_embedding_models_across_time_windows_and_context_sizes}).
Again, Qwen3-Emb-8B was the only \ac{llm} embedding model without a pronounced performance drop at longer time horizons.
In contrast, the count-based baseline improved with larger time windows, further highlighting its robustness when many training examples are available.
A one-hour window caused a large performance drop across all models, indicating insufficient information for prediction in EHRSHOT.

\subsection{Fine-Tuned LLM Embedding and LLM Decoder Models}
\label{res:qwen3_finetune}

\begin{figure}[t!]
    \centering
    \includegraphics[width=0.62\textwidth]{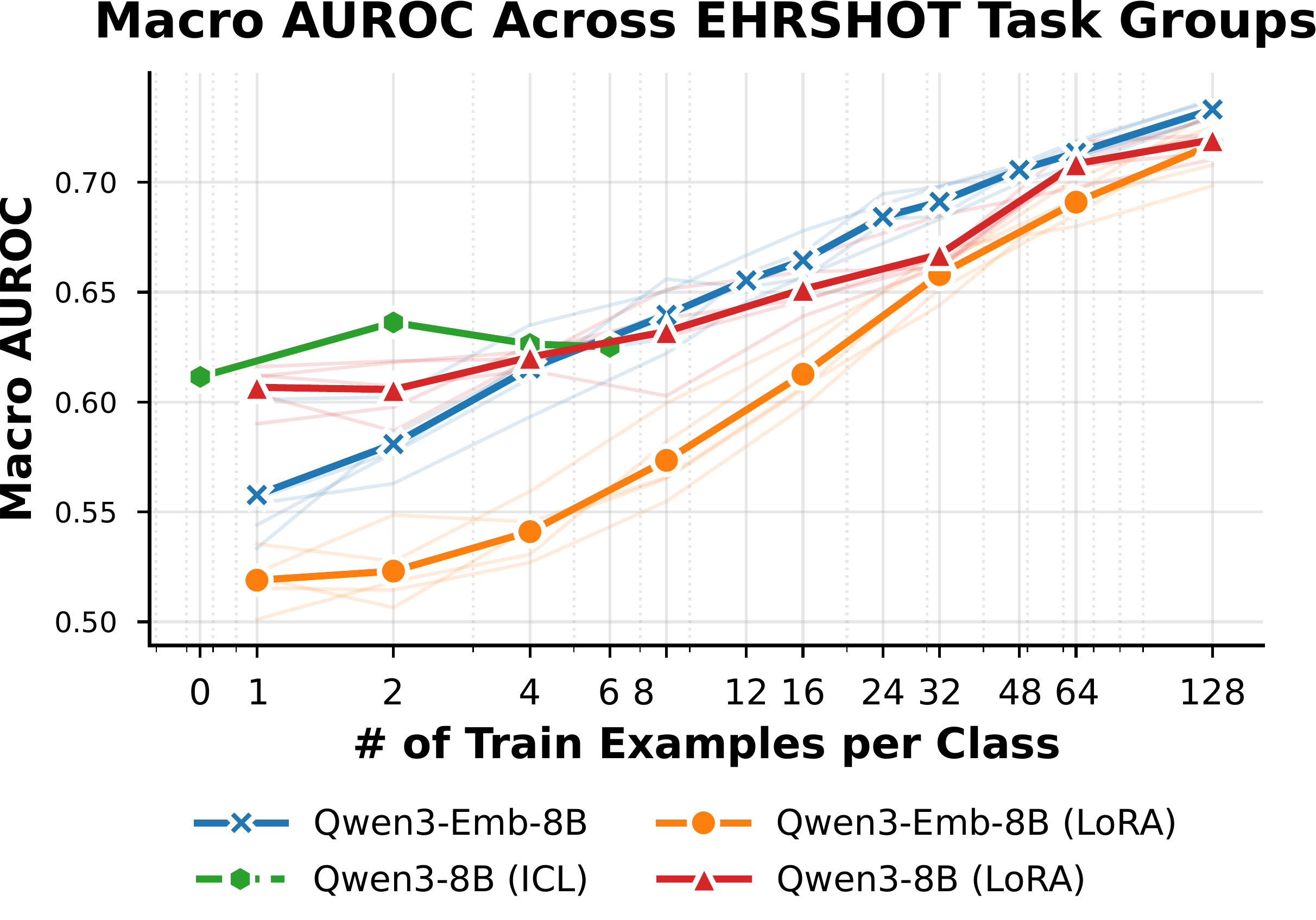}  
    \caption{\textbf{Few-Shot AUROC Performance of Encoder and Decoder Models on EHRSHOT.} Macro-averaged \acf{auroc} across all EHRSHOT subtasks for zero to \np{128} training examples per class. The frozen Qwen3-Emb-8B encoder baseline (blue) is compared with its \ac{lora}-tuned variant (orange), decoder \ac{icl} runs at \np{0}, \np{2}, \np{4}, and \np{6} shots (green), and the \ac{lora}-tuned Qwen3-8B decoder (red). Decoder \ac{icl} beyond \np{6} shots was not computationally feasible.}
    \label{fig:qwen3_encoder_decoder_overall}
\end{figure}

We evaluated decoder-style \acp{llm} that generate text, where token probabilities can be used for classification.
We extended the prompt with an additional instruction to produce \texttt{Yes} and \texttt{No} tokens leveraged for prediction (\cref{tab:instructions_for_llm_embedding_models}).
In addition, we fine-tuned both the \ac{llm} embedding model and the LLM decoder model using \ac{lora}, and evaluated decoder in-context learning with \np{2}, \np{4}, and \np{6} examples.
The decoder experiments train the instruction-tuned causal language model Qwen3-8B to generate direct \texttt{Yes} and \texttt{No} predictions for each serialized \ac{ehr}, while the encoder experiments pair Qwen3-Emb-8B with a supervised classification head (\cref{subsec:methods/qwen3_finetune}).
Both approaches preserve the EHRSHOT task definitions, operate on the same default list-based \ac{ehr} serialization, and employ the same few-shot sampling strategy across all shot sizes and replicates.
Due to model constraints, we restricted the decoder \ac{icl} experiments to at most \np{6} in-context examples and used a maximum input length of \np{4096} tokens.

The decoder with in-context examples achieved its best performance at \np{2}-shot, with a macro-\ac{auroc} of \citext{0.636}{0.588}{0.685}, compared with \citext{0.611}{0.563}{0.659} in the zero-shot setting (see \cref{fig:qwen3_encoder_decoder_overall}).
This early advantage is consistent with architectural differences between the two approaches: the decoder can directly exploit pretrained logits for \texttt{Yes} and \texttt{No} tokens, whereas embedding variants must learn a classification head from limited supervision.
Increasing the number of in-context examples beyond \np{2} did not lead to further gains.

With more supervision, the fine-tuned decoder improved steadily, reaching macro-\ac{auroc} values of \citext{0.632}{0.584}{0.679} at $k=8$, \citext{0.708}{0.664}{0.752} at $k=64$, and \citext{0.719}{0.673}{0.764} at $k=128$.
The fine-tuned encoder showed a complementary trade-off with slightly lower overall performance, reaching \citext{0.573}{0.524}{0.623} at $k=8$, \citext{0.691}{0.643}{0.737} at $k=64$, and \citext{0.717}{0.669}{0.762} at $k=128$.
Although both fine-tuned variants benefited from additional labeled data, neither surpassed the frozen Qwen3-Emb-8B baseline from the main experiments, which reached \citext{0.733}{0.686}{0.777} at $k=128$ (\cref{tab:ehrshot_k128_encoder_decoder}).

\section{Discussion}
\label{sec:discussion}

This study shows that general-purpose \ac{llm} embedding models pretrained on large-scale natural-language corpora can serve as effective encoders of longitudinal \ac{ehr} data for clinical prediction.
On the EHRSHOT benchmark, Qwen3-Emb-8B performed on par with the domain-specific \ac{ehr} foundation model CLMBR-T-Base \citep{steinberg_language_2021,wornow_ehrshot_2023}, with most task-level differences not reaching statistical significance.
In the external \ac{ukb} validation, Qwen3-Emb-8B performed modestly better overall and showed significant improvements on a subset of tasks.
Our sensitivity analysis, in which Qwen3-Emb-8B was restricted to codes mappable to CLMBR-T-Base, suggests that this advantage reflects both broader vocabulary coverage and slightly better generalization \cite{kirchler_large_2026}.
Together, these findings indicate that \ac{llm} embeddings can match strong \ac{ehr} foundation models on in-domain benchmarks while transferring effectively across settings with different coding practices.
A key practical advantage of Qwen3-Emb-8B is that it operates on a simple list-based \ac{ehr} serialization with minimal preprocessing.
Because it uses natural-language code descriptions rather than a fixed medical vocabulary, it can incorporate arbitrary clinical codes with textual representations, and it does not require institution-specific code mapping.
Despite this simple representation, the model captured clinically meaningful structure, including information from laboratory values embedded in raw \ac{ehr} text.
These results extend prior work on encoder-based models for \ac{ehr} modalities \cite{lee_clinical_2025} and on \ac{llm}-based embeddings at the medical-code level \cite{kirchler_large_2026}.
Overall, \ac{llm} embeddings provide a flexible and powerful alternative for \ac{ehr} representation learning.

Our results also highlight a trade-off between the flexibility of \ac{llm} embeddings and the efficiency of specialized \ac{ehr} foundation models.
CLMBR-T-Base encoded the EHRSHOT benchmark in several minutes with substantially lower memory requirements.
However, it relies on pretraining on a large private \ac{ehr} dataset of approximately \np{1.8} billion clinical events and operates with a fixed vocabulary.
In contrast, Qwen3-Emb-8B required more than 20 hours for encoding, but was pretrained on a much larger and more diverse general-text corpus of approximately \np{36} trillion tokens without requiring access to sensitive clinical data.
The limitations of fixed-vocabulary models became apparent in the \ac{ukb}, where only \np{16}\% of codes could be mapped to the CLMBR-T-Base vocabulary, covering only \np{25}\% of clinical events and requiring a complex, potentially lossy mapping procedure \cite{papez_transforming_2023, lehne_why_2019}.
By operating directly on natural-language descriptions, \ac{llm} embeddings can incorporate all available codes without retraining.
This text-based interface may also facilitate future multimodal extensions that integrate structured records with notes, reports, and imaging metadata \cite{moor_foundation_2023}.
Thus, domain-specific \ac{ehr} models remain substantially more computationally efficient, whereas \ac{llm} embeddings offer greater flexibility and portability across datasets.

Our analyses provide insights into why \ac{llm} embeddings perform well for \ac{ehr} data.
A simple newline-separated list of medical code descriptions was sufficient for strong performance with Qwen3-Emb-8B, indicating that extensive preprocessing or structured input formats were not required.
However, model behavior differed markedly with respect to long-context inputs.
Among similarly sized models, only Qwen3-Emb-8B handled list inputs up to \np{8192} tokens robustly, whereas the other embedding models degraded at longer lengths and often performed better with shorter or more structured inputs such as Markdown.
This suggests that model scale alone does not ensure effective long-context representations of EHRs.
Task-specific instructions further improved performance, indicating that instruction-tuned embeddings can better focus on clinically relevant information \citep{behnamghader_llm2vec_2024,li_towards_2023}.
Performance was driven largely by recent events, although this is partly confounded by the longer inputs induced by larger time windows, and explicit temporal information did not improve results in our setup.
Ablation studies revealed substantial redundancy across \ac{ehr} modalities, with laboratory values being particularly important for lab prediction tasks, but no single modality matched the full representation.
Overall, these findings suggest that \ac{llm} embeddings of \ac{ehr} data depend less on elaborate input engineering and more on model-specific capabilities, particularly robustness to long contexts and sensitivity to clinically relevant signals.

Conventional baselines provide important context for interpreting our results.
Encoder-only language models remain a common approach for \ac{ehr} representation learning \cite{jiang_health_2023, lee_clinical_2025}, but they required input chunking to handle long sequences, which increases computational cost and may limit performance.
Across 14 evaluated configurations, BioClinicalBERT with mean pooling over \np{512}-token chunks performed best among encoder models \cite{alsentzer_publicly_2019}.
In our setting, \ac{llm} embedding models consistently outperformed BioClinicalBERT across multiple few-shot regimes and several tasks with statistically significant differences, although at substantially higher computational cost.
This suggests that \ac{llm} embeddings are a stronger option than encoder-only models for long-context \ac{ehr} representation learning.
At the same time, the count-based model with ontology expansion, string and numeric values, and time binning remained highly competitive when trained on all available data \cite{razavian_population-level_2015, rajkomar_scalable_2018}.
However, its performance deteriorated in few-shot settings, indicating lower data efficiency than pretrained representations.
These findings indicate that the benefits of pretraining are most pronounced under limited supervision and diminish as more labeled data become available.
They also underscore the importance of including strong count-based baselines in \ac{ehr} prediction studies.

We also compared frozen LLM embeddings, decoder \ac{icl}, and parameter-efficiently fine-tuned variants of both model families with two main observations.
First, a small number of in-context examples improved the decoder relative to pure zero-shot prompting, likely because decoder-style models can directly exploit pretrained logits for the constrained \texttt{Yes}/\texttt{No} outputs.
These gains plateaued for \np{4} to \np{6} examples, and larger \ac{icl} settings were not computationally feasible for our long-context \ac{ehr} inputs.
Second, decoder fine-tuning and encoder fine-tuning both improved steadily with additional labels, but neither tuned variant exceeded the frozen Qwen3-Emb-8B baseline within the explored budgets.
This supports using a frozen embedding model with a simple classifier as the default configuration for LLM-based \ac{ehr} prediction.
Decoder \ac{icl} and decoder fine-tuning might remain viable alternatives when very-few-shot performance or task-specific generative behavior justifies their higher optimization and inference costs.

This study has several limitations.
First, our approach relies on a manually designed text-based \ac{ehr} serialization and task-specific instructions, and only captures the most recent occurrence of each code.
Second, comparisons between CLMBR-T-Base and Qwen3-Emb-8B are confounded by large differences in architecture, model scale, and pretraining data, making it difficult to isolate the contribution of each factor.
Third, \ac{llm} embedding models incur substantially higher computational costs, with large memory requirements and long runtimes, and our analyses were limited to contexts of up to \np{4096} to \np{8192} tokens and three \ac{llm} embedding models.
In addition, the hyperparameter tuning for our fine-tuning experiments was limited to two task groups and two shot sizes due to computational constraints, and we only used a decoder with simple token logits for comparison.
Finally, we evaluated two large cohorts, EHRSHOT and \ac{ukb}, but further validation across diverse healthcare systems will be necessary to assess generalizability, robustness, and fairness at scale.

Future work should address these limitations by improving long-context modeling and temporal reasoning in \ac{llm} embeddings.
Serialization-free approaches that allow models to process raw \ac{ehr} tables and timelines may reduce bias introduced by manual text transformation.
Scaling to larger models may provide additional gains in long-context understanding.
At the same time, improving efficiency through distillation or smaller architectures will be important for practical deployment.
It will also be important to evaluate \ac{llm} embedding models on combined structured and unstructured clinical data, including notes, imaging, and reports.
Future work should also evaluate decoder \acp{llm} with stronger learning signals such as LLM-JEPA \citep{huang_llm-jepa_2025} instead of using only simple token logits.
Finally, expanding evaluations to real-world deployments and investigating how to combine the complementary strengths of domain-specific \ac{ehr} models and general-purpose \acp{llm} will be essential for building robust and scalable \ac{ehr} foundation models.

\section{Methods}
\label{sec:methods}

\subsection{EHRSHOT Database and Prediction Task}
\label{subsec:methods/ehrshot_database_and_prediction_tasks}
The EHR data used in our experiments are from the EHRSHOT benchmark for few-shot evaluation of clinical EHR prediction \cite{wornow_ehrshot_2023}.
We obtained version \np{2.1} of the dataset, which is accessible via gated access under a research data use agreement.
This dataset comprises longitudinal records for \np{6739} patients, \np{921499} visits, and \np{41661637} clinical events collected between 1990 and February 8, 2023 (\cref{tab:cohort_ehrshot_ukb_overview}).
Each clinical event is linked to a specific patient and includes information such as start time, end time, a semantic code, a value, unit, visit ID, and the corresponding \ac{omop} source table.
We used the official ehrshot-benchmark GitHub repository as a starting point to design our experiments \cite{wornow_ehrshot_2023}, enabling us to build on existing functionalities and facilitate comparisons with prior methods.
The benchmark uses the \ac{femr} GitHub repository, which provides Python classes for efficient loading and processing of \ac{ehr} data.
All extensions and experiments conducted for this paper are publicly available via our GitHub repository: \url{https://github.com/stefanhgm/ehrshot-benchmark}.
The EHRSHOT benchmark defines a rigorous evaluation including \np{15} clinical prediction tasks categorized into four groups: operational outcomes, anticipating lab-test results, assignment of new diagnoses, and anticipating chest X-ray findings \cite{wornow_ehrshot_2023}.
Task labels are derived from clinical events, so a single patient can contribute multiple labels per task, resulting in substantial variation in task-specific sample sizes (\cref{tab:ehrshot_prediction_tasks_overview}).
For instance, frequent events such as laboratory tests yield many more examples than rarer events such as incident diagnoses.
The benchmark focuses on analyzing model performance in a few-shot setting, which is particularly relevant for large pretrained foundation models \cite{bommasani_opportunities_2022}.
Specifically, for $k \in \{\np{1}, \np{2}, \np{4}, \np{8}, \np{12}, \np{16}, \np{24}, \np{32}, \np{48}, \np{64}, \np{128}\}$, the protocol uses $k$ positive and $k$ negative training examples and $k$ positive and $k$ negative validation examples sampled with replacement to train and tune supervised classifiers.
If fewer than $k$ positive or negative training examples were available, the next larger shot size was used with resampling.
Using the full data for training and validation, which are often unbalanced, is included for all tasks.
Testing is always performed on the full set of available examples for each task.
The classifiers evaluated within the EHRSHOT framework include logistic regression, random forests, and \acp{gbm} \cite{ke_lightgbm_2017}.
Performance is reported using the \ac{auroc}, the \ac{auprc}, and the Brier score.
For few-shot settings, results are averaged over multiple runs with different random seeds, and variability across runs is summarized using the standard deviation of the test set performance \cite{wornow_ehrshot_2023}.
For each individual task, uncertainty is additionally estimated via bootstrap resampling of the test set to obtain standard deviations and percentile-based \np{95}\% confidence intervals.
For task groups, mean performance is computed by averaging across tasks, and approximate \np{95}\% confidence intervals are derived from the uncertainties of the individual task estimates by combining their variances and applying a normal approximation.
An overall macro average is computed by averaging across task groups, with corresponding uncertainty estimated analogously.

\subsection{EHR Text Serialization}
\label{subsec:methods/ehr_text_serialization}

To apply \acp{llm} embedding models to \ac{ehr} data, we serialized each patient record into task-agnostic text.
Following EHRSHOT, we included all events occurring before the label time and truncated inputs to \np{8192} tokens (approximately \np{32000} characters) due to computational constraints (\cref{subsec:computational_setup_and_running_times}).
Each clinical event in EHRSHOT is represented by a semantic identifier of the form “ontology/code”.
We resolved these identifiers to text using the provided vocabularies and custom mappings.
To assess ontology coverage, we analyzed events from \np{200} patients across the operational-outcomes and new-diagnoses task groups (\np{2968} labels).
We observed events from the following ontologies: \ac{loinc}, SNOMED, RxNorm, CPT4, Domain, CARE\textunderscore{}SITE, RxNorm Extension, Medicare Specialty, ICD10PCS, CMS Place of Service, Cancer Modifier, ICD9Proc, CVX, ICDO3, HCPCS, OMOP Extension, and Condition Type.
Codes from CPT4, CARE\textunderscore{}SITE, ICD10PCS, Cancer Modifier, CVX, and ICDO3 were not fully resolved by the provided resources.
We parsed UICC cancer stages from Cancer Modifier codes and added manual mapping files for CPT4, ICD10PCS, and CVX.
We excluded CARE\textunderscore{}SITE and ICDO3 because we could not map them to useful descriptions.

We used a simple newline-separated list of event descriptions with minimal preprocessing to emphasize the embedding model’s ability to interpret raw clinical content.
To remain within the token budget, we kept only the most recent occurrence of each code.
This default serialization therefore corresponds to a list of unique codes per patient record with code-to-description mapping and deduplication as the only mandatory preprocessing steps.
When available, we appended values and units in brackets, and values of datatype float were rounded to two decimals.
We also evaluated variants that kept the first occurrence of each code and that added date and time information, but these did not improve performance (\cref{tab:ehrshot_performance_formats}).

\subsection{Testing Alternative EHR Text Serializations}
\label{subsec:methods/ehr_alternative_text_serialization}

To evaluate whether more structured inputs improve performance, we implemented an alternative Markdown-based serialization, a common \ac{llm} input format \cite{sui_table_2024}.
We included visit dates and the number of days before prediction time, normalizing all dates relative to January 1, 2024, the prediction reference time.
The serialization started with demographics, and birthdates were converted to age in years.
Because \np{65}\% of events were LOINC-coded time series, we aggregated LOINC events.
Using the same \np{200}-patient subset as above, we selected the \np{24} most frequent concepts across vital signs, body metrics, and laboratory values and merged synonymous codes (\cref{tab:semantic_codes_for_aggregated_concepts_in_ehr_serialization}).
For each concept, we reported the last three values and removed implausible measurements.
We additionally added default units and categorical interpretations (low, normal, high) based on standard ranges (\cref{tab:semantic_codes_for_aggregated_concepts_in_ehr_serialization}).
Following the aggregated data, a summary of all visits was included to address the potential truncation of older visits.
Events not associated with visits were then presented, using the same aggregation logic to display the last three values where applicable.
Finally, a detailed reverse-chronological presentation of all visits was included, with events categorized into conditions (SNOMED, Visit, Cancer Modifier, CVX, HCPCS), medications (RxNorm, RxNorm Extension), and procedures (CPT4, ICD10PCS, ICD9Proc).

Additionally, we converted the Markdown serialization into JSON, XML, and YAML using standard Python libraries to assess format effects under a fixed token budget.
Because these formats have different overhead, the amount of clinical content retained can vary for the same token limit.

\subsection{LLM Embedding Models and Baselines}
\label{subsec:methods/llm_embedding_models_and_baselines}

We evaluated three LLM embedding models: Qwen3-Embedding-8B (Qwen3-Emb-8B) \cite{yang_qwen3_2025, zhang_qwen3_2025}, GTE-Qwen2-7B-Instruct (Qwen2-Emb-7B) \cite{yang_qwen2_2024, li_towards_2023}, and LLM2Vec-Llama-3.1-8B-Instruct (Llama3.1-LLM2Vec-8B) \cite{grattafiori_llama_2024, behnamghader_llm2vec_2024}, based on state-of-the-art decoder-only LLMs.
We selected these models for their ability to handle the \np{8192}-token EHR serializations used in our experiments.
For comparison, we also tested smaller variants of Qwen3-Emb-8B and Qwen2-Emb-7B.
All \ac{llm}-based models received task-specific instructions prepended to the \ac{ehr} serialization (\cref{subsec:methods/instructions_for_llm_embedding_models}), and we used each model’s default embedding configuration.
For Qwen-based embedding models, we used the last-layer hidden state of the final \texttt{[EOS]} token as the patient embedding \cite{yang_qwen2_2024, zhang_qwen3_2025}.
For LLM2Vec models, we applied mean pooling over the last-layer hidden states corresponding to the \ac{ehr} serialization, excluding the instruction tokens from pooling \cite{behnamghader_llm2vec_2024}.
Importantly, the instruction still conditions the internal representation and thus influences the resulting embedding.

As additional baselines, we included commonly used encoder-only embedding models with shorter input limits of \np{512} tokens.
To apply them to \np{8192}-token inputs, we split each input into up to 16 chunks of \np{512} tokens and evaluated mean pooling and concatenation of these chunk embeddings to obtain a single representation.
For the biomedical fine-tuned encoders MedBERT and BioClinicalBERT, we also tested the \ac{meme} method to encode domain-specific data separately with a logistic regression classification head to follow our study setup \cite{lee_clinical_2025}.
Overall, this yielded \np{14} encoder-only baseline configurations for long-context EHR prediction.
Encoder-only models did not use task instructions.
Below is an overview of all models.

\textit{Qwen3-Embedding-8B/4B/0.6B} is a model family built on the Qwen3 foundation LLMs \cite{yang_qwen3_2025} and trained as instruction-aware embedding models via a multi-stage recipe \cite{zhang_qwen3_2025}.
All variants use decoder-only Transformers with causal attention and produce embeddings from the last-layer hidden state of the final \texttt{[EOS]} token (\cref{subsec:methods/instructions_for_llm_embedding_models}).
The 8B and 4B models have \np{36} layers, while the 0.6B model has \np{28} layers, and all support a context window of up to \np{32000} tokens.
By default, the embedding dimensionalities are \np{4096} (8B), \np{2560} (4B), and \np{1024} (0.6B), with support for flexible output dimensions.
The base model Qwen3 was trained on a multilingual corpus of 36T tokens \cite{yang_qwen3_2025}.
Training of the embedding models comprises large-scale weakly supervised pretraining on synthetic query–document pairs, followed by supervised fine-tuning on labeled and filtered synthetic data, optimizing a contrastive learning objective.
Finally, different checkpoints are merged to improve robustness \cite{zhang_qwen3_2025}.

\textit{Qwen2-Embedding-7B/1.5B} is based on GTE-Qwen2-7B-Instruct and GTE-Qwen2-1.5B-Instruct \cite{yang_qwen2_2024}.
These models use decoder-only Transformers (7B: \np{28} layers, \np{28} heads, hidden size \np{3584}; 1.5B: \np{28} layers, \np{12} heads, hidden size \np{1536}).
They are first trained with autoregressive next-token prediction and then converted to embedding models via the General Text Embedding (GTE) procedure \cite{li_towards_2023}, which replaces the causal mask with BERT-like bidirectional attention during embedding extraction.
Contrastive learning is applied using a mixture of private datasets to enhance embedding performance.
The models also incorporate instructions tailored for embedding tasks and support a context size of up to \np{32000} tokens.

\textit{Llama3.1-LLM2Vec-8B} is built upon the Llama-3.1-8B-Instruct LLM \cite{grattafiori_llama_2024}, which has a decoder-only Transformer architecture with \np{32} layers, \np{32} attention heads, and a hidden size of \np{4096} (Hugging Face identifier: \texttt{McGill-NLP/LLM2Vec-Meta-Llama-31-8B-Instruct-mntp-supervised}).
Initially trained for next-token prediction, it was converted to an embedding model using the LLM2Vec method \cite{behnamghader_llm2vec_2024}.
This method adds bidirectional attention and fine-tunes the model with supervised contrastive learning on embedding tasks.
Fine-tuning used curated data from the public E5 dataset \cite{springer_repetition_2024, wang_fine-tuning_2024}, containing approximately \np{1.5} million entries.
The model supports task-specific instructions and a context size of up to \np{128000} tokens.

\textit{DeBERTa v3 base/large} are encoder-only Transformer models designed for token embeddings \cite{he_debertav3_2022}.
They improve upon their predecessor by replacing the masked language modeling objective with replaced token detection and using Gradient-Disentangled Embedding Sharing.
We evaluated the base variant (\np{12} layers, \np{12} attention heads, \np{768} hidden size) and the large variant (\np{24} layers, \np{16} attention heads, \np{1024} hidden size), with parameter counts of \np{183}M and \np{434}M, respectively (Hugging Face identifier: \texttt{microsoft/deberta-v3-\{base,large\}}).

\textit{BERT base/large}: BERT is a well-established text embedding model using an encoder-only Transformer trained with the masked language modeling objective \cite{devlin_bert_2019}.
We included both the base (\np{12} layers, \np{12} attention heads, \np{768} hidden size, \np{110}M parameters) and large (\np{24} layers, \np{16} attention heads, \np{1024} hidden size, \np{340}M parameters) variants as benchmarks (Hugging Face identifier: \texttt{google-bert/bert-\{base,large\}-uncased}).
While not state of the art, BERT models remain widely used in embedding tasks.

\textit{Bio$\_$ClinicalBERT} builds on BERT-Base, further fine-tuned on biomedical \cite{lee_biobert_2020} and clinical data \cite{alsentzer_publicly_2019}.
It is a widely adopted embedding model for medical text and was included as a baseline for comparison (Hugging Face identifier: \texttt{emilyalsentzer/Bio$\_$ClinicalBERT}).

\textit{MedBERT} builds on Bio$\_$ClinicalBERT through continued domain pretraining on heterogeneous biomedical corpora (N2C2, BioNLP, CRAFT, and biomedical Wikipedia) and was originally introduced for biomedical NER \cite{vasantharajan_medbert_2022}.
We include it as a domain-adapted baseline for comparison (Hugging Face identifier: \texttt{Charangan/MedBERT}).

\textit{CLMBR-T-Base} is a specialized EHR foundation model trained on \np{2.57} million de-identified EHRs from Stanford Medicine with autoregressive next-code prediction \cite{steinberg_language_2021, wornow_ehrshot_2023}.
With the estimated 706 mean clinical events per patient \cite{wornow_ehrshot_2023}, this leads to approximately 1.8B clinical events for training.
CLMBR-T-Base has \np{12} transformer layers and a hidden dimension of \np{768} (Hugging Face identifier: \texttt{StanfordShahLab/clmbr-t-base}).
The model has \np{141}M parameters and allows for a context window of \np{496} codes.
CLMBR-T-Base has demonstrated improvements over count-based baselines across a variety of clinical prediction tasks \cite{wornow_ehrshot_2023}.
It serves as the main baseline for our experiments to test specialized EHR models against general-purpose text embedding models for representing EHR records.

\textit{LLM Embedding Model and CLMBR-T-Base} is a model combination used to test whether the LLM embedding models and the EHR foundation model learn orthogonal information.
To this end, we simply appended both embeddings.
The resulting embeddings have dimensions of \np{4864} for Qwen3-Emb-8B.

\textit{Count-based Models} have proven to be strong baselines for EHR prediction tasks \cite{rajkomar_scalable_2018, rasmy_med-bert_2021, steinberg_language_2021}.
All EHR events of a patient are encoded in a single vector, where each entry represents the number of occurrences of a medical concept.
We used the count-based baseline introduced in \cite{wornow_ehrshot_2023}, which uses counts for all clinical events that occurred in a patient's timeline prior to the prediction time and extends this approach with ontology expansion, enriching the vectors with parent and child concepts.
We also included three extensions of the above count baseline implemented in the Electronic Medical Records (FEMR) toolkit \cite{steinberg_language_2021}, using (1) time binning of 0–30, 30–180, 180–365, and 365+ days \cite{steinberg_language_2021}, (2) numeric values partitioned into deciles \cite{rajkomar_scalable_2018}, (3) string values \cite{rajkomar_scalable_2018}, and a combination of all three (\cref{tab:ehrshot_count_models}).

Based on the embeddings or count vectors generated by the methods described above, a classification head was trained and validated for each prediction task.
For the embedding models, we used a logistic regression head as default.
For the count-based model, we used a \ac{gbm} \cite{ke_lightgbm_2017} as the primary classifier because it performs better for high-dimensional count features (\cref{tab:ehrshot_performance_on_all_examples_full}).
For completeness, we also report count-based results with a logistic regression head.
We adopted the parameter tuning of the classification heads from the EHRSHOT benchmark to ensure comparability \cite{wornow_ehrshot_2023}.

\subsection{Instructions for LLM Embedding Models}
\label{subsec:methods/instructions_for_llm_embedding_models}
The Qwen2, Qwen3, and LLM2Vec models use instruction-tuned embeddings.
We therefore added simple prompts for each prediction task based on their respective templates.
For instance, for prediction of anemia, we added “Given a patient's electronic healthcare record (EHR) as a newline-separated list, retrieve relevant passages that answer the query: has the patient anemia”.
The existing EHRSHOT benchmark encodes EHRs for the same patient and identical prediction times only once for efficiency.
To support task-specific instructions, we changed this behavior and encoded each (patient, task, prediction time) instance, resulting in \np{1161412} instead of \np{406379} EHRs and longer processing times.
The difference between the \np{1161412} labels used in our experiments and the total number of \np{1178665} labels (\cref{tab:ehrshot_prediction_tasks_overview}) arises because some labels share the same task and prediction time and are therefore merged.
For experiments in \ac{ukb}, we designed analogous instructions.
We list all instructions in \cref{tab:instructions_for_llm_embedding_models} and perform ablations to test their effect.

\subsection{Fine-Tuned LLM Embedding and LLM Decoder Models}
\label{subsec:methods/qwen3_finetune}

The main approach evaluated in this paper used LLM embedding models that encode the EHR serialization into a fixed-length vector prior to classification.
For comparison, we also tested LLM decoder models that generate text whose token probabilities are used for classification.
We further assessed the effect of fine-tuning for both approaches.
For comparability, all encoder, decoder, and decoder-\ac{icl} experiments used the same default list-based \ac{ehr} serialization and the same task-specific instructions.
In contrast to the other experiments, inputs were truncated to \np{4096} tokens due to computational restrictions.

For the encoder setting, we fine-tuned Qwen3-Emb-8B \cite{yang_qwen3_2025, zhang_qwen3_2025} with \ac{lora} adapters \cite{hu_lora_2021} applied to the attention and MLP projection modules (\texttt{q\_proj}, \texttt{k\_proj}, \texttt{v\_proj}, \texttt{o\_proj}, \texttt{up\_proj}, \texttt{down\_proj}, \texttt{gate\_proj}).
The encoder was paired with a lightweight classification head consisting of a dropout layer and a linear projection, and optimized end-to-end with cross-entropy loss.
Inputs were created by prepending the task-specific instruction to the serialized \ac{ehr} and using tokenization with left padding to preserve recency under truncation.

For the decoder setting, we adapted the instruction-tuned causal language model Qwen3-8B (Hugging Face identifier: \texttt{Qwen/Qwen3-8B}) with \ac{lora} adapters on the same projection modules.
The decoder received a chat-style prompt containing the task instruction and the serialized \ac{ehr} (\cref{tab:instructions_for_llm_embedding_models}), and was trained to predict the literal next token \texttt{Yes} or \texttt{No}.
At evaluation time, we scored each example by aggregating the probability mass assigned to single-token variants of \texttt{Yes} and \texttt{No}, including capitalization and punctuation variants, and then normalizing to a two-way probability.

We also evaluated decoder \ac{icl} without weight updates.
These prompts used the same task-specific instruction and deterministic balanced example selection.
We evaluated \np{0}-, \np{2}-, \np{4}-, and \np{6}-shot \ac{icl} only; larger \ac{icl} settings were not computationally feasible.
To limit prompt growth, we applied separate \np{4096}-token caps to the rendered \ac{icl} examples block and to the base prompt containing the target patient record.

For both fine-tuned model families, we kept the optimizer and most training settings fixed and tuned only the main adaptation hyperparameters.
The tuning grid covered learning rates of $5\times10^{-5}$, $10^{-4}$, and $2\times10^{-4}$, LoRA ranks of $8$, $16$, and $64$, and LoRA dropout values of $0.0$, $0.05$, and $0.1$, yielding \np{27} configurations per model family.
We fixed LoRA $\alpha=32$, warmup ratio $0.03$, a maximum of \np{20} epochs, AdamW optimization via the Hugging Face \texttt{Trainer}, cosine learning-rate decay, early stopping with patience \np{5}, gradient checkpointing, and an effective batch size of \np{8}.
Per-task tuning across all EHRSHOT subtasks was computationally infeasible.
We therefore performed tuning on the operational-outcomes and new-diagnosis tasks only (\texttt{guo\_*} and \texttt{new\_*}) at $k \in \{8, 16\}$ using validation \ac{auroc} as the selection criterion, and then chose one shared setting per model family based on mean validation \ac{auroc} across those tasks (\cref{tab:ehrshot_tuning_hparams}).
This selected an encoder setting with learning rate $5\times10^{-5}$, LoRA rank $8$, and dropout $0.1$, and a decoder setting with learning rate $2\times10^{-4}$, LoRA rank $8$, and dropout $0.05$.
Using these fixed settings, we reran the full fine-tuning matrix across all EHRSHOT subtasks, $k \in \{1, 2, 4, 8, 16, 32, 64, 128\}$, and five replicates.
For both model families, evaluation computed \ac{auroc}, \ac{auprc}, and Brier score on the held-out test set together with \np{1000} patient-level bootstrap replicates to estimate standard deviations and \np{95}\% confidence intervals.

\subsection{Existing Methods Using Language Models for EHR Prediction}
\label{subsec:methods/existing_methods}

Two broad approaches have emerged for leveraging language models with structured \ac{ehr} data: generation-based prediction and embedding-based prediction.
Early work on generation-based approaches converted structured claims into short text snippets and queried encoder–decoder models with a 512-token context limit, showing improvements in few-shot regimes (16–64 examples) for end-of-life, surgery, and length-of-stay tasks \cite{hegselmann_tabllm_2023}.
Subsequent studies explored larger \acp{llm}, prompt design, and fine-tuning strategies on serialized \ac{ehr} inputs, typically treating prediction as text generation or calibrated scoring \cite{shoham_cpllm_2024, zhu_prompting_2024, cui_llms-based_2024, acharya_clinical_2024, chen_clinicalbench_2024, makarov_large_2025}.
One variant constrains the model’s output space to medical codes to better align free-text reasoning with code-based clinical labels, thereby bridging the gap between general-purpose language models and \ac{ehr}-specific foundation models \cite{ma_memorize_2025}.

Embedding-based methods instead extract fixed-dimensional representations for downstream classifiers \cite{gao_when_2024, lee_clinical_2025, contreras_dellirium_2024, kirchler_large_2026}.
Gao et~al.\ provide a systematic comparison across serialization formats, \acp{llm}, and embedding strategies on MIMIC-III and a private 660-patient clinical-deterioration dataset, reporting that \ac{llm} embeddings paired with a prediction head can be competitive in some settings, while raw numerical features remain strong baselines \cite{gao_when_2024}.
They also evaluate a decoder-based approach producing \texttt{Yes} and \texttt{No} tokens in the zero-shot setting but find no discriminatory ability for Mistral-7B-Instruct and Llama3-8B-Instruct.
The GRASP framework shows that combining LLM-derived embeddings of medical codes with transformer predictors improves cross-dataset generalization \cite{kirchler_large_2026}.
The DeLLiriuM method introduced in \cite{contreras_dellirium_2024} also evaluates embeddings derived from different \acp{llm} for delirium prediction and performs detailed model introspection via a SHAP analysis.
Most similar to our work is the \ac{meme} method, which encodes different \ac{ehr} modalities such as vitals, medications, and diagnoses into separate embeddings and fuses them with a self-attention layer \cite{lee_clinical_2025}.
This framework uses the relatively small and efficient MedBERT language model for embedding creation and reports superior performance to GPT-4 \cite{lee_clinical_2025}.

Our study follows the embedding paradigm but differs from prior work in three ways.
First, we use \acp{llm} explicitly adapted for embedding via contrastive learning.
Second, we operate with substantially longer inputs (up to 8{,}192 tokens), which is important for longitudinal \ac{ehr} serialization where earlier studies often processed up to 3{,}076 tokens \cite{gao_when_2024} or relied on chunking \cite{lee_clinical_2025}.
Third, we evaluate on large public longitudinal cohorts and standardized benchmarks from EHRSHOT and \ac{ukb}.
We also include a \ac{meme}-style baseline with a linear head in place of self-attention due to data sparsity \cite{lee_clinical_2025}.
Finally, consistent with \cite{gao_when_2024}, we explicitly compare embeddings to direct \ac{llm} outputs and examine parameter-efficient fine-tuning to assess robustness and practical trade-offs (\cref{res:qwen3_finetune}).

\subsection{Computational Setup and Runtime}
\label{subsec:computational_setup_and_running_times}

All experiments were conducted on the Charité High-Performance Cluster offering different GPU setups.
Due to the larger memory requirements of the \acp{llm}, we used a smaller batch size for inference compared to encoder-based models and CLMBR-T-Base.
For the Llama3.1-LLM2Vec-8B model, runtime errors occurred during multi-GPU experiments with the full dataset.
These issues were resolved by splitting the data into smaller batches, which introduced additional overhead.
Additionally, we optimized the LLM2Vec code by removing an initial word-boundary token-limit routine that iteratively re-tokenizes to ensure the final cut occurs at a word boundary, with minimal effect on performance limited to a potentially incomplete trailing word.
To quantify runtime differences, we measured the wall-clock encoding time for the EHRSHOT benchmark on the same 8-GPU Nvidia H200 cluster for CLMBR-T-Base, LLM embedding models, and encoder-based models (\cref{tab:ehrshot_encoding_time}).
We tried to use the maximum possible batch sizes.
Encoder-based models used 512-token chunks of the up to \np{8192}-token inputs, which substantially increased their runtime.
On the \ac{ukb}, encoding all \np{387464} UKB patients for one task with a maximum context length of \np{8192} on the same H200 cluster took 4:46:34 hours for Qwen3-Emb-8B, 3:49:21 hours for Qwen2-Emb-7B, and 3:55:12 hours for Llama3.1-LLM2Vec-8B.

\subsection{Performance Results on EHRSHOT Prediction Tasks and Few-Shot Setting}
Following the EHRSHOT benchmark, we evaluated all models across \np{15} prediction tasks under various few-shot settings.
The benchmark includes a modular pipeline designed to execute key tasks, with the flexibility to optionally use a Slurm cluster for distributed execution.
Running all steps within this pipeline ensures full reproducibility of results.
Step four of the pipeline, which generates EHR representations with CLMBR-T-Base and the count-based model, was extended to incorporate our method for creating language model-based EHR representations.
This adaptation allowed reuse of significant portions of the existing code, including the task evaluation framework.
Additionally, we implemented new functionality for EHR serialization and slightly modified other steps of the benchmark to accommodate our experimental setup.
For instance, the label creation process was adjusted (step three) to enable task-specific instructions for the LLM embedding models.
All modifications have been documented and can be tracked in our public GitHub repository.

\subsection{External Validation on UK Biobank}
\label{subsec:methods/external_validation_ukb}
External validation was performed using data from the \ac{ukb}, a large-scale prospective cohort study comprising \np{502489} UK participants recruited between 2006 and 2010, with a median follow-up of \np{13.8} years (\cref{tab:cohort_ehrshot_ukb_overview}).
We used linked \ac{ehr} data from primary care (General Practitioner (GP)) and secondary care (Hospital Episode Statistics (HES)), providing information on diagnoses, procedures, and prescriptions.
We aimed to closely replicate the experimental setup of the EHRSHOT benchmark by using predefined splits of training, validation, and test data (\cref{tab:ukb_prediction_tasks_overview}), defining tasks and task groups accordingly, and applying the same few-shot analysis, hyperparameter tuning scheme, and statistical evaluation.

Initial data preprocessing, including cleaning, feature extraction, missing-value imputation, and endpoint selection, followed the methodology described in \cite{steinfeldt_medical_2025}.
All health records were mapped to the \ac{omop} CDM using mapping tables provided by the \ac{ukb}, SNOMED International, and the OHDSI community for mapping concepts from provider- and country-specific non-standard vocabularies to \ac{omop} standard vocabularies.
Participants not having full demographic information were excluded.
Participants lacking any recorded GP or HES events either before or after their recruitment date were excluded, resulting in a validation cohort of \np{387464} individuals.
Diagnostic codes were mapped to Phecodes X \cite{wu_mapping_2019,wei_evaluating_2017} primarily for standardized endpoint definition and cohort selection.
To avoid redundancy with medical codes used as features, the Phecodes were excluded during the creation of patient representations.
Due to significant challenges in mapping and harmonizing UKB laboratory values \cite{denaxas_semi-supervised_2020}, and to ensure comparability across models and tasks, medical codes of laboratory values were used without numerical values \cite{steinfeldt_medical_2025}.
The final feature set comprised conditions (SNOMED, CVX), medications (RxNorm), and procedures (SNOMED).

We defined the following tasks for the UKB: (1) prediction of all-cause hospitalization within the next year (operational outcomes), (2) prediction of all-cause mortality (mortality prediction), and (3) prediction of incident diagnoses for a set of selected conditions (assignment of new diagnoses).
The selection of diseases for the assignment of new diagnoses task group largely followed \cite{steinfeldt_medical_2025}, focusing on common conditions, diseases lacking established risk stratification tools, and specific cardiovascular conditions.
From the initial \np{24} endpoints proposed in \cite{steinfeldt_medical_2025}, we treated all-cause mortality as a separate task, leading to a total of 25 tasks.
For the assignment of new diagnoses and mortality tasks, patients with a diagnosis of the respective endpoint recorded prior to and on the day of their UKB recruitment date were excluded.
This exclusion was not applied to the hospitalization task due to the high incidence of hospitalization events before the recruitment date (\cref{tab:ukb_prediction_tasks_overview}).
In the external validation on UKB, each patient had only one prediction date, which was marked by their recruitment date.
Analogously to the main experiments, we used a simple EHR list serialization of the most recent occurrence of each medical code, with birth and race events added at the time of birth.

The pretrained CLMBR-T-Base model operates with a fixed vocabulary of \np{26249} unique codes.
To use this model, we mapped the \np{50702} unique medical codes (SNOMED CT, RxNorm, CVX) present in our processed UKB cohort to the CLMBR-T-Base vocabulary.
This mapping followed steps similar to those implemented in the \ac{femr} package, involving direct code matching where possible, supplemented by indirect mapping via the OHDSI ATHENA vocabulary using ``Maps to'' relationships and inclusion of ancestor concepts.
This process constituted a major effort and required approximately two weeks to complete.
Overall, \np{7969} (16\%) unique UKB codes were successfully mapped to the CLMBR-T-Base vocabulary (in the format ontology/code), which were responsible for 25\% of medical events in the UKB.
The relatively low mapping coverage can be attributed to differences in underlying hospital systems and the distinct purposes of the datasets.
Unmapped codes were excluded for CLMBR-T-Base.
Additionally, UKB ethnicities were converted to the ethnicity groups used by CLMBR-T-Base.
The final data of mapped medical events, birth date, ethnicity, and visits were converted into the MEDS standard \cite{arnrich_medical_2024}.
For transforming patient information into embeddings, we mainly followed the code provided by \ac{femr}, primarily using the \texttt{convert\_patient} function with minor modifications to enable batch processing.
To disentangle effects of vocabulary coverage from generalization capabilities, we performed a sensitivity analysis on \ac{ukb} by restricting Qwen3-Emb-8B to CLMBR-T-Base-mappable codes.
We performed the same few-shot experiments and statistical analyses as for the main experiments to compare Qwen3-Emb-8B with all UKB codes, Qwen3-Emb-8B restricted to CLMBR-T-Base-mappable codes, and CLMBR-T-Base (\cref{tab:qwen_win_tie_loss_shots}, \cref{fig:sensitivity_analysis_auroc_ukb}, and \cref{tab:ukb_significance_qwen_sensitivity}).

Analogously to the best count-based model for EHRSHOT, the count model for the UKB included ontology expansion \cite{wornow_ehrshot_2023}, increasing the number of unique codes from \np{50702} to \np{71393}, and time binning of 0–30, 30–180, 180–365, and 365+ days \cite{steinberg_language_2021}.
In contrast to the EHRSHOT experiments, the count-based model did not include string or numeric values because the UKB did not provide them.
We also added normalized age at prediction time and coded sex as additional features.
Lastly, we evaluated the best encoder-based language model from the EHRSHOT experiments, BioClinicalBERT, on the UKB.

\subsection{Statistical Testing}
To assess the significance of performance differences between the best LLM embedding model, Qwen3-Emb-8B, the EHR foundation model CLMBR-T-Base, the encoder-only model BioClinicalBERT, and the count-based baseline, we performed statistical tests for per-task performance on EHRSHOT and the \ac{ukb}.
We evaluated three training regimes: a very few-shot setting with 8 positive and 8 negative examples, a few-shot setting with 64 positive and 64 negative examples, and training on all available data.
We excluded tasks with insufficient training data for the 64-shot setting.
Few-shot experiments with five-fold cross-validation already evaluate learning efficiency and robustness under limited training data, whereas the statistical tests assess whether the final trained models differ significantly in population-level task performance.
Statistical significance of AUROC differences between models was assessed using a paired, patient-level bootstrap on the held-out test set with \np{10000} bootstrap replicates.
For each shot setting, we used all evaluation replicates by averaging predicted probabilities across replicates for each test example before computing the bootstrap statistic.
For the chest X-ray findings task, which comprises 14 binary sub-tasks, the bootstrap statistic is the macro-averaged AUROC, defined as the mean of per-sub-task AUROCs computed within each bootstrap replicate, matching the macro-average metric reported in the main results.
We report \np{95}\% percentile confidence intervals and two-sided p-values computed from the bootstrap \(\Delta\)AUROC distribution.
For each shot setting, p-values were adjusted jointly across all tasks and baseline comparisons using Holm’s procedure to control the family-wise error rate.
This corresponds to a correction over 45 hypotheses per shot setting for 15 tasks and three model comparisons on EHRSHOT, and 75 (8-shot, all data) and 60 (64-shot) hypotheses for the UKB (\cref{tab:ehrshot_significance_qwen} and \cref{tab:ukb_significance_qwen}).
We also performed an analogous statistical test for the sensitivity analysis of Qwen3-Emb-8B restricted to CLMBR-T-Base-mappable codes on the \ac{ukb} (\cref{tab:ukb_significance_qwen_sensitivity}).

\subsection{Ablation Studies of EHR Serialization}
To better understand the contribution of various components in the EHR serialization process to the performance of the LLM embedding models, we conducted a series of ablation studies.
These ablations used Qwen3-Emb-8B with the default list serialization and a \np{8192}-token limit.
We first examined the role of task-specific instructions by replacing them with a generic prompt (\cref{tab:instructions_for_llm_embedding_models}) or removing instructions entirely, thereby isolating their contribution to the resulting embeddings.
Next, we systematically excluded medical events from six different categories: demographics, visits, medications, procedures, labs, and conditions.
To this end, we grouped medical codes into six mutually exclusive categories based on ontology prefixes and hierarchical SNOMED parent concepts (\cref{tab:code_categories_ablation}).
Finally, we constructed serializations that retained only medical events from one category.

\subsection{Effect of Different Time Windows}
To examine the influence of recency on predictive performance, we varied the time window preceding the prediction date used during EHR text serialization with a maximum context size of \np{8192} tokens.
We evaluated Qwen3-Emb-8B, Qwen2-Emb-7B, Llama3.1-LLM2Vec-8B, and the count-based baseline with a \ac{gbm} head across seven intervals: one hour, one day, one week (\np{7} days), one month (\np{30} days), one year (\np{365} days), three years (\np{1095} days), and full history.
For each window, only events occurring within the specified interval before the prediction time were included.
Thus, only data from the respective time window contributed to the aggregated information and visit data in the \ac{ehr} serialization.
All other aspects of the serialization, including structure, formatting, and instruction prompts, remained unchanged to isolate the effect of the temporal window.

\subsection{Effect of Different Context Sizes}
We investigated the impact of varying context sizes in the LLM embedding models.
Specifically, we evaluated Qwen3-Emb-8B, Qwen2-Emb-7B, and Llama3.1-LLM2Vec-8B with input token limits of \np{512}, \np{1024}, \np{2048}, \np{4096}, and \np{8192} tokens.
Input tokens exceeding these thresholds were discarded.
Due to the EHR list serialization including the most recent occurrences of each medical code, additional input tokens primarily consisted of earlier medical concepts.
All other preprocessing choices and task-specific instructions were held fixed across context-size settings.
By testing these varying context sizes, we aimed to assess the balance between capturing historical medical data and preserving the clarity of high-priority information within the embeddings.

\section*{Data Availability}
The EHRSHOT data are available through gated access via \url{https://doi.org/10.57761/0gv9-nd83}.
UK Biobank data, including all linked routine health records, are publicly available to bona fide researchers upon application at \url{http://www.ukbiobank.ac.uk/using-the-resource}.
In this study, only primary care data not subject to the Government’s Control of Patient Information (COPI) notice were used (UK Biobank Category 3000).

\section*{Code Availability}
All extensions and experiments conducted for this paper are publicly available via our GitHub repository: \url{https://github.com/stefanhgm/ehrshot-benchmark}.

\section*{Acknowledgements}
This work was funded by the Deutsche Forschungsgemeinschaft (DFG, German Research Foundation) – Project-ID 437531118 – SFB 1470 and the German Federal Ministry of Education and Research (BMBF) Project-ID 01ZZ2317G.
Stefan Hegselmann is a participant in the BIH Charité Junior Digital Clinician Scientist Program funded by Charité – Universitätsmedizin Berlin and the Berlin Institute of Health at Charité (BIH).

\section*{Author Contributions}
S.H., G.v.A., and B.W. designed the study.
T.R., N.K., D.S., G.H., and R.E. reviewed and revised the experimental setup.
S.H. implemented and conducted the experiments on EHRSHOT.
G.v.A. and B.W. implemented and conducted the experiments on the UK Biobank.
S.H. and G.v.A. created the initial figures, which were revised and finalized with contributions from T.R., N.K., and B.W.
S.H. drafted the initial manuscript, and G.v.A., T.R., N.K., D.S., G.H., R.E., and B.W. provided substantial feedback.
S.H., G.v.A., and B.W. revised and prepared the final version of the manuscript.
All authors reviewed and approved the final manuscript.

\section*{Competing Interests}
The authors declare no competing interests.

\bibliography{references}

\clearpage
\appendix

\renewcommand\thesection{\arabic{section}}
\renewcommand\theHsection{app.\arabic{section}}
\setcounter{section}{0}

\renewcommand\thesubsection{\thesection.\arabic{subsection}}
\renewcommand\theHsubsection{app.\arabic{section}.\arabic{subsection}}

\setcounter{page}{1}
\pagenumbering{arabic}

\renewcommand{\thefigure}{S\arabic{figure}}
\renewcommand{\theHfigure}{app.S\arabic{figure}}
\setcounter{figure}{0}

\renewcommand{\thetable}{S\arabic{table}}
\renewcommand{\theHtable}{app.S\arabic{table}}
\setcounter{table}{0}

\renewcommand{\figurename}{Figure}
\renewcommand{\tablename}{Table}
\section{Supplementary Information}

\subsection{Additional Experimental Details}

\begin{table}[htbp]
    \caption{\textbf{UK Biobank Prediction Tasks Overview.} The external validation on the UKB includes three task groups. Operational outcomes and mortality prediction only consist of a single prediction task, while the assignment of new diagnosis spans 23 conditions. Following \cite{wornow_ehrshot_2023}, canonical splits for training, validation, and testing were defined.}
    \label{tab:ukb_prediction_tasks_overview}
    \centering
    \footnotesize
    \begin{tabular}{@{}>{\raggedright\arraybackslash}p{3.15cm} p{1.9cm} p{1.9cm} p{1.9cm} p{2.05cm}@{}}
    \toprule
    \textbf{Attribute} & 
    \textbf{\begin{tabular}[c]{@{}l@{}}Train Labels \\ (Positive)\end{tabular}} &
    \textbf{\begin{tabular}[c]{@{}l@{}}Valid Labels \\ (Positive)\end{tabular}} &
    \textbf{\begin{tabular}[c]{@{}l@{}}Test Labels \\ (Positive)\end{tabular}} &
    \textbf{\begin{tabular}[c]{@{}l@{}}Total Labels \\ (Positive)\end{tabular}} \\
    \midrule
\multicolumn{5}{l}{\textbf{Operational Outcomes}} \\
\midrule
Hospitalization & 119463 (22992) & 119463 (22913) & 119463 (22624) & 358389 (68529) \\
\midrule
\multicolumn{5}{l}{\textbf{Mortality Prediction}} \\
\midrule
Death & 119459 (268) & 119463 (284) & 119462 (250) & 358384 (802) \\
\midrule
\multicolumn{5}{l}{\textbf{Assignment of New Diagnoses}} \\
\midrule
Hypertension & 100049 (2515) & 99866 (2532) & 99613 (2600) & 299528 (7647) \\
Diabetes Mellitus & 114423 (621) & 114531 (616) & 114380 (604) & 343334 (1841) \\
Atrial Fibrillation & 118804 (107) & 118784 (94) & 118704 (105) & 356292 (306) \\
Pneumonia & 117409 (270) & 117527 (299) & 117437 (232) & 352373 (801) \\
Chronic Obstructive Pulmonary Disease [COPD] & 117868 (305) & 117899 (284) & 117886 (334) & 353653 (923) \\
Chronic Kidney Disease & 117542 (308) & 117512 (346) & 117478 (347) & 352532 (1001) \\
Ischemic Heart Disease & 113627 (609) & 113634 (583) & 113467 (596) & 340728 (1788) \\
Myocardial Infarction [heart Attack] & 117049 (214) & 117070 (204) & 116995 (201) & 351114 (619) \\
Cerebral Infarction [ischemic Stroke] & 118788 (100) & 118788 (116) & 118745 (94) & 356321 (310) \\
Heart Failure & 118563 (152) & 118626 (158) & 118609 (125) & 355798 (435) \\
Cardiac Arrest & 119370 (25) & 119362 (29) & 119370 (33) & 358102 (87) \\
Abdominal Aortic Aneurysm & 119359 (17) & 119352 (17) & 119354 (39) & 358065 (73) \\
Pulmonary Embolism & 118750 (81) & 118795 (99) & 118754 (68) & 356299 (248) \\
Aortic Stenosis & 119170 (38) & 119174 (37) & 119179 (39) & 357523 (114) \\
Mitral Valve Insufficiency & 119003 (63) & 119061 (58) & 119024 (65) & 357088 (186) \\
Endocarditis & 119005 (24) & 119012 (24) & 118998 (31) & 357015 (79) \\
Rheumatic Fever And Chronic Rheumatic Heart Diseases & 119047 (51) & 119073 (46) & 119083 (57) & 357203 (154) \\
Anemia & 114135 (606) & 114119 (587) & 114142 (620) & 342396 (1813) \\
Back Pain & 100129 (1323) & 99663 (1322) & 99876 (1285) & 299668 (3930) \\
Parkinson's Disease (primary) & 119278 (27) & 119306 (31) & 119284 (28) & 357868 (86) \\
Rheumatoid Arthritis & 118392 (117) & 118400 (107) & 118380 (92) & 355172 (316) \\
Psoriasis & 117482 (148) & 117461 (119) & 117594 (142) & 352537 (409) \\
Suicide Ideation And Attempt Or Self Harm & 118953 (62) & 118896 (60) & 118943 (53) & 356792 (175) \\
    \bottomrule
    \end{tabular}
\end{table}

\clearpage
\begin{table}[]
    \caption{\textbf{Semantic Codes for Aggregated Concepts in EHR Markdown Serialization.} We aggregated time-series data encoded via \ac{loinc} concepts and identified the most frequent concepts from which we selected 24 key medical concepts. To reduce duplicate information, we merged synonymous semantic codes. The primary \ac{loinc} code is presented first in the column Semantic Codes followed by identified duplicates.  We also defined a unit, minimum and maximum allowed  values for filtering, a normal range to classify values in low, normal, and high, and a formatting strategy to create our EHR serialization.}
    \label{tab:semantic_codes_for_aggregated_concepts_in_ehr_serialization}
    \centering
    \setlength{\tabcolsep}{3pt} 
    \renewcommand{\arraystretch}{0.92}
    \footnotesize
    \begin{tabularx}{\textwidth}{@{}X p{3cm} p{1.6cm} p{1.3cm} p{1.2cm} p{1.7cm}@{}} \toprule
    \textbf{Medical Concept}  & \textbf{Semantic Codes}                                                        & \textbf{Unit} & \textbf{Min-Max Range} & \textbf{Normal Range} & \textbf{Formatting} \\ \midrule
    \multicolumn{6}{@{}l@{}}{\textbf{Recent Body Metrics}} \\ \midrule
    Body weight               & LOINC/29463-7                                                                  & oz            & 350-10000              &                       & One decimal         \\
    Body height               & LOINC/8302-2                                                                   & inch          & 5-100                  &                       & One decimal         \\
    Body mass index /   BMI   & LOINC/39156-5                                                                  & kg/m2         & 10-100                 & 18.5-24.9             & One decimal         \\
    Body surface area         & LOINC/8277-6, SNOMED/301898006                                                 & m2            & 0.1-10                 &                       & Two decimals        \\ \midrule
    \multicolumn{6}{@{}l@{}}{\textbf{Recent Vital Signs}} \\ \midrule
    Heart rate                & LOINC/8867-4,   SNOMED/364075005, SNOMED/78564009                              & bpm           & 5-300                  & 60-100                & Integer             \\
    Systolic blood   pressure & LOINC/8480-6,   SNOMED/271649006                                               & mmHg          & 20-300                 & 90-140                & Integer             \\
    Diastolic blood pressure  & LOINC/8462-4,   SNOMED/271650006                                               & mmHg          & 20-300                 & 60-90                 & Integer             \\
    Body temperature          & LOINC/8310-5                                                                   & °F            & 80-120                 & 95-100.4              & One decimal         \\
    Respiratory rate          & LOINC/9279-1                                                                   & breaths/min   & 1-100                  & 12-18                 & Integer             \\
    Oxygen saturation         & LOINC/LP21258-6                                                                & \%            & 1-100                  & 95-100                & Integer             \\ \midrule
    \multicolumn{6}{@{}l@{}}{\textbf{Recent Lab Results}} \\ \midrule
    Hemoglobin                & LOINC/718-7,   SNOMED/271026005, SNOMED/441689006                              & g/dL          & 1-20                   & 12-17                 & One decimal         \\
    Hematocrit                & LOINC/4544-3,   LOINC/20570-8, LOINC/48703-3, SNOMED/28317006                  & \%            & 10-100                 & 36-51                 & Integer             \\
    Erythrocytes              & LOINC/789-8,   LOINC/26453-1                                                   & 106/uL        & 1-10                   & 4.2-5.9               & Two decimals        \\
    Leukocytes                & LOINC/20584-9,   LOINC/6690-2                                                  & 103/uL        & 1-100                  & 4-10                  & One decimal         \\
    Platelets                 & LOINC/777-3,   SNOMED/61928009                                                 & 103/uL        & 10-1000                & 150-350               & Integer             \\
    Sodium                    & LOINC/2951-2,   LOINC/2947-0, SNOMED/25197003                                  & mmol/L        & 100-200                & 136-145               & Integer             \\
    Potassium                 & LOINC/2823-3,   SNOMED/312468003, LOINC/6298-4, SNOMED/59573005                & mmol/L        & 0.1-10                 & 3.5-5.0               & One decimal         \\
    Chloride                  & LOINC/2075-0,   SNOMED/104589004, LOINC/2069-3                                 & mmol/L        & 50-200                 & 98-106                & Integer             \\
    Carbon dioxide,   total   & LOINC/2028-9                                                                   & mmol/L        & 10-100                 & 23-28                 & Integer             \\
    Calcium                   & LOINC/17861-6,   SNOMED/271240001                                              & mg/dL         & 1-20                   & 9-10.5                & One decimal         \\
    Glucose                   & LOINC/2345-7,   SNOMED/166900001, LOINC/2339-0, SNOMED/33747003, LOINC/14749-6 & mg/dL         & 10-1000                & 70-100                & Integer             \\
    Urea nitrogen             & LOINC/3094-0,   SNOMED/105011006                                               & mg/dL         & 1-200                  & 8-20                  & Integer             \\
    Creatinine                & LOINC/2160-0,   SNOMED/113075003                                               & mg/dL         & 0.1-10                 & 0.7-1.3               & One decimal         \\
    Anion gap                 & LOINC/33037-3,   LOINC/41276-7, SNOMED/25469001                                & mmol/L        & -20-50                 & 3-11                  & Integer \\ \bottomrule
\end{tabularx}
\end{table}

\clearpage
\begin{table}[h]
    \caption{\textbf{Instructions for LLM Embedding Models.} The LLM embedding models were trained using instructions; hence, we also defined simple task-specific prompts for each of the 15 clinical prediction tasks. Each prompt is prepended by the prefix given below, containing a general task description. The three tasks used for the external validation on UKB use a similar design. For the decoder model we added an additional instruction to the prompt enforcing the output of \texttt{Yes} and \texttt{No} tokens used for prediction.}
    \label{tab:instructions_for_llm_embedding_models}
    \centering
    \begin{tabular}{@{}p{3.9cm} p{8.4cm}@{}} \toprule
    \textbf{Task}        & \textbf{Prompt} \\ \midrule
    Prefix (for all tasks) & Given a patient's electronic healthcare record (EHR) as a newline separated list, retrieve relevant passages that answer the query: \\ \midrule
    \textbf{EHRSHOT} \\ \midrule
    Long Length of Stay  & will the patient stay in the hospital for more than 7 days \\
    30-day Readmission   & will the patient be readmitted to the hospital within 30 days \\
    ICU Transfer         & will the patient be transferred to the intensive care unit \\
    Thrombocytopenia     & has the patient thrombocytopenia \\
    Hyperkalemia         & has the patient hyperkalemia \\
    Hypoglycemia         & has the patient hypoglycemia \\
    Hyponatremia         & has the patient hyponatremia \\
    Anemia               & has the patient anemia \\
    Hypertension         & has the patient hypertension \\
    Hyperlipidemia       & has the patient hyperlipidemia \\
    Pancreatic Cancer    & has the patient pancreatic cancer \\
    Celiac               & has the patient celiac disease \\
    Lupus                & has the patient lupus \\
    Acute MI             & has the patient an acute myocardial infarction \\
    Chest X-Ray Findings & what are the chest x-ray findings of the patient \\ 
    Generic (ablation)   & what are the key clinical features of the patient to predict future medical events \\
    \midrule
    \textbf{UK Biobank (UKB)} \\ \midrule
    Mortality Prediction & will the patient die within one year \\
    Hospitalization      & will the patient be admitted to the hospital within one year \\
    Assignment of New Diagnoses & has the patient \{diagnosis name\} \\
    \midrule
    \multicolumn{2}{@{}l@{}}{\textbf{Additional Prompt Added for Decoder Qwen3-8B}} \\ \midrule
    Decoder Prompt & Answer STRICTLY with a single token: Yes or No. No punctuation, no extra words. \\
    \bottomrule
    \end{tabular}
\end{table}

\clearpage

\subsection{Full Results on EHRSHOT}
\begin{table}[h]
    \caption{\textbf{Performance for All Examples on EHRSHOT.} Mean \acf{auroc} performance with approximate 95\% confidence intervals of all included models for four task groups. The macro-averaged performance across all task groups is given in the right-most column. All LLM embedding models use a context size of \np{8192} tokens.}
    \label{tab:ehrshot_performance_on_all_examples_full}
    \centering
    \footnotesize
    \setlength{\tabcolsep}{2.6pt} 
    \begin{tabular}{>{\raggedright\arraybackslash}p{3.1cm} 
                >{\raggedright\arraybackslash}p{1.8cm} 
                >{\raggedright\arraybackslash}p{1.8cm} 
                >{\raggedright\arraybackslash}p{1.8cm} 
                >{\raggedright\arraybackslash}p{1.8cm} 
                >{\raggedright\arraybackslash}p{1.8cm}@{}}
    \toprule
\textbf{Model}                           & \textbf{Operational Outcomes} & \textbf{Anticipating Lab Test Results} & \textbf{Assignment of New Diagnosis} & \textbf{Anticipating Chest X-ray Findings} & \textbf{Macro Avg. Across Task Groups} \\ \midrule
\multicolumn{6}{l}{\textbf{Baselines} \cite{wornow_ehrshot_2023}} \\ \midrule
CLMBR-T-Base                             & $\ci{0.824}{.803}{.845}$ & $\ci{0.832}{.824}{.840}$ & $\ci{0.707}{.667}{.746}$ & $\ci{0.713}{.702}{.724}$ & $\ci{0.769}{.746}{.792}$ \\
Count-based + GBM                       & $\ci{0.824}{.804}{.844}$ & $\ci{0.841}{.833}{.849}$ & $\ci{0.758}{.724}{.793}$ & $\ci{0.686}{.674}{.699}$ & $\ci{0.777}{.756}{.799}$ \\
Count-based + LR                        &$\ci{0.764}{.741}{.787}$ & $\ci{0.742}{.729}{.756}$ & $\ci{0.734}{.687}{.782}$ & $\ci{0.673}{.656}{.690}$ & $\ci{0.728}{.700}{.757}$ \\ \midrule
\multicolumn{6}{l}{\textbf{LLM Embedding Models}} \\ \midrule
Qwen3-Emb-8B                             & $\ci{0.797}{.773}{.820}$ & $\ci{0.842}{.835}{.850}$ & $\ci{0.714}{.672}{.757}$ & $\ci{0.722}{.711}{.733}$ & $\ci{0.769}{.744}{.794}$ \\
Qwen3-Emb-4B                             & $\ci{0.787}{.764}{.810}$ & $\ci{0.824}{.816}{.831}$ & $\ci{0.718}{.667}{.768}$ & $\ci{0.708}{.696}{.719}$ & $\ci{0.759}{.730}{.787}$ \\
Qwen3-Emb-0.6B                           & $\ci{0.778}{.753}{.803}$ & $\ci{0.742}{.732}{.753}$ & $\ci{0.684}{.631}{.737}$ & $\ci{0.705}{.694}{.716}$ & $\ci{0.727}{.697}{.758}$ \\
Qwen2-Emb-7B                             & $\ci{0.772}{.749}{.796}$ & $\ci{0.746}{.735}{.757}$ & $\ci{0.744}{.705}{.784}$ & $\ci{0.685}{.671}{.699}$ & $\ci{0.737}{.712}{.761}$ \\
Qwen2-Emb-1.5B                           & $\ci{0.756}{.735}{.778}$ & $\ci{0.710}{.699}{.721}$ & $\ci{0.696}{.647}{.744}$ & $\ci{0.680}{.668}{.691}$ & $\ci{0.710}{.683}{.738}$ \\
Llama3.1-LLM2Vec-8B                      & $\ci{0.763}{.738}{.787}$ & $\ci{0.726}{.714}{.738}$ & $\ci{0.727}{.688}{.766}$ & $\ci{0.686}{.673}{.699}$ & $\ci{0.725}{.701}{.750}$ \\ \midrule
\multicolumn{6}{l}{\textbf{LLM Embedding Model + EHR Foundation Model} \cite{wornow_ehrshot_2023}} \\ \midrule
Qwen3-Emb-8B + CLMBR-T-Base              & $\ci{0.821}{.800}{.842}$ & $\ci{0.864}{.858}{.871}$ & $\ci{0.736}{.695}{.777}$ & $\ci{0.731}{.721}{.742}$ & $\ci{0.788}{.764}{.812}$ \\ \midrule
\multicolumn{6}{l}{\textbf{LLM Embedding Model + GBM Head}} \\ \midrule
Qwen3-Emb-8B + GBM           & $\ci{0.774}{.749}{.799}$ & $\ci{0.812}{.804}{.820}$ & $\ci{0.685}{.644}{.727}$ & $\ci{0.696}{.686}{.706}$ & $\ci{0.742}{.717}{.767}$ \\ \midrule
\multicolumn{6}{l}{\textbf{Multiple Embedding Model for EHR (MEME)} \cite{lee_clinical_2025} \textbf{with Linear Head} } \\ \midrule
Qwen3-Emb-8B MEME                        & $\ci{0.814}{.793}{.834}$ & $\ci{0.845}{.837}{.852}$ & $\ci{0.728}{.673}{.784}$ & $\ci{0.717}{.705}{.728}$ & $\ci{0.776}{.746}{.806}$ \\
BioClinicalBERT MEME                             & $\ci{0.756}{.733}{.778}$ & $\ci{0.699}{.686}{.713}$ & $\ci{0.704}{.651}{.758}$ & $\ci{0.648}{.635}{.661}$ & $\ci{0.702}{.671}{.732}$ \\ 
MedBERT MEME                             & $\ci{0.759}{.736}{.782}$ & $\ci{0.695}{.682}{.708}$ & $\ci{0.715}{.676}{.755}$ & $\ci{0.620}{.605}{.635}$ & $\ci{0.697}{.673}{.722}$ \\ \midrule
\multicolumn{6}{l}{\textbf{Encoder Language Models with Mean Embeddings of Chunked Inputs}} \\ \midrule
BioClinicalBERT                             & $\ci{0.738}{.712}{.763}$ & $\ci{0.698}{.685}{.711}$ & $\ci{0.707}{.668}{.746}$ & $\ci{0.679}{.666}{.691}$ & $\ci{0.705}{.680}{.730}$ \\
MedBERT                                  & $\ci{0.742}{.718}{.767}$ & $\ci{0.694}{.683}{.706}$ & $\ci{0.663}{.614}{.713}$ & $\ci{0.683}{.671}{.696}$ & $\ci{0.696}{.667}{.725}$ \\
DeBERTaV3 large                          & $\ci{0.737}{.713}{.762}$ & $\ci{0.694}{.681}{.706}$ & $\ci{0.676}{.629}{.724}$ & $\ci{0.660}{.648}{.673}$ & $\ci{0.692}{.664}{.720}$ \\
DeBERTaV3 base                           & $\ci{0.727}{.701}{.753}$ & $\ci{0.680}{.668}{.693}$ & $\ci{0.672}{.635}{.710}$ & $\ci{0.659}{.647}{.672}$ & $\ci{0.685}{.660}{.709}$ \\
BERT large                               & $\ci{0.748}{.725}{.771}$ & $\ci{0.688}{.675}{.700}$ & $\ci{0.671}{.625}{.718}$ & $\ci{0.669}{.656}{.682}$ & $\ci{0.694}{.667}{.722}$ \\
BERT base                                & $\ci{0.748}{.724}{.772}$ & $\ci{0.693}{.681}{.706}$ & $\ci{0.685}{.641}{.729}$ & $\ci{0.677}{.664}{.690}$ & $\ci{0.701}{.674}{.727}$ \\ \midrule
\multicolumn{6}{l}{\textbf{Encoder Language Models with Concatenated Embeddings of Chunked Inputs}} \\ \midrule
BioClinicalBERT                             & $\ci{0.740}{.715}{.764}$ & $\ci{0.707}{.696}{.718}$ & $\ci{0.690}{.651}{.730}$ & $\ci{0.653}{.641}{.665}$ & $\ci{0.698}{.673}{.722}$ \\
MedBERT                                  & $\ci{0.751}{.726}{.775}$ & $\ci{0.712}{.701}{.723}$ & $\ci{0.658}{.618}{.698}$ & $\ci{0.658}{.646}{.670}$ & $\ci{0.695}{.670}{.720}$ \\
DeBERTaV3 large                          & $\ci{0.714}{.687}{.742}$ & $\ci{0.690}{.679}{.701}$ & $\ci{0.632}{.595}{.669}$ & $\ci{0.624}{.612}{.636}$ & $\ci{0.665}{.641}{.690}$ \\
DeBERTaV3 base                           & $\ci{0.675}{.647}{.703}$ & $\ci{0.680}{.670}{.690}$ & $\ci{0.623}{.583}{.662}$ & $\ci{0.615}{.603}{.628}$ & $\ci{0.648}{.623}{.674}$ \\
BERT large                               & $\ci{0.736}{.710}{.762}$ & $\ci{0.697}{.686}{.708}$ & $\ci{0.657}{.613}{.701}$ & $\ci{0.643}{.631}{.655}$ & $\ci{0.683}{.657}{.710}$ \\
BERT base                                & $\ci{0.752}{.728}{.776}$ & $\ci{0.705}{.695}{.716}$ & $\ci{0.657}{.613}{.701}$ & $\ci{0.650}{.639}{.662}$ & $\ci{0.691}{.665}{.717}$ \\
\bottomrule
\end{tabular}
\end{table}

\clearpage
\newcommand{\stripzero}[1]{%
  \begingroup
  \edef\temp{#1}%
  \expandafter\stripzeroaux\temp\relax
  \endgroup
}

\def\stripzeroaux#1#2.#3\relax{%
  \ifx#1+%
    \ifnum#2=0 +.#3\else +#2.#3\fi
  \else\ifx#1-%
    \ifnum#2=0 -.#3\else -#2.#3\fi
  \else
    \ifnum#1=0 .#3\else #1.#3\fi
  \fi\fi
}

\newcommand{\estcip}[4]{%
$#1_{\scriptsize[#2,#3]}$ {\tiny(#4)}%
}

\newcommand{\bestcip}[4]{%
$\mathbf{#1}_{\scriptsize[#2,#3]}$ {\tiny(#4)}%
}

\begin{table}[]
\caption{\textbf{Per-task $\Delta$AUROC (Qwen3-Emb-8B minus baseline) on EHRSHOT.}
Cells report the AUROC difference ($\Delta$AUROC), 95\% bootstrap confidence intervals, and Holm-adjusted $p$-values obtained from paired patient-level bootstrap tests with 10{,}000 resamples. Positive values indicate better performance of Qwen3-Emb-8B. 
Multiple testing was controlled separately for each shot setting ($k=8$, $k=64$, and all training data) using Holm’s procedure across all tasks and baseline comparisons (45 tests per setting). Bold indicates statistically significant differences ($p_{\text{adj}}<0.05$).}
\label{tab:ehrshot_significance_qwen}
\centering
\footnotesize
\setlength{\tabcolsep}{2pt}
\begin{tabular}{lccc}
\toprule
\textbf{Task} & \textbf{CLMBR-T-Base} & \textbf{BioClinicalBERT} & \textbf{Count-based Model} \\
\midrule
\multicolumn{4}{l}{\textbf{8-shot (8 positive / 8 negative examples)}} \\
\midrule
Long LOS & \estcip{+0.011}{-0.025}{+0.047}{1.000} & \estcip{-0.001}{-0.034}{+0.031}{1.000} & \estcip{-0.042}{-0.080}{-0.002}{0.677} \\
30-day Readmission & \estcip{+0.001}{-0.034}{+0.034}{1.000} & \bestcip{+0.079}{+0.038}{+0.121}{0.009} & \bestcip{+0.085}{+0.039}{+0.131}{0.009} \\
ICU Transfer & \estcip{-0.043}{-0.089}{+0.002}{1.000} & \estcip{+0.071}{+0.009}{+0.132}{0.572} & \estcip{+0.069}{-0.015}{+0.151}{1.000} \\
Thrombocytopenia & \bestcip{+0.036}{+0.014}{+0.057}{0.041} & \bestcip{+0.089}{+0.068}{+0.110}{0.009} & \bestcip{+0.100}{+0.074}{+0.123}{0.009} \\
Hyperkalemia & \bestcip{+0.113}{+0.082}{+0.141}{0.009} & \bestcip{+0.197}{+0.160}{+0.236}{0.009} & \bestcip{+0.194}{+0.146}{+0.241}{0.009} \\
Hypoglycemia & \estcip{-0.042}{-0.080}{-0.004}{0.584} & \estcip{+0.050}{+0.009}{+0.089}{0.415} & \estcip{+0.024}{-0.021}{+0.067}{1.000} \\
Hyponatremia & \bestcip{+0.029}{+0.011}{+0.047}{0.041} & \bestcip{+0.085}{+0.061}{+0.109}{0.009} & \bestcip{+0.070}{+0.046}{+0.095}{0.009} \\
Anemia & \bestcip{-0.067}{-0.080}{-0.056}{0.009} & \bestcip{+0.145}{+0.130}{+0.161}{0.009} & \bestcip{+0.032}{+0.021}{+0.043}{0.009} \\
Hypertension & \estcip{-0.068}{-0.125}{-0.012}{0.415} & \estcip{-0.012}{-0.060}{+0.037}{1.000} & \estcip{+0.002}{-0.065}{+0.070}{1.000} \\
Hyperlipidemia & \estcip{+0.019}{-0.028}{+0.069}{1.000} & \estcip{+0.030}{-0.027}{+0.088}{1.000} & \estcip{+0.076}{+0.011}{+0.141}{0.474} \\
Pancreatic Cancer & \estcip{+0.089}{+0.008}{+0.174}{0.600} & \estcip{+0.071}{-0.024}{+0.164}{1.000} & \bestcip{+0.167}{+0.074}{+0.259}{0.030} \\
Celiac & \estcip{+0.070}{-0.113}{+0.299}{1.000} & \estcip{+0.209}{-0.166}{+0.453}{1.000} & \estcip{+0.009}{-0.262}{+0.232}{1.000} \\
Lupus & \estcip{+0.054}{-0.088}{+0.184}{1.000} & \estcip{+0.123}{-0.005}{+0.270}{1.000} & \estcip{+0.063}{-0.097}{+0.237}{1.000} \\
Acute MI & \bestcip{+0.126}{+0.056}{+0.191}{0.025} & \bestcip{+0.161}{+0.081}{+0.237}{0.009} & \bestcip{+0.172}{+0.088}{+0.251}{0.009} \\
Chest X-ray Findings & \estcip{-0.015}{-0.028}{-0.001}{0.600} & \estcip{+0.024}{+0.009}{+0.037}{0.062} & \bestcip{+0.026}{+0.010}{+0.042}{0.043} \\
\midrule
\multicolumn{4}{l}{\textbf{64-shot (64 positive / 64 negative examples)}} \\
\midrule
Long LOS & \estcip{-0.037}{-0.060}{-0.014}{0.054} & \estcip{+0.019}{-0.011}{+0.049}{1.000} & \estcip{-0.025}{-0.054}{+0.004}{1.000} \\
30-day Readmission & \estcip{-0.010}{-0.029}{+0.009}{1.000} & \bestcip{+0.050}{+0.030}{+0.069}{0.009} & \estcip{+0.022}{+0.002}{+0.041}{0.846} \\
ICU Transfer & \estcip{-0.069}{-0.123}{-0.014}{0.418} & \estcip{+0.056}{-0.007}{+0.120}{1.000} & \estcip{-0.041}{-0.097}{+0.016}{1.000} \\
Thrombocytopenia & \bestcip{+0.023}{+0.009}{+0.036}{0.026} & \bestcip{+0.078}{+0.066}{+0.092}{0.009} & \bestcip{-0.047}{-0.060}{-0.036}{0.009} \\
Hyperkalemia & \bestcip{+0.044}{+0.026}{+0.062}{0.009} & \bestcip{+0.143}{+0.115}{+0.169}{0.009} & \estcip{+0.020}{-0.001}{+0.040}{1.000} \\
Hypoglycemia & \bestcip{-0.044}{-0.070}{-0.018}{0.043} & \bestcip{+0.094}{+0.057}{+0.131}{0.009} & \estcip{+0.034}{+0.002}{+0.065}{1.000} \\
Hyponatremia & \estcip{+0.016}{-0.003}{+0.035}{1.000} & \bestcip{+0.118}{+0.097}{+0.137}{0.009} & \bestcip{-0.072}{-0.091}{-0.055}{0.009} \\
Anemia & \bestcip{-0.056}{-0.064}{-0.049}{0.009} & \bestcip{+0.156}{+0.141}{+0.172}{0.009} & \estcip{+0.004}{-0.003}{+0.011}{1.000} \\
Hypertension & \estcip{-0.038}{-0.073}{-0.002}{1.000} & \estcip{+0.028}{-0.018}{+0.074}{1.000} & \estcip{+0.002}{-0.040}{+0.045}{1.000} \\
Hyperlipidemia & \estcip{+0.030}{-0.009}{+0.068}{1.000} & \estcip{+0.037}{-0.013}{+0.087}{1.000} & \estcip{+0.020}{-0.021}{+0.063}{1.000} \\
Pancreatic Cancer & \bestcip{+0.090}{+0.038}{+0.148}{0.038} & \estcip{+0.073}{+0.001}{+0.141}{1.000} & \estcip{+0.011}{-0.042}{+0.071}{1.000} \\
Celiac & \estcip{+0.101}{-0.080}{+0.257}{1.000} & \estcip{+0.169}{-0.121}{+0.317}{1.000} & \estcip{-0.029}{-0.138}{+0.100}{1.000} \\
Lupus & \estcip{-0.079}{-0.205}{+0.040}{1.000} & \estcip{+0.019}{-0.122}{+0.163}{1.000} & \estcip{+0.039}{-0.133}{+0.220}{1.000} \\
Acute MI & \estcip{-0.028}{-0.066}{+0.010}{1.000} & \estcip{-0.047}{-0.089}{-0.002}{1.000} & \estcip{-0.041}{-0.083}{+0.002}{1.000} \\
Chest X-ray Findings & \estcip{+0.000}{-0.009}{+0.010}{1.000} & \bestcip{+0.031}{+0.018}{+0.045}{0.009} & \bestcip{+0.021}{+0.010}{+0.032}{0.014} \\\midrule
\multicolumn{4}{l}{\textbf{All training data}} \\
\midrule
Long LOS & \estcip{-0.006}{-0.024}{+0.012}{1.000} & \bestcip{+0.079}{+0.058}{+0.100}{0.009} & \estcip{-0.018}{-0.035}{-0.001}{0.840} \\
30-day Readmission & \estcip{-0.015}{-0.037}{+0.007}{1.000} & \bestcip{+0.045}{+0.025}{+0.067}{0.011} & \estcip{-0.002}{-0.029}{+0.026}{1.000} \\
ICU Transfer & \estcip{-0.060}{-0.110}{-0.012}{0.463} & \estcip{+0.053}{+0.005}{+0.103}{0.638} & \estcip{-0.062}{-0.115}{-0.007}{0.658} \\
Thrombocytopenia & \bestcip{+0.045}{+0.035}{+0.054}{0.009} & \bestcip{+0.116}{+0.100}{+0.132}{0.009} & \bestcip{-0.040}{-0.049}{-0.032}{0.009} \\
Hyperkalemia & \bestcip{+0.030}{+0.015}{+0.045}{0.009} & \bestcip{+0.142}{+0.114}{+0.172}{0.009} & \estcip{+0.022}{+0.006}{+0.038}{0.178} \\
Hypoglycemia & \bestcip{-0.035}{-0.053}{-0.019}{0.009} & \bestcip{+0.146}{+0.115}{+0.176}{0.009} & \bestcip{+0.061}{+0.036}{+0.086}{0.009} \\
Hyponatremia & \bestcip{+0.034}{+0.022}{+0.046}{0.009} & \bestcip{+0.179}{+0.163}{+0.195}{0.009} & \bestcip{-0.027}{-0.037}{-0.017}{0.009} \\
Anemia & \bestcip{-0.021}{-0.023}{-0.018}{0.009} & \bestcip{+0.139}{+0.126}{+0.153}{0.009} & \bestcip{-0.009}{-0.012}{-0.006}{0.009} \\
Hypertension & \estcip{-0.042}{-0.084}{-0.002}{0.840} & \estcip{+0.036}{-0.009}{+0.082}{1.000} & \estcip{-0.033}{-0.079}{+0.011}{1.000} \\
Hyperlipidemia & \estcip{+0.022}{-0.017}{+0.063}{1.000} & \estcip{+0.017}{-0.024}{+0.057}{1.000} & \estcip{-0.018}{-0.064}{+0.028}{1.000} \\
Pancreatic Cancer & \estcip{+0.049}{+0.002}{+0.100}{0.840} & \estcip{+0.052}{-0.012}{+0.110}{1.000} & \estcip{-0.022}{-0.072}{+0.019}{1.000} \\
Celiac & \estcip{+0.023}{-0.153}{+0.171}{1.000} & \estcip{-0.118}{-0.281}{-0.019}{0.565} & \estcip{-0.130}{-0.322}{+0.009}{1.000} \\
Lupus & \estcip{-0.019}{-0.134}{+0.088}{1.000} & \estcip{+0.059}{-0.046}{+0.174}{1.000} & \estcip{-0.070}{-0.206}{+0.059}{1.000} \\
Acute MI & \estcip{+0.014}{-0.015}{+0.042}{1.000} & \estcip{-0.003}{-0.037}{+0.029}{1.000} & \estcip{+0.012}{-0.029}{+0.053}{1.000} \\
Chest X-ray Findings & \estcip{+0.009}{-0.001}{+0.018}{1.000} & \bestcip{+0.043}{+0.031}{+0.057}{0.009} & \bestcip{+0.036}{+0.023}{+0.048}{0.009} \\
\bottomrule
\end{tabular}
\end{table}

\clearpage
\begin{table}[h]
    \caption{\textbf{Performance for All Examples on EHRSHOT Across Different Count-based Models.} Mean \acf{auroc} performance with approximate 95\% confidence intervals for the count-based model from \cite{wornow_ehrshot_2023} using ontology expansion. We tested a \ac{gbm} and LR model and  extensions including the encoding of string values (SV), numeric values (NV), time binning with four time bins (TB), and the extension with all three.}
    \label{tab:ehrshot_count_models}
    \centering
    \footnotesize
    \setlength{\tabcolsep}{2.6pt} 
    \begin{tabular}{>{\raggedright\arraybackslash}p{3.05cm} 
                >{\raggedright\arraybackslash}p{1.8cm} 
                >{\raggedright\arraybackslash}p{1.8cm} 
                >{\raggedright\arraybackslash}p{1.8cm} 
                >{\raggedright\arraybackslash}p{1.8cm} 
                >{\raggedright\arraybackslash}p{1.8cm}@{}}
    \toprule
\textbf{Model}                           & \textbf{Operational Outcomes} & \textbf{Anticipating Lab Test Results} & \textbf{Assignment of New Diagnosis} & \textbf{Anticipating Chest X-ray Findings} & \textbf{Macro Avg.   Across Task Groups} \\ \midrule
\multicolumn{6}{l}{\textbf{GBM Model}} \\ \midrule
    Counts \cite{wornow_ehrshot_2023} & $\ci{0.774}{.752}{.797}$ & $\ci{0.728}{.716}{.741}$ & $\ci{0.719}{.669}{.768}$ & $\ci{0.656}{.641}{.671}$ & $\ci{0.719}{.691}{.748}$ \\
    Counts + SV                  & $\ci{0.785}{.764}{.806}$ & $\ci{0.731}{.719}{.744}$ & $\ci{0.732}{.685}{.779}$ & $\ci{0.662}{.649}{.675}$ & $\ci{0.727}{.700}{.755}$ \\
    Counts + NV                  & $\ci{0.786}{.763}{.809}$ & $\ci{0.789}{.779}{.800}$ & $\ci{0.752}{.711}{.792}$ & $\ci{0.656}{.642}{.670}$ & $\ci{0.746}{.721}{.771}$ \\ 
    Counts + TB                  & $\ci{0.815}{.795}{.836}$ & $\ci{0.752}{.741}{.764}$ & $\ci{0.768}{.736}{.801}$ & $\ci{0.686}{.674}{.698}$ & $\ci{0.756}{.735}{.776}$ \\
    Counts + SV/NV/TB            & $\ci{0.824}{.804}{.844}$ & $\ci{0.841}{.833}{.849}$ & $\ci{0.758}{.724}{.793}$ & $\ci{0.686}{.674}{.699}$ & $\ci{0.777}{.756}{.799}$ \\ \midrule
\multicolumn{6}{l}{\textbf{LR Model}} \\ \midrule
    Counts \cite{wornow_ehrshot_2023} & $\ci{0.719}{.692}{.746}$ & $\ci{0.669}{.653}{.685}$ & $\ci{0.749}{.711}{.787}$ & $\ci{0.646}{.633}{.660}$ & $\ci{0.696}{.670}{.721}$ \\
    Counts + SV                  & $\ci{0.720}{.693}{.747}$ & $\ci{0.669}{.654}{.685}$ & $\ci{0.750}{.712}{.788}$ & $\ci{0.647}{.633}{.661}$ & $\ci{0.696}{.671}{.722}$ \\
    Counts + NV                  & $\ci{0.718}{.691}{.746}$ & $\ci{0.687}{.671}{.703}$ & $\ci{0.752}{.715}{.789}$ & $\ci{0.655}{.641}{.669}$ & $\ci{0.703}{.678}{.728}$ \\
    Counts + TB                  & $\ci{0.763}{.741}{.785}$ & $\ci{0.717}{.703}{.730}$ & $\ci{0.741}{.695}{.787}$ & $\ci{0.666}{.650}{.683}$ & $\ci{0.722}{.694}{.749}$ \\
    Counts + SV/NV/TB            & $\ci{0.764}{.741}{.787}$ & $\ci{0.742}{.729}{.756}$ & $\ci{0.734}{.687}{.782}$ & $\ci{0.673}{.656}{.690}$ & $\ci{0.728}{.700}{.757}$ \\
    \bottomrule
    \end{tabular}
\end{table}

\begin{table}[]
    \caption{\textbf{Performance for All Examples on EHRSHOT for Different Serializations.} 
    Mean \acf{auroc} performance with approximate 95\% confidence intervals of the list serialization used in this work and three alternatives using the first occurrences of each code and adding timestamps to each code. We also tested a handcrafted Markdown EHR serialization with three LLMs and JSON, XML, and YAML data formats with Qwen3-Emb-8B.}
    \label{tab:ehrshot_performance_formats_full}
    \centering
    \footnotesize
    \setlength{\tabcolsep}{2.6pt} 
    \begin{tabular}{>{\raggedright\arraybackslash}p{3.05cm} 
                >{\raggedright\arraybackslash}p{1.8cm} 
                >{\raggedright\arraybackslash}p{1.8cm} 
                >{\raggedright\arraybackslash}p{1.8cm} 
                >{\raggedright\arraybackslash}p{1.8cm} 
                >{\raggedright\arraybackslash}p{1.8cm}@{}}
    \toprule
\textbf{Model}                           & \textbf{Operational Outcomes} & \textbf{Anticipating Lab Test Results} & \textbf{Assignment of New Diagnosis} & \textbf{Anticipating Chest X-ray Findings} & \textbf{Macro Avg. Across Task Groups} \\ \midrule
\multicolumn{6}{l}{\textbf{EHR List Serializations for Qwen3-Emb-8B}} \\ \midrule
List codes recent (ours)                              & $\ci{0.797}{.773}{.820}$ & $\ci{0.842}{.835}{.850}$ & $\ci{0.714}{.672}{.757}$ & $\ci{0.722}{.711}{.733}$ & $\ci{0.769}{.744}{.794}$ \\
List codes first                          & $\ci{0.761}{.736}{.785}$ & $\ci{0.715}{.701}{.728}$ & $\ci{0.731}{.688}{.773}$ & $\ci{0.676}{.663}{.690}$ & $\ci{0.721}{.694}{.747}$ \\
List codes recent + time                       & $\ci{0.795}{.772}{.818}$ & $\ci{0.844}{.837}{.851}$ & $\ci{0.692}{.637}{.748}$ & $\ci{0.727}{.716}{.738}$ & $\ci{0.765}{.734}{.795}$ \\
List codes first + time                   & $\ci{0.746}{.720}{.772}$ & $\ci{0.703}{.689}{.716}$ & $\ci{0.718}{.674}{.761}$ & $\ci{0.645}{.629}{.660}$ & $\ci{0.703}{.675}{.730}$ \\  \midrule
\multicolumn{6}{l}{\textbf{EHR Markdown Serializations}} \\ \midrule
Qwen3-Emb-8B                             & $\ci{0.773}{.749}{.797}$ & $\ci{0.859}{.852}{.866}$ & $\ci{0.725}{.683}{.767}$ & $\ci{0.694}{.681}{.707}$ & $\ci{0.763}{.737}{.788}$ \\
Qwen2-Emb-7B                             & $\ci{0.756}{.731}{.781}$ & $\ci{0.767}{.756}{.778}$ & $\ci{0.717}{.671}{.764}$ & $\ci{0.677}{.664}{.690}$ & $\ci{0.729}{.702}{.757}$ \\
Llama3.1-LLM2Vec-8B                      & $\ci{0.769}{.746}{.792}$ & $\ci{0.732}{.720}{.744}$ & $\ci{0.705}{.651}{.759}$ & $\ci{0.692}{.679}{.705}$ & $\ci{0.724}{.694}{.755}$ \\ \midrule
\multicolumn{6}{l}{\textbf{EHR Alternative Serialization Formats for Qwen3-Emb-8B}} \\ \midrule
JSON                                        & $\ci{0.773}{.749}{.796}$ & $\ci{0.858}{.851}{.865}$ & $\ci{0.736}{.692}{.780}$ & $\ci{0.690}{.677}{.704}$ & $\ci{0.764}{.738}{.790}$ \\
XML                                         & $\ci{0.771}{.747}{.795}$ & $\ci{0.862}{.855}{.868}$ & $\ci{0.726}{.681}{.771}$ & $\ci{0.676}{.663}{.690}$ & $\ci{0.759}{.732}{.785}$ \\
YAML                                        & $\ci{0.773}{.749}{.796}$ & $\ci{0.863}{.856}{.870}$ & $\ci{0.723}{.677}{.769}$ & $\ci{0.684}{.670}{.698}$ & $\ci{0.761}{.734}{.787}$ \\ 
\bottomrule
\end{tabular}
\end{table}

\clearpage
\begin{table}[]
    \caption{\textbf{Code Categories Used for Ablation Study.} For the content ablation study, we grouped all codes into six mutually exclusive categories based on ontology prefixes and hierarchical SNOMED parent concepts. Counts were computed across all events from all patients, excluding codes from CARE\textunderscore{}SITE and ICDO3, as well as codes with empty descriptions.}
    \label{tab:code_categories_ablation}
    \centering
    \setlength{\tabcolsep}{4pt}
    \footnotesize
    \begin{tabularx}{\textwidth}{@{}p{1.8cm} p{7.8cm} p{2.6cm}@{}}
        \toprule
        \textbf{Category} & \textbf{Included Codes} & \textbf{\# Codes} \\
        \midrule
        Demographics &
        Race/, Gender/, Ethnicity/, SNOMED/3950001 (birth) &
        \np{18189} (0.0\%)
        \\
        Visits &
        Visit/, Medicare Specialty/, CMS Place of Service/ &
        \np{606798} (1.5\%)
        \\
        Conditions &
        Cancer Modifier/, OMOP Extension/, Condition Type/, all remaining SNOMED codes &
        \np{2995518} (7.3\%)
        \\
        Medications &
        RxNorm/, RxNorm Extension/, CVX/, SNOMED descendants of 373873005 (pharmaceutical product), 105590001 (substance) &
        \np{2453197} (5.9\%)
        \\
        Procedures &
        CPT4/, ICD10PCS/, ICD9Proc/, Domain/, HCPCS/, SNOMED descendants of 71388002 (procedure) &
        \np{1840172} (4.5\%)
        \\
        Lab Results &
        LOINC/, SNOMED descendants of 108252007 (laboratory procedure), 430925007 (measurement of substance) &
        \np{33351150} (80.8\%)
        \\
        \midrule
        \textbf{Total} & * & \np{41265024} (100.0\%) \\
        \bottomrule
    \end{tabularx}
\end{table}

\begin{table}[]
    \caption{\textbf{Performance for All Examples on EHRSHOT for EHR Serialization Experiments.} Mean \acf{auroc} performance with approximate 95\% confidence intervals of the EHR list serialization used in this work (Full EHR) and different EHR serialization variants. We evaluated a generic and an empty instruction, serializations with specific components removed, and serialization consisting of an individual component.}
    \label{tab:ehrshot_ehr_serialization_ablation_experiments_for_llm_embeddings_models}
    \centering
    \footnotesize
    \setlength{\tabcolsep}{2.6pt} 
    \begin{tabular}{>{\raggedright\arraybackslash}p{3.05cm} 
                >{\raggedright\arraybackslash}p{1.8cm} 
                >{\raggedright\arraybackslash}p{1.8cm} 
                >{\raggedright\arraybackslash}p{1.8cm} 
                >{\raggedright\arraybackslash}p{1.8cm} 
                >{\raggedright\arraybackslash}p{1.8cm}@{}}
    \toprule
\textbf{Model}                           & \textbf{Operational Outcomes} & \textbf{Anticipating Lab Test Results} & \textbf{Assignment of New Diagnosis} & \textbf{Anticipating Chest X-ray Findings} & \textbf{Macro Avg.   Across Task Groups} \\ \midrule
\multicolumn{6}{l}{\textbf{Original List Serialization}} \\ \midrule
Full EHR & $\ci{0.797}{.773}{.820}$ & $\ci{0.842}{.835}{.850}$ & $\ci{0.714}{.672}{.757}$ & $\ci{0.722}{.711}{.733}$ & $\ci{0.769}{.744}{.794}$ \\ \midrule
\multicolumn{6}{l}{\textbf{Instruction Experiments}} \\ \midrule
Generic Instruction & $\ci{0.788}{.765}{.812}$ & $\ci{0.772}{.762}{.782}$ & $\ci{0.706}{.665}{.747}$ & $\ci{0.718}{.707}{.730}$ & $\ci{0.746}{.722}{.771}$ \\
Empty Instruction & $\ci{0.791}{.767}{.815}$ & $\ci{0.756}{.745}{.766}$ & $\ci{0.698}{.651}{.745}$ & $\ci{0.715}{.704}{.726}$ & $\ci{0.740}{.712}{.767}$ \\ \midrule
\multicolumn{6}{l}{\textbf{Removing Medical Code Categories}} \\ \midrule
No Demographics & $\ci{0.797}{.774}{.820}$ & $\ci{0.842}{.834}{.849}$ & $\ci{0.708}{.666}{.750}$ & $\ci{0.721}{.710}{.732}$ & $\ci{0.767}{.742}{.792}$ \\
No Visits & $\ci{0.796}{.773}{.819}$ & $\ci{0.842}{.835}{.849}$ & $\ci{0.717}{.676}{.758}$ & $\ci{0.723}{.712}{.734}$ & $\ci{0.769}{.745}{.794}$ \\
No Conditions & $\ci{0.797}{.774}{.820}$ & $\ci{0.847}{.840}{.854}$ & $\ci{0.701}{.665}{.737}$ & $\ci{0.716}{.705}{.727}$ & $\ci{0.765}{.743}{.788}$ \\
No Medications & $\ci{0.798}{.775}{.822}$ & $\ci{0.846}{.838}{.853}$ & $\ci{0.702}{.657}{.746}$ & $\ci{0.722}{.711}{.733}$ & $\ci{0.767}{.741}{.793}$ \\
No Procedures & $\ci{0.787}{.764}{.810}$ & $\ci{0.847}{.839}{.854}$ & $\ci{0.721}{.678}{.764}$ & $\ci{0.718}{.707}{.729}$ & $\ci{0.768}{.743}{.793}$ \\ 
No Lab Results & $\ci{0.798}{.775}{.821}$ & $\ci{0.752}{.741}{.763}$ & $\ci{0.722}{.681}{.762}$ & $\ci{0.711}{.699}{.722}$ & $\ci{0.746}{.721}{.770}$ \\ \midrule
\multicolumn{6}{l}{\textbf{Individual Medical Code Categories}} \\ \midrule
Only Demographics & $\ci{0.549}{.518}{.580}$ & $\ci{0.543}{.528}{.559}$ & $\ci{0.606}{.570}{.643}$ & $\ci{0.513}{.494}{.531}$ & $\ci{0.553}{.526}{.579}$ \\
Only Visits & $\ci{0.591}{.564}{.618}$ & $\ci{0.628}{.615}{.642}$ & $\ci{0.560}{.523}{.596}$ & $\ci{0.613}{.600}{.625}$ & $\ci{0.598}{.574}{.622}$ \\
Only Conditions & $\ci{0.773}{.750}{.797}$ & $\ci{0.708}{.695}{.721}$ & $\ci{0.693}{.649}{.737}$ & $\ci{0.686}{.673}{.698}$ & $\ci{0.715}{.688}{.741}$ \\
Only Medications & $\ci{0.777}{.756}{.799}$ & $\ci{0.710}{.698}{.722}$ & $\ci{0.688}{.641}{.735}$ & $\ci{0.657}{.644}{.670}$ & $\ci{0.708}{.681}{.735}$ \\
Only Procedures & $\ci{0.781}{.756}{.807}$ & $\ci{0.718}{.705}{.731}$ & $\ci{0.672}{.640}{.704}$ & $\ci{0.697}{.685}{.708}$ & $\ci{0.717}{.695}{.739}$ \\
Only Lab Results & $\ci{0.777}{.755}{.799}$ & $\ci{0.856}{.849}{.862}$ & $\ci{0.665}{.607}{.724}$ & $\ci{0.702}{.691}{.713}$ & $\ci{0.750}{.718}{.782}$ \\
    \bottomrule
    \end{tabular}
\end{table}

\clearpage
\begin{table}[h]
    \caption{\textbf{Performance for All Examples on EHRSHOT Across Context Sizes.} Mean \acf{auroc} performance with approximate 95\% confidence intervals for different context sizes of the LLM embedding models.}
    \label{tab:ehrshot_performance_of_llm_embedding_models_across_context_sizes}
    \centering
    \footnotesize
    \setlength{\tabcolsep}{2.6pt} 
    \begin{tabular}{>{\raggedright\arraybackslash}p{3.05cm} 
                >{\raggedright\arraybackslash}p{1.8cm} 
                >{\raggedright\arraybackslash}p{1.8cm} 
                >{\raggedright\arraybackslash}p{1.8cm} 
                >{\raggedright\arraybackslash}p{1.8cm} 
                >{\raggedright\arraybackslash}p{1.8cm}@{}}
    \toprule
\textbf{Model}                           & \textbf{Operational Outcomes} & \textbf{Anticipating Lab Test Results} & \textbf{Assignment of New Diagnosis} & \textbf{Anticipating Chest X-ray Findings} & \textbf{Macro Avg.   Across Task Groups} \\ \midrule
\multicolumn{6}{l}{\textbf{Qwen3-Emb-8B}} \\ \midrule
    8,192 context size               & $\ci{0.797}{.773}{.820}$ & $\ci{0.842}{.835}{.850}$ & $\ci{0.714}{.672}{.757}$ & $\ci{0.722}{.711}{.733}$ & $\ci{0.769}{.744}{.794}$ \\
    4,096 context size               & $\ci{0.800}{.778}{.823}$ & $\ci{0.850}{.843}{.857}$ & $\ci{0.711}{.666}{.757}$ & $\ci{0.727}{.716}{.737}$ & $\ci{0.772}{.746}{.798}$ \\
    2,048 context size               & $\ci{0.805}{.783}{.828}$ & $\ci{0.859}{.852}{.865}$ & $\ci{0.691}{.656}{.726}$ & $\ci{0.718}{.706}{.729}$ & $\ci{0.768}{.746}{.790}$ \\
    1,024 context size               & $\ci{0.799}{.776}{.822}$ & $\ci{0.832}{.825}{.839}$ & $\ci{0.707}{.668}{.746}$ & $\ci{0.697}{.686}{.709}$ & $\ci{0.759}{.735}{.782}$ \\
    512 context size                 & $\ci{0.790}{.766}{.814}$ & $\ci{0.786}{.778}{.794}$ & $\ci{0.703}{.660}{.745}$ & $\ci{0.683}{.672}{.694}$ & $\ci{0.741}{.715}{.766}$ \\ \midrule
\multicolumn{6}{l}{\textbf{Qwen2-Emb-7B}} \\ \midrule
    8,192 context size               & $\ci{0.772}{.749}{.796}$ & $\ci{0.746}{.735}{.757}$ & $\ci{0.744}{.705}{.784}$ & $\ci{0.685}{.671}{.699}$ & $\ci{0.737}{.712}{.761}$    \\
    4,096 context size         & $\ci{0.794}{.771}{.816}$ & $\ci{0.799}{.790}{.808}$ & $\ci{0.740}{.698}{.781}$ & $\ci{0.708}{.696}{.720}$ & $\ci{0.760}{.736}{.785}$ \\
    2,048 context size               & $\ci{0.806}{.783}{.828}$ & $\ci{0.846}{.839}{.853}$ & $\ci{0.718}{.674}{.761}$ & $\ci{0.720}{.709}{.731}$ & $\ci{0.772}{.747}{.798}$ \\
    1,024 context size               & $\ci{0.797}{.776}{.817}$ & $\ci{0.833}{.826}{.841}$ & $\ci{0.709}{.665}{.752}$ & $\ci{0.700}{.689}{.711}$ & $\ci{0.760}{.735}{.785}$ \\
    512 context size                 & $\ci{0.784}{.763}{.805}$ & $\ci{0.786}{.778}{.794}$ & $\ci{0.741}{.702}{.779}$ & $\ci{0.683}{.672}{.694}$ & $\ci{0.748}{.726}{.771}$ \\ \midrule
\multicolumn{6}{l}{\textbf{Llama3.1-LLM2Vec-8B}} \\ \midrule
    8,192 context size               & $\ci{0.763}{.738}{.787}$ & $\ci{0.726}{.714}{.738}$ & $\ci{0.727}{.688}{.766}$ & $\ci{0.686}{.673}{.699}$ & $\ci{0.725}{.701}{.750}$ \\
    4,096 context size               & $\ci{0.780}{.756}{.804}$ & $\ci{0.757}{.746}{.768}$ & $\ci{0.714}{.664}{.763}$ & $\ci{0.716}{.705}{.727}$ & $\ci{0.742}{.713}{.770}$ \\
    2,048 context size               & $\ci{0.803}{.781}{.825}$ & $\ci{0.826}{.819}{.834}$ & $\ci{0.680}{.636}{.725}$ & $\ci{0.721}{.710}{.731}$ & $\ci{0.757}{.732}{.783}$ \\
    1,024 context size               & $\ci{0.790}{.766}{.813}$ & $\ci{0.843}{.836}{.850}$ & $\ci{0.696}{.654}{.738}$ & $\ci{0.700}{.689}{.710}$ & $\ci{0.757}{.732}{.782}$ \\
    512 context size                 & $\ci{0.775}{.751}{.800}$ & $\ci{0.789}{.781}{.796}$ & $\ci{0.689}{.643}{.735}$ & $\ci{0.686}{.676}{.697}$ & $\ci{0.735}{.708}{.762}$ \\
    \bottomrule
    \end{tabular}
\end{table}

\clearpage
\begin{table}[h]
    \caption{\textbf{Performance for All Examples on EHRSHOT Across Time Windows.} Mean \acf{auroc} performance with approximate 95\% confidence intervals for different time windows of the LLM embedding models and the count-based baseline.}
    \label{tab:ehrshot_performance_of_llm_embedding_models_across_time_windows}
    \centering
    \footnotesize
    \setlength{\tabcolsep}{2.6pt} 
    \begin{tabular}{>{\raggedright\arraybackslash}p{3.05cm} 
                >{\raggedright\arraybackslash}p{1.8cm} 
                >{\raggedright\arraybackslash}p{1.8cm} 
                >{\raggedright\arraybackslash}p{1.8cm} 
                >{\raggedright\arraybackslash}p{1.8cm} 
                >{\raggedright\arraybackslash}p{1.8cm}@{}}
    \toprule
\textbf{Model}                           & \textbf{Operational Outcomes} & \textbf{Anticipating Lab Test Results} & \textbf{Assignment of New Diagnosis} & \textbf{Anticipating Chest X-ray Findings} & \textbf{Macro Avg.   Across Task Groups} \\ \midrule
\multicolumn{6}{l}{\textbf{Qwen3-Emb-8B}} \\ \midrule
    Full patient history         & $\ci{0.797}{.773}{.820}$ & $\ci{0.842}{.835}{.850}$ & $\ci{0.714}{.672}{.757}$ & $\ci{0.722}{.711}{.733}$ & $\ci{0.769}{.744}{.794}$ \\
    3 years (1,095 days) & $\ci{0.798}{.775}{.821}$ & $\ci{0.842}{.835}{.849}$ & $\ci{0.717}{.674}{.759}$ & $\ci{0.723}{.712}{.734}$ & $\ci{0.770}{.745}{.795}$ \\
    1 year (365 days)   & $\ci{0.802}{.780}{.824}$ & $\ci{0.843}{.836}{.850}$ & $\ci{0.720}{.677}{.762}$ & $\ci{0.723}{.712}{.734}$ & $\ci{0.772}{.747}{.797}$ \\
    1 month (30 days)    & $\ci{0.814}{.794}{.835}$ & $\ci{0.840}{.832}{.847}$ & $\ci{0.695}{.651}{.739}$ & $\ci{0.717}{.707}{.727}$ & $\ci{0.767}{.742}{.792}$ \\
    1 week (7 days)      & $\ci{0.813}{.792}{.834}$ & $\ci{0.832}{.824}{.839}$ & $\ci{0.683}{.645}{.722}$ & $\ci{0.702}{.692}{.712}$ & $\ci{0.758}{.735}{.780}$ \\
    1 day                & $\ci{0.812}{.790}{.834}$ & $\ci{0.800}{.791}{.808}$ & $\ci{0.697}{.660}{.735}$ & $\ci{0.664}{.654}{.675}$ & $\ci{0.743}{.721}{.766}$ \\
    1 hour               & $\ci{0.764}{.740}{.788}$ & $\ci{0.677}{.668}{.686}$ & $\ci{0.627}{.585}{.668}$ & $\ci{0.602}{.591}{.612}$ & $\ci{0.667}{.642}{.692}$ \\ \midrule
    \multicolumn{6}{l}{\textbf{Qwen2-Emb-7B}} \\ \midrule
    Full patient history         & $\ci{0.772}{.749}{.796}$ & $\ci{0.746}{.735}{.757}$ & $\ci{0.744}{.705}{.784}$ & $\ci{0.685}{.671}{.699}$ & $\ci{0.737}{.712}{.761}$ \\
    3 years (1,095 days) & $\ci{0.773}{.750}{.796}$ & $\ci{0.754}{.743}{.765}$ & $\ci{0.725}{.681}{.770}$ & $\ci{0.688}{.675}{.701}$ & $\ci{0.735}{.709}{.761}$ \\
    1 year (365 days)    & $\ci{0.781}{.758}{.803}$ & $\ci{0.766}{.756}{.776}$ & $\ci{0.724}{.685}{.763}$ & $\ci{0.696}{.683}{.709}$ & $\ci{0.742}{.718}{.766}$ \\
    1 month (30 days)    & $\ci{0.810}{.789}{.831}$ & $\ci{0.789}{.780}{.798}$ & $\ci{0.715}{.679}{.752}$ & $\ci{0.711}{.701}{.722}$ & $\ci{0.756}{.734}{.779}$ \\
    1 week (7 days)      & $\ci{0.812}{.791}{.833}$ & $\ci{0.801}{.792}{.809}$ & $\ci{0.700}{.658}{.741}$ & $\ci{0.702}{.692}{.712}$ & $\ci{0.754}{.729}{.778}$ \\
    1 day                & $\ci{0.817}{.797}{.837}$ & $\ci{0.796}{.788}{.804}$ & $\ci{0.720}{.669}{.771}$ & $\ci{0.663}{.653}{.673}$ & $\ci{0.749}{.721}{.777}$ \\
    1 hour               & $\ci{0.771}{.749}{.794}$ & $\ci{0.674}{.665}{.682}$ & $\ci{0.599}{.558}{.640}$ & $\ci{0.600}{.589}{.610}$ & $\ci{0.661}{.637}{.685}$ \\ \midrule
    \multicolumn{6}{l}{\textbf{Llama3.1-LLM2Vec-8B}} \\ \midrule
    Full patient history & $\ci{0.763}{.738}{.787}$ & $\ci{0.726}{.714}{.738}$ & $\ci{0.727}{.688}{.766}$ & $\ci{0.686}{.673}{.699}$ & $\ci{0.725}{.701}{.750}$ \\
    3 years (1,095 days) & $\ci{0.763}{.739}{.787}$ & $\ci{0.733}{.721}{.744}$ & $\ci{0.696}{.662}{.730}$ & $\ci{0.694}{.682}{.705}$ & $\ci{0.721}{.699}{.744}$ \\
    1 year (365 days)    & $\ci{0.776}{.753}{.800}$ & $\ci{0.742}{.731}{.754}$ & $\ci{0.706}{.657}{.756}$ & $\ci{0.705}{.694}{.717}$ & $\ci{0.733}{.704}{.761}$ \\
    1 month (30 days)    & $\ci{0.802}{.780}{.824}$ & $\ci{0.770}{.760}{.780}$ & $\ci{0.696}{.642}{.751}$ & $\ci{0.718}{.707}{.728}$ & $\ci{0.747}{.716}{.777}$ \\
    1 week (7 days)      & $\ci{0.816}{.795}{.836}$ & $\ci{0.782}{.773}{.791}$ & $\ci{0.687}{.634}{.740}$ & $\ci{0.702}{.691}{.712}$ & $\ci{0.747}{.718}{.776}$ \\
    1 day                & $\ci{0.811}{.790}{.833}$ & $\ci{0.786}{.777}{.794}$ & $\ci{0.680}{.626}{.734}$ & $\ci{0.666}{.656}{.676}$ & $\ci{0.736}{.706}{.766}$ \\ 
     1 hour              & $\ci{0.766}{.743}{.790}$ & $\ci{0.673}{.664}{.681}$ & $\ci{0.581}{.538}{.625}$ & $\ci{0.595}{.585}{.606}$ & $\ci{0.654}{.628}{.680}$ \\  \midrule 
    \multicolumn{6}{l}{\textbf{Count-based + GBM}} \\ \midrule
     Full patient history & $\ci{0.824}{.804}{.844}$ & $\ci{0.841}{.833}{.849}$ & $\ci{0.758}{.724}{.793}$ & $\ci{0.686}{.674}{.699}$ & $\ci{0.777}{.756}{.799}$ \\
     3 years (1,095 days) & $\ci{0.819}{.799}{.839}$ & $\ci{0.841}{.833}{.850}$ & $\ci{0.740}{.691}{.790}$ & $\ci{0.688}{.676}{.700}$ & $\ci{0.772}{.745}{.800}$ \\
     1 year (365 days)    & $\ci{0.824}{.803}{.844}$ & $\ci{0.841}{.833}{.849}$ & $\ci{0.749}{.705}{.794}$ & $\ci{0.679}{.666}{.692}$ & $\ci{0.773}{.748}{.799}$ \\
    1 month (30 days)     & $\ci{0.824}{.805}{.844}$ & $\ci{0.836}{.828}{.844}$ & $\ci{0.707}{.669}{.746}$ & $\ci{0.688}{.677}{.700}$ & $\ci{0.764}{.741}{.787}$ \\
    1 week (7 days)       & $\ci{0.814}{.793}{.835}$ & $\ci{0.842}{.835}{.850}$ & $\ci{0.693}{.653}{.734}$ & $\ci{0.686}{.674}{.697}$ & $\ci{0.759}{.735}{.782}$ \\
    1 day                 & $\ci{0.820}{.799}{.841}$ & $\ci{0.830}{.823}{.837}$ & $\ci{0.702}{.660}{.745}$ & $\ci{0.650}{.638}{.662}$ & $\ci{0.751}{.726}{.775}$ \\
    1 hour                & $\ci{0.742}{.717}{.767}$ & $\ci{0.670}{.662}{.679}$ & $\ci{0.616}{.569}{.664}$ & $\ci{0.572}{.559}{.586}$ & $\ci{0.650}{.622}{.678}$ \\
    \bottomrule
    \end{tabular}
\end{table}

\clearpage
\begin{table}[h]
    \caption{\textbf{Encoding Times for Models on EHRSHOT.} Time required to encode EHR entries for EHRSHOT experiments on an 8-GPU H200 cluster. LLM embedding models and encoder language models use EHR list serialization with inputs of up to \np{8192} tokens. Encoder language models process the input in separate 512-token chunks. Differences in multi-GPU optimization across models may affect runtime. We attempted to optimize each model for the given GPU cluster.}
    \label{tab:ehrshot_encoding_time}
    \centering
    \footnotesize
    \setlength{\tabcolsep}{2.6pt} 
    \begin{tabular}{>{\raggedright\arraybackslash}p{3.1cm} 
                >{\raggedright\arraybackslash}p{2.4cm}
                >{\raggedright\arraybackslash}p{2.8cm}@{}}
    \toprule
\textbf{Model}                           & \textbf{Encoding Time} & \textbf{Macro Avg. Across Task Groups} \\ \midrule
\multicolumn{3}{l}{\textbf{Baselines} \cite{wornow_ehrshot_2023}} \\ \midrule
CLMBR-T-Base                             & 6:04 & $\ci{0.769}{.746}{.792}$ \\ \midrule
\multicolumn{3}{l}{\textbf{LLM Embedding Models}} \\ \midrule
Qwen3-Emb-8B                             & 21:48:56 & $\ci{0.769}{.744}{.794}$ \\
Qwen3-Emb-4B                             & 12:20:28 & $\ci{0.759}{.730}{.787}$ \\
Qwen3-Emb-0.6B                           & 5:14:07 & $\ci{0.727}{.697}{.758}$ \\
Qwen2-Emb-7B                             & 12:47:18 & $\ci{0.737}{.712}{.761}$ \\
Qwen2-Emb-1.5B                           & 5:23:55 & $\ci{0.710}{.683}{.738}$ \\
Llama3.1-LLM2Vec-8B                      & 21:39:06 & $\ci{0.725}{.701}{.750}$ \\ \midrule
\multicolumn{3}{l}{\textbf{Encoder Language Models}} \\ \midrule
DeBERTaV3 large                          & 7:00:21 & $\ci{0.692}{.664}{.720}$ \\
DeBERTaV3 base                           & 2:49:35 & $\ci{0.685}{.660}{.709}$ \\
BERT large                               & 5:31:18 & $\ci{0.694}{.667}{.722}$ \\
BERT base                                & 2:21:35 & $\ci{0.701}{.674}{.727}$ \\ 
\bottomrule
\end{tabular}
\end{table}

\clearpage
\begin{table}[]
    \caption{\textbf{Hyperparameter Tuning Results for Fine-Tuning Experiments.} Mean validation AUROC across the tuning sweep for the operational outcomes and assignment of new diagnoses tasks at $k \in \{8, 16\}$. Rows marked with a * denote the selected configuration for each model family.}
    \label{tab:ehrshot_tuning_hparams}
    \centering
    \footnotesize
    \setlength{\tabcolsep}{3.2pt}
    \begin{tabular}{>{\raggedright\arraybackslash}p{2.8cm}
                >{\raggedright\arraybackslash}p{1.35cm}
                >{\raggedright\arraybackslash}p{0.8cm}
                >{\raggedright\arraybackslash}p{1.1cm}
                >{\raggedright\arraybackslash}p{1.45cm}
                >{\raggedright\arraybackslash}p{1.55cm}
                >{\raggedright\arraybackslash}p{1.7cm}}
    \toprule
\textbf{Config ID} & \textbf{LR} & \textbf{$r$} & \textbf{Dropout} & \textbf{$k=8$ Val. AUROC} & \textbf{$k=16$ Val. AUROC} & \textbf{Macro Avg.} \\ \midrule
\multicolumn{7}{l}{\textbf{Encoder Models}} \\ \midrule
\texttt{enc\_lr5e5\_r8\_d000} & 5e-5 & 8 & 0.00 & 0.667 & 0.688 & 0.677 \\
\texttt{enc\_lr5e5\_r8\_d005} & 5e-5 & 8 & 0.05 & 0.641 & 0.696 & 0.668 \\
\texttt{enc\_lr5e5\_r8\_d010}* & 5e-5 & 8 & 0.10 & 0.686 & 0.720 & 0.703 \\
\texttt{enc\_lr5e5\_r16\_d000} & 5e-5 & 16 & 0.00 & 0.616 & 0.651 & 0.634 \\
\texttt{enc\_lr5e5\_r16\_d005} & 5e-5 & 16 & 0.05 & 0.606 & 0.686 & 0.646 \\
\texttt{enc\_lr5e5\_r16\_d010} & 5e-5 & 16 & 0.10 & 0.578 & 0.617 & 0.597 \\
\texttt{enc\_lr5e5\_r64\_d000} & 5e-5 & 64 & 0.00 & 0.639 & 0.696 & 0.667 \\
\texttt{enc\_lr5e5\_r64\_d005} & 5e-5 & 64 & 0.05 & 0.668 & 0.677 & 0.672 \\
\texttt{enc\_lr5e5\_r64\_d010} & 5e-5 & 64 & 0.10 & 0.658 & 0.634 & 0.646 \\
\texttt{enc\_lr1e4\_r8\_d000} & 1e-4 & 8 & 0.00 & 0.691 & 0.630 & 0.660 \\
\texttt{enc\_lr1e4\_r8\_d005} & 1e-4 & 8 & 0.05 & 0.592 & 0.657 & 0.625 \\
\texttt{enc\_lr1e4\_r8\_d010} & 1e-4 & 8 & 0.10 & 0.668 & 0.681 & 0.675 \\
\texttt{enc\_lr1e4\_r16\_d000} & 1e-4 & 16 & 0.00 & 0.615 & 0.660 & 0.637 \\
\texttt{enc\_lr1e4\_r16\_d005} & 1e-4 & 16 & 0.05 & 0.670 & 0.630 & 0.650 \\
\texttt{enc\_lr1e4\_r16\_d010} & 1e-4 & 16 & 0.10 & 0.641 & 0.653 & 0.647 \\
\texttt{enc\_lr1e4\_r64\_d000} & 1e-4 & 64 & 0.00 & 0.623 & 0.689 & 0.656 \\
\texttt{enc\_lr1e4\_r64\_d005} & 1e-4 & 64 & 0.05 & 0.584 & 0.670 & 0.627 \\
\texttt{enc\_lr1e4\_r64\_d010} & 1e-4 & 64 & 0.10 & 0.587 & 0.686 & 0.637 \\
\texttt{enc\_lr2e4\_r8\_d000} & 2e-4 & 8 & 0.00 & 0.616 & 0.698 & 0.657 \\
\texttt{enc\_lr2e4\_r8\_d005} & 2e-4 & 8 & 0.05 & 0.655 & 0.655 & 0.655 \\
\texttt{enc\_lr2e4\_r8\_d010} & 2e-4 & 8 & 0.10 & 0.616 & 0.686 & 0.651 \\
\texttt{enc\_lr2e4\_r16\_d000} & 2e-4 & 16 & 0.00 & 0.588 & 0.632 & 0.610 \\
\texttt{enc\_lr2e4\_r16\_d005} & 2e-4 & 16 & 0.05 & 0.596 & 0.703 & 0.650 \\
\texttt{enc\_lr2e4\_r16\_d010} & 2e-4 & 16 & 0.10 & 0.705 & 0.681 & 0.693 \\
\texttt{enc\_lr2e4\_r64\_d000} & 2e-4 & 64 & 0.00 & 0.670 & 0.703 & 0.687 \\
\texttt{enc\_lr2e4\_r64\_d005} & 2e-4 & 64 & 0.05 & 0.627 & 0.630 & 0.628 \\
\texttt{enc\_lr2e4\_r64\_d010} & 2e-4 & 64 & 0.10 & 0.588 & 0.663 & 0.625 \\
\midrule
\multicolumn{7}{l}{\textbf{Decoder Models}} \\ \midrule
\texttt{dec\_lr5e5\_r8\_d000} & 5e-5 & 8 & 0.00 & 0.614 & 0.606 & 0.610 \\
\texttt{dec\_lr5e5\_r8\_d005} & 5e-5 & 8 & 0.05 & 0.616 & 0.611 & 0.614 \\
\texttt{dec\_lr5e5\_r8\_d010} & 5e-5 & 8 & 0.10 & 0.624 & 0.615 & 0.620 \\
\texttt{dec\_lr5e5\_r16\_d000} & 5e-5 & 16 & 0.00 & 0.609 & 0.622 & 0.616 \\
\texttt{dec\_lr5e5\_r16\_d005} & 5e-5 & 16 & 0.05 & 0.615 & 0.606 & 0.610 \\
\texttt{dec\_lr5e5\_r16\_d010} & 5e-5 & 16 & 0.10 & 0.615 & 0.633 & 0.624 \\
\texttt{dec\_lr5e5\_r64\_d000} & 5e-5 & 64 & 0.00 & 0.609 & 0.609 & 0.609 \\
\texttt{dec\_lr5e5\_r64\_d005} & 5e-5 & 64 & 0.05 & 0.629 & 0.624 & 0.627 \\
\texttt{dec\_lr5e5\_r64\_d010} & 5e-5 & 64 & 0.10 & 0.616 & 0.625 & 0.621 \\
\texttt{dec\_lr1e4\_r8\_d000} & 1e-4 & 8 & 0.00 & 0.606 & 0.629 & 0.617 \\
\texttt{dec\_lr1e4\_r8\_d005} & 1e-4 & 8 & 0.05 & 0.619 & 0.625 & 0.622 \\
\texttt{dec\_lr1e4\_r8\_d010} & 1e-4 & 8 & 0.10 & 0.610 & 0.613 & 0.612 \\
\texttt{dec\_lr1e4\_r16\_d000} & 1e-4 & 16 & 0.00 & 0.641 & 0.619 & 0.630 \\
\texttt{dec\_lr1e4\_r16\_d005} & 1e-4 & 16 & 0.05 & 0.627 & 0.628 & 0.627 \\
\texttt{dec\_lr1e4\_r16\_d010} & 1e-4 & 16 & 0.10 & 0.621 & 0.605 & 0.613 \\
\texttt{dec\_lr1e4\_r64\_d000} & 1e-4 & 64 & 0.00 & 0.623 & 0.624 & 0.623 \\
\texttt{dec\_lr1e4\_r64\_d005} & 1e-4 & 64 & 0.05 & 0.628 & 0.615 & 0.622 \\
\texttt{dec\_lr1e4\_r64\_d010} & 1e-4 & 64 & 0.10 & 0.625 & 0.628 & 0.627 \\
\texttt{dec\_lr2e4\_r8\_d000} & 2e-4 & 8 & 0.00 & 0.632 & 0.651 & 0.642 \\
\texttt{dec\_lr2e4\_r8\_d005}* & 2e-4 & 8 & 0.05 & 0.706 & 0.637 & 0.671 \\
\texttt{dec\_lr2e4\_r8\_d010} & 2e-4 & 8 & 0.10 & 0.652 & 0.638 & 0.645 \\
\texttt{dec\_lr2e4\_r16\_d000} & 2e-4 & 16 & 0.00 & 0.657 & 0.645 & 0.651 \\
\texttt{dec\_lr2e4\_r16\_d005} & 2e-4 & 16 & 0.05 & 0.643 & 0.621 & 0.632 \\
\texttt{dec\_lr2e4\_r16\_d010} & 2e-4 & 16 & 0.10 & 0.668 & 0.632 & 0.650 \\
\texttt{dec\_lr2e4\_r64\_d000} & 2e-4 & 64 & 0.00 & 0.674 & 0.647 & 0.660 \\
\texttt{dec\_lr2e4\_r64\_d005} & 2e-4 & 64 & 0.05 & 0.651 & 0.619 & 0.635 \\
\texttt{dec\_lr2e4\_r64\_d010} & 2e-4 & 64 & 0.10 & 0.646 & 0.636 & 0.641 \\
    \bottomrule
    \end{tabular}
\end{table}

\clearpage

\subsection{Additional Performance Results on EHRSHOT}
\begin{figure}[h]
    \centering
    \includegraphics[width=\linewidth]{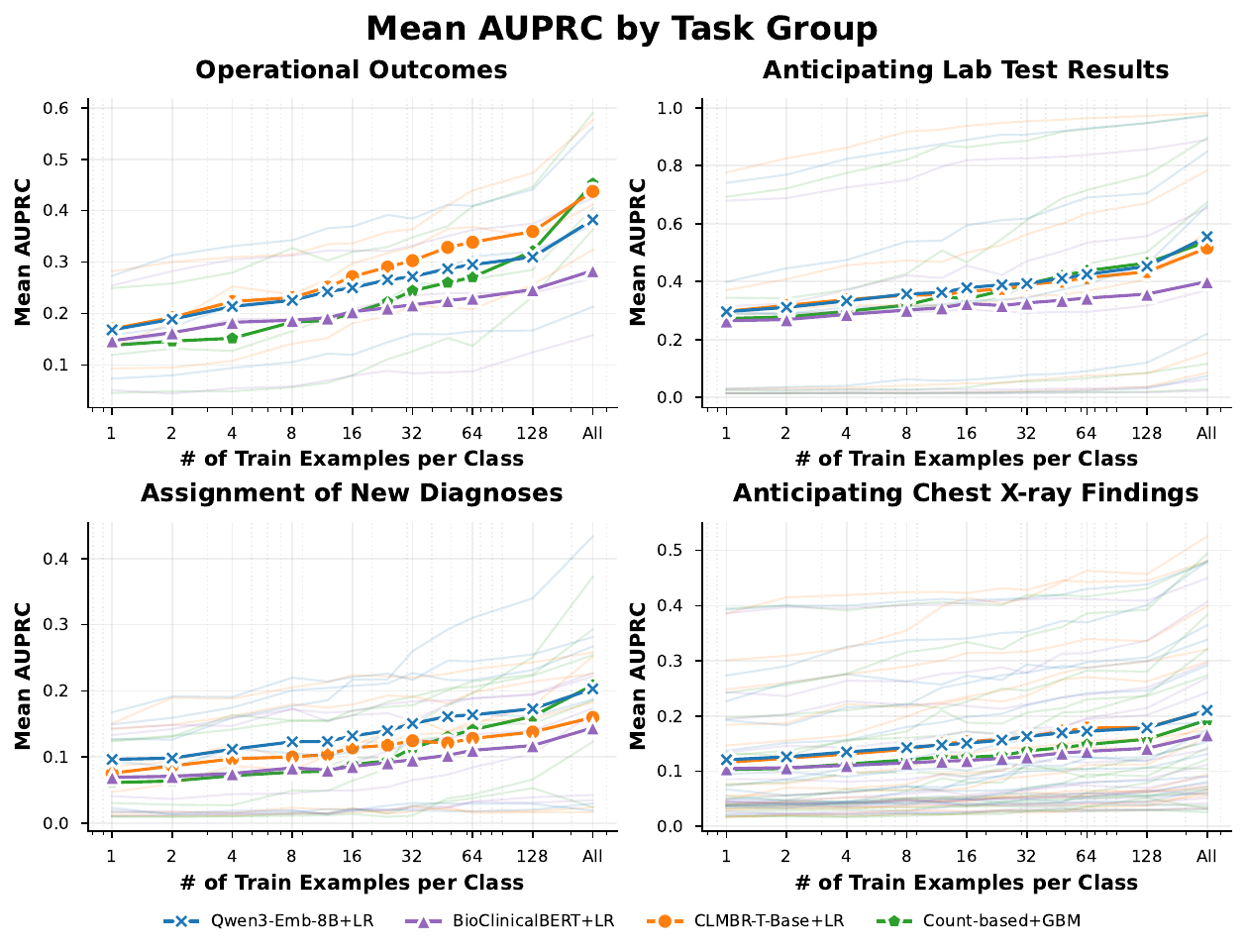}
    \caption{\textbf{Few-Shot AUPRC Performance on EHRSHOT.} Mean \acf{auprc} performance across subtasks for four task groups (bold). Blurred lines show averages across five bootstrapped runs using different seeds \cite{wornow_ehrshot_2023}.}
    \label{fig:auprc_performance_in_few_shot_settings}
\end{figure}
\begin{figure}[h]
    \centering
    \includegraphics[width=\linewidth]{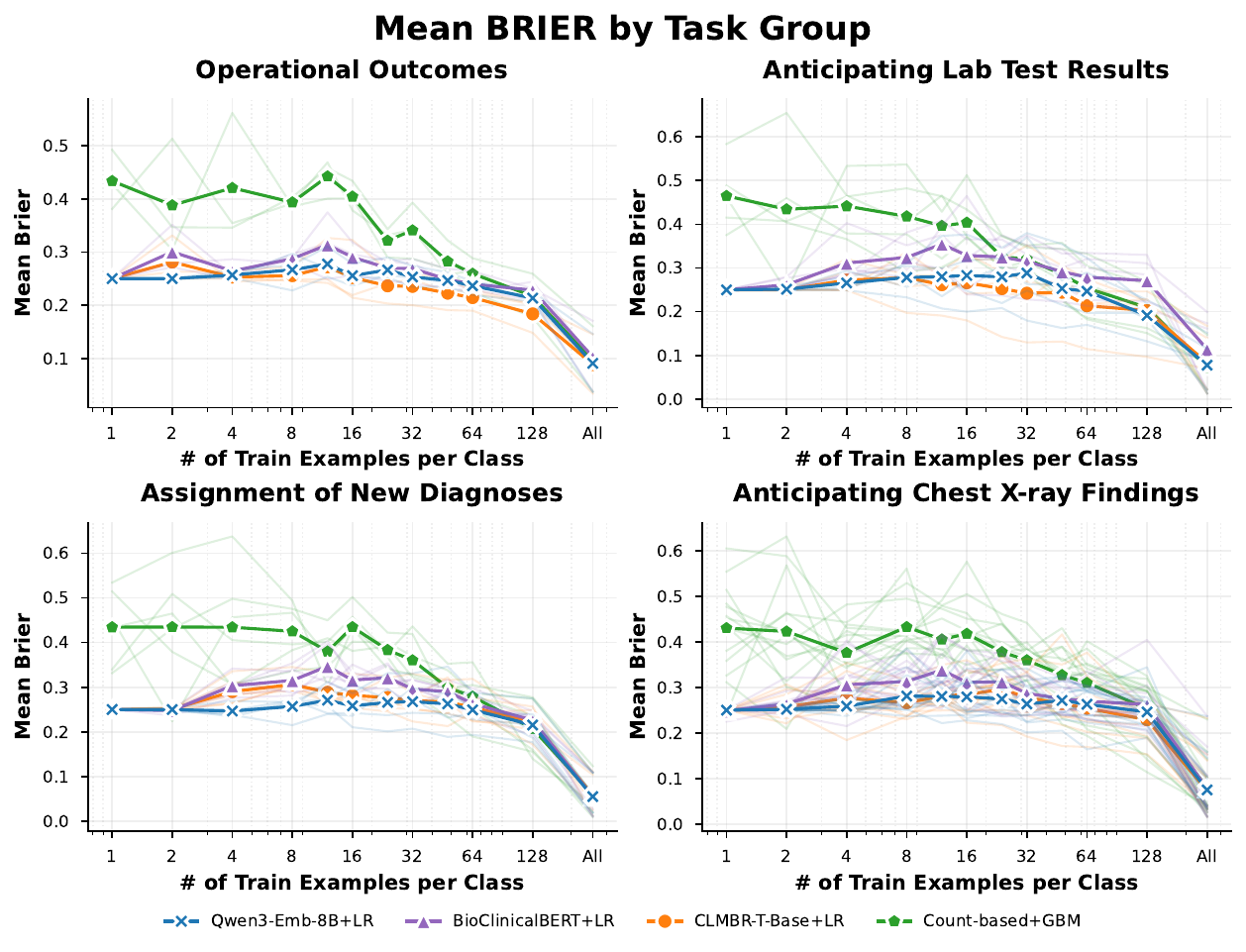}
    \caption{\textbf{Few-Shot Brier Score on EHRSHOT.} Mean Brier score across subtasks for four task groups (bold). Blurred lines show averages across five bootstrapped runs using different seeds \cite{wornow_ehrshot_2023}.}
    \label{fig:brier_performance_in_few_shot_settings}
\end{figure}
\clearpage
\begin{figure}[h]
    \centering    
    \includegraphics[width=0.9\linewidth]{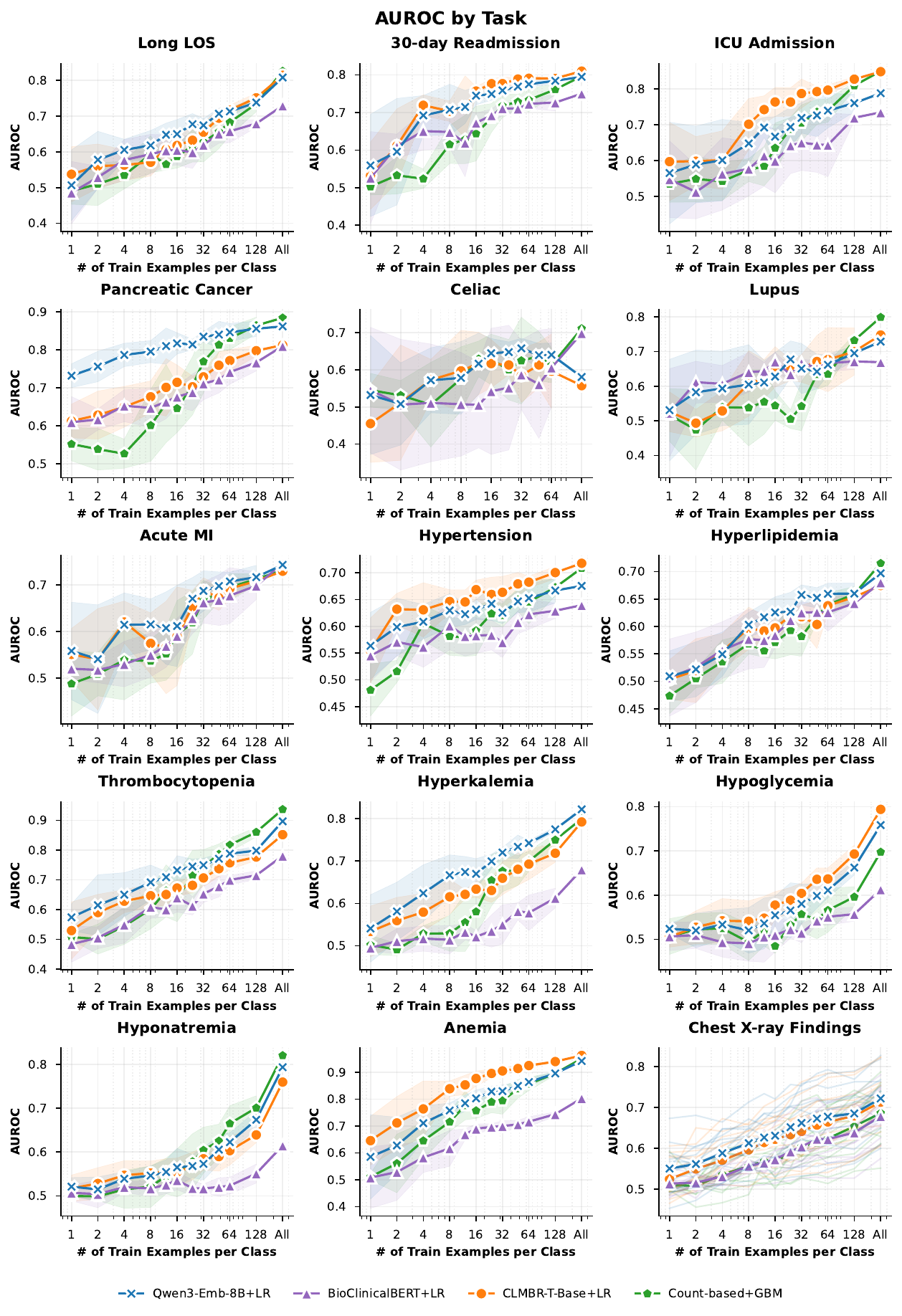}
    \caption{\textbf{Task-Specific AUROC Performance on EHRSHOT.} \Acf{auroc} performance with standard deviation across five few-shot replicates for all \np{15} prediction tasks \cite{wornow_ehrshot_2023}.}
    \label{fig:task_specific_auroc_performance}
\end{figure}
\clearpage
\begin{figure}[h]
    \centering
    \includegraphics[width=0.9\linewidth]{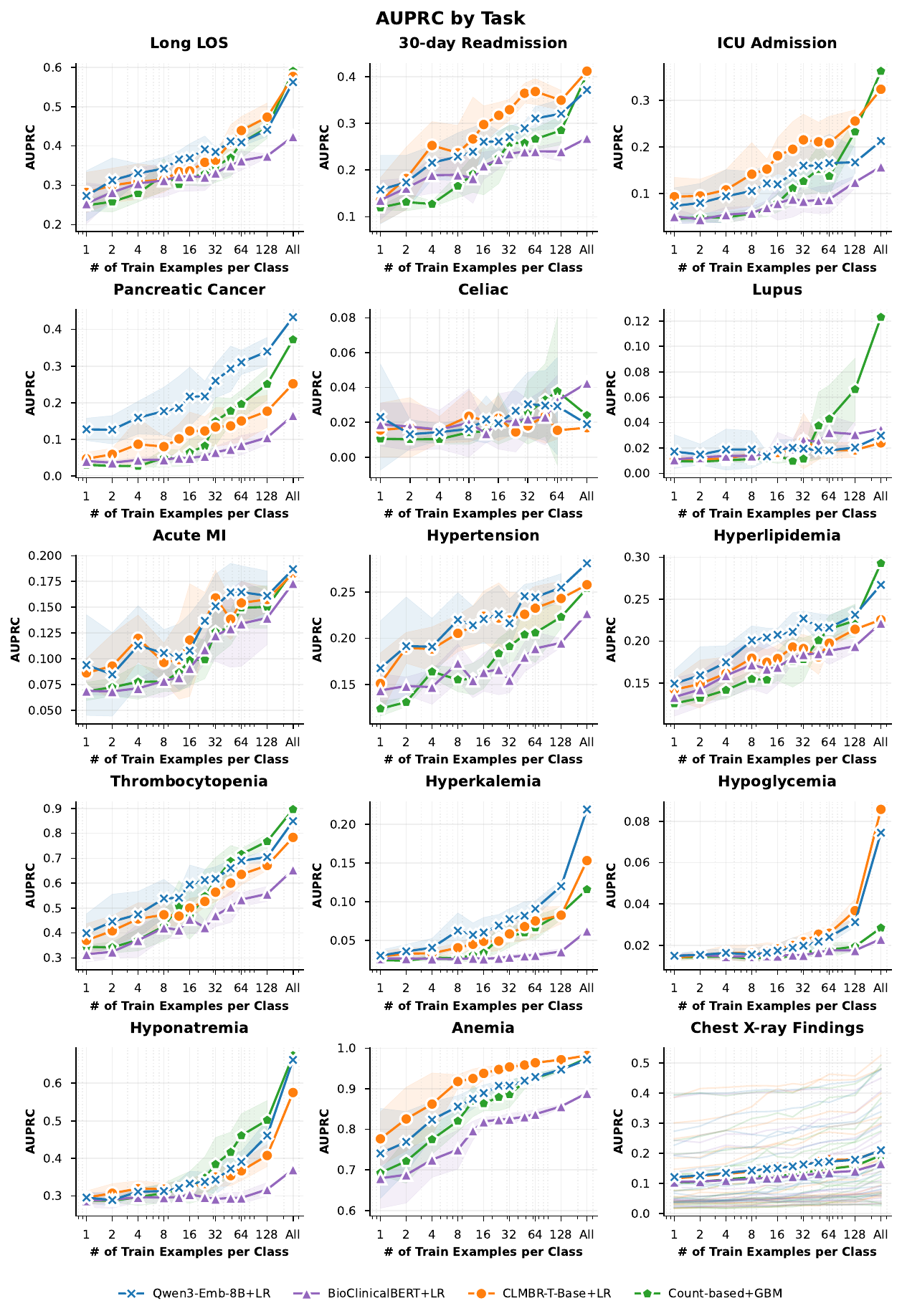}
    \caption{\textbf{Task-Specific AUPRC Performance on EHRSHOT.} \Acf{auprc} performance with standard deviation across five few-shot replicates for all \np{15} prediction tasks \cite{wornow_ehrshot_2023}.}
    \label{fig:task_specific_auprc_performance}
\end{figure}
\clearpage
\begin{figure}[h]
    \centering
    \includegraphics[width=0.9\linewidth]{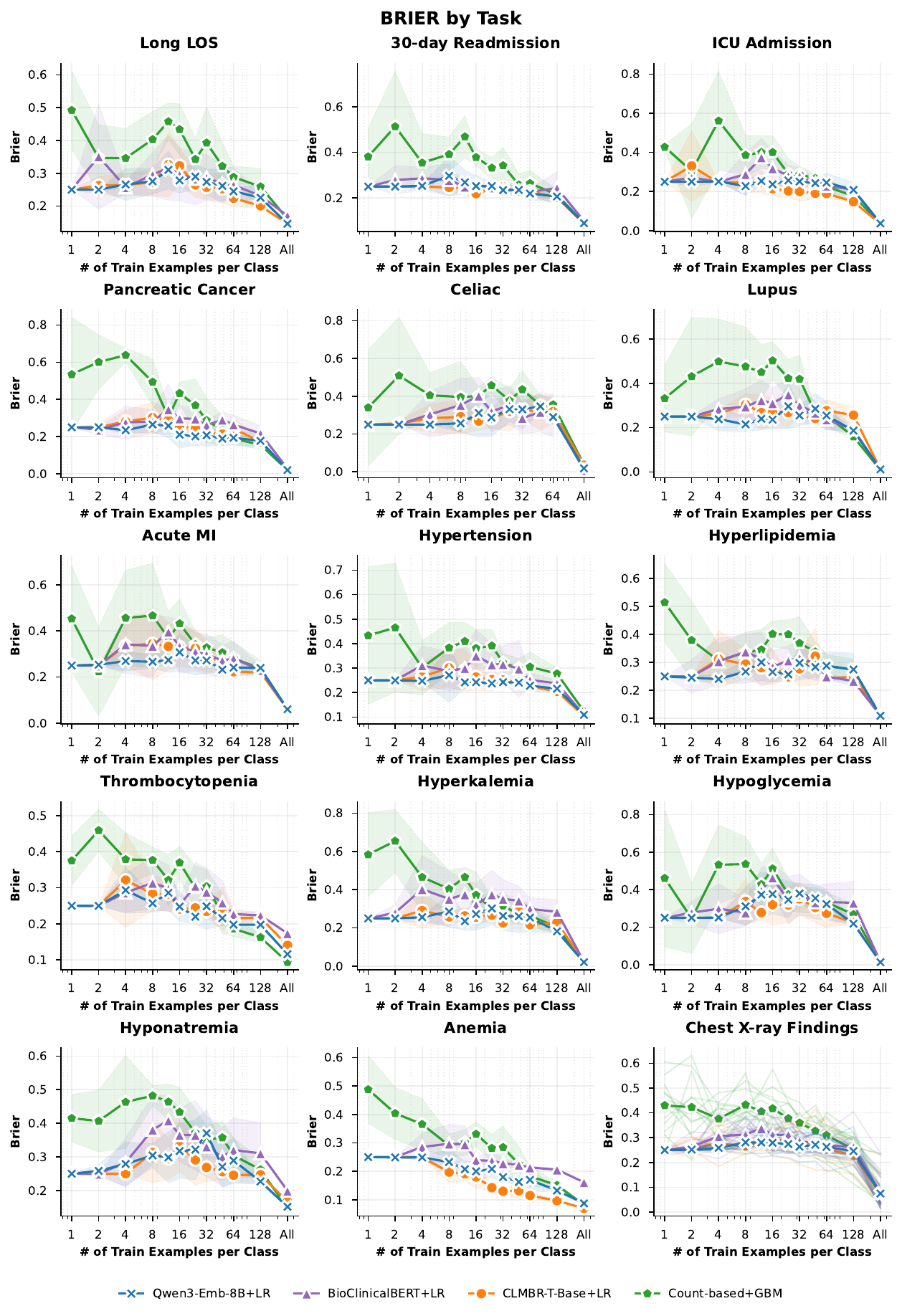}
    \caption{\textbf{Task-Specific Brier Score on EHRSHOT.} Brier score with standard deviation across five few-shot replicates for all \np{15} prediction tasks \cite{wornow_ehrshot_2023}.}
    \label{fig:task_specific_brier_performance}
\end{figure}
\clearpage

\subsection{Additional Performance Results on UK Biobank}
\begin{table}[h]
    \caption{\textbf{Performance for All Examples on UKB.} Mean \acf{auroc} performance with approximate 95\% confidence intervals for three task groups. The assignment of new diagnoses prediction is based on the mean of all 23 provided diseases. The macro-averaged performance across all task groups is given in the right-most column. All LLM embedding models use a context size of 8,192 tokens.}
    \label{tab:ukb_performance_on_all_examples_full}
    \centering
    \footnotesize
    \setlength{\tabcolsep}{2.6pt} 
    \begin{tabular}{>{\raggedright\arraybackslash}p{3.2cm} 
                >{\raggedright\arraybackslash}p{2.2cm} 
                >{\raggedright\arraybackslash}p{2.4cm} 
                >{\raggedright\arraybackslash}p{2.2cm} 
                >{\raggedright\arraybackslash}p{2.2cm}@{}}
    \toprule
\textbf{Model}                           & \textbf{Mortality prediction} & \textbf{Operational Outcomes (Hospitalization)} & \textbf{Assignment of New Diagnoses}  & \textbf{Macro Avg. Across Task Groups}  \\ \midrule
\multicolumn{5}{l}{\textbf{Baselines} \cite{wornow_ehrshot_2023}} \\ \midrule
CLMBR-T-Base                & $\ci{0.801}{.772}{.830}$  & $\ci{0.689}{.685}{.693}$ & $\ci{0.719}{.708}{.731}$  & $\ci{0.736}{.726}{.747}$ \\
Count-based + GBM           & $\ci{0.780}{.749}{.810}$  & $\ci{0.709}{.705}{.712}$ & $\ci{0.634}{.622}{.646}$  & $\ci{0.708}{.697}{.719}$ \\ 
\midrule
\multicolumn{5}{l}{\textbf{LLM Embedding Model}} \\ \midrule
Qwen3-Emb-8B                & $\ci{0.811}{.781}{.840}$  & $\ci{0.698}{.694}{.702}$ & $\ci{0.743}{.731}{.755}$  & $\ci{0.751}{.740}{.761}$ \\ 
Qwen2-Emb-7B                & $\ci{0.804}{.774}{.835}$  & $\ci{0.695}{.691}{.699}$ & $\ci{0.747}{.732}{.762}$  & $\ci{0.749}{.737}{.760}$ \\
Llama3.1-LLM2Vec-8B        & $\ci{0.796}{.766}{.827}$  & $\ci{0.690}{.686}{.694}$ & $\ci{0.731}{.717}{.746}$  & $\ci{0.739}{.728}{.751}$ \\ \midrule
\multicolumn{5}{l}{\textbf{Sensitivity Analysis Restricted to CLMBR Codes}} \\ \midrule
Qwen3-Emb-8B CLMBR-T-Base codes    & $\ci{0.806}{.775}{.836}$  & $\ci{0.687}{.683}{.691}$ & $\ci{0.736}{.723}{.749}$  & $\ci{0.743}{.732}{.754}$ \\ \midrule
\multicolumn{5}{l}{\textbf{Encoder Language Models with Chunked Inputs}} \\ \midrule
BioClinicalBERT               & $\ci{0.778}{.746}{.811}$  & $\ci{0.675}{.671}{.678}$ & $\ci{0.705}{.693}{.717}$  & $\ci{0.719}{.708}{.731}$ \\
\bottomrule
\end{tabular}
\end{table}

\begin{table}[]
\caption{\textbf{Per-task $\Delta$AUROC (Qwen3-Emb-8B minus baseline) on UKB.}
Cells report the AUROC difference ($\Delta$AUROC), 95\% bootstrap confidence intervals, and Holm-adjusted $p$-values obtained from paired patient-level bootstrap tests with 10{,}000 resamples. Positive values indicate better performance of Qwen3-Emb-8B. 
Multiple testing was controlled separately for each shot setting ($k=8$, $k=64$, and all training data) using Holm’s procedure across all tasks and baseline comparisons (75/60 tests per setting). Bold indicates statistically significant differences ($p_{\text{adj}}<0.05$).}
\label{tab:ukb_significance_qwen}
\centering
\footnotesize
\setlength{\tabcolsep}{2pt}
\begin{tabular}{lccc}
\toprule
\textbf{Task} & \textbf{CLMBR-T-Base} & \textbf{BioClinicalBERT} & \textbf{Count-based Model} \\
\midrule
\multicolumn{4}{l}{\textbf{8-shot (8 positive / 8 negative examples)}} \\
\midrule
Hospitalization & \bestcip{+0.044}{+0.039}{+0.049}{0.015} & \bestcip{+0.042}{+0.038}{+0.047}{0.015} & \bestcip{-0.032}{-0.036}{-0.027}{0.015} \\
Death & \estcip{+0.032}{+0.000}{+0.065}{1.000} & \bestcip{+0.083}{+0.046}{+0.121}{0.015} & \bestcip{+0.076}{+0.036}{+0.117}{0.015} \\
Hypertension & \estcip{+0.010}{-0.004}{+0.024}{1.000} & \bestcip{+0.043}{+0.029}{+0.058}{0.015} & \estcip{+0.006}{-0.008}{+0.020}{1.000} \\
Diabetes Mellitus & \estcip{+0.016}{-0.014}{+0.045}{1.000} & \estcip{-0.044}{-0.073}{-0.015}{0.312} & \estcip{+0.012}{-0.019}{+0.042}{1.000} \\
Atrial Fibrillation & \estcip{-0.023}{-0.069}{+0.023}{1.000} & \estcip{-0.004}{-0.040}{+0.033}{1.000} & \estcip{-0.013}{-0.048}{+0.024}{1.000} \\
Pneumonia & \estcip{-0.045}{-0.084}{-0.006}{1.000} & \estcip{+0.029}{-0.005}{+0.063}{1.000} & \estcip{-0.028}{-0.069}{+0.012}{1.000} \\
COPD & \estcip{+0.021}{-0.010}{+0.052}{1.000} & \bestcip{+0.072}{+0.040}{+0.105}{0.015} & \estcip{-0.006}{-0.038}{+0.027}{1.000} \\
Chronic Kidney Disease & \estcip{+0.046}{+0.013}{+0.078}{0.399} & \estcip{-0.020}{-0.052}{+0.012}{1.000} & \estcip{-0.021}{-0.051}{+0.009}{1.000} \\
Ischemic Heart Disease & \estcip{+0.036}{+0.011}{+0.061}{0.360} & \estcip{-0.007}{-0.030}{+0.016}{1.000} & \estcip{-0.029}{-0.056}{-0.002}{1.000} \\
Myocardial Infarction & \estcip{+0.052}{-0.001}{+0.104}{1.000} & \bestcip{+0.114}{+0.061}{+0.166}{0.015} & \estcip{+0.030}{-0.019}{+0.078}{1.000} \\
Cerebral Infarction & \estcip{+0.114}{+0.041}{+0.182}{0.126} & \estcip{-0.013}{-0.070}{+0.045}{1.000} & \estcip{-0.050}{-0.101}{-0.002}{1.000} \\
Heart Failure & \estcip{-0.079}{-0.133}{-0.025}{0.319} & \estcip{-0.040}{-0.100}{+0.019}{1.000} & \estcip{-0.031}{-0.085}{+0.023}{1.000} \\
Cardiac Arrest & \estcip{+0.010}{-0.105}{+0.123}{1.000} & \estcip{+0.022}{-0.073}{+0.112}{1.000} & \estcip{+0.144}{+0.029}{+0.264}{0.795} \\
Abdominal Aortic Aneurysm & \estcip{+0.036}{-0.031}{+0.100}{1.000} & \estcip{+0.013}{-0.034}{+0.062}{1.000} & \estcip{+0.033}{-0.052}{+0.118}{1.000} \\
Pulmonary Embolism & \estcip{-0.113}{-0.207}{-0.017}{1.000} & \estcip{+0.122}{+0.039}{+0.203}{0.232} & \estcip{-0.059}{-0.153}{+0.031}{1.000} \\
Aortic Stenosis & \estcip{-0.011}{-0.082}{+0.059}{1.000} & \estcip{+0.033}{-0.078}{+0.148}{1.000} & \estcip{-0.018}{-0.144}{+0.114}{1.000} \\
Mitral Valve Insufficiency & \estcip{+0.072}{-0.003}{+0.143}{1.000} & \estcip{-0.049}{-0.124}{+0.027}{1.000} & \estcip{+0.078}{-0.015}{+0.173}{1.000} \\
Endocarditis & \estcip{+0.042}{-0.061}{+0.142}{1.000} & \estcip{-0.011}{-0.102}{+0.078}{1.000} & \estcip{+0.007}{-0.095}{+0.108}{1.000} \\
Rheumatic Fever & \estcip{+0.080}{+0.016}{+0.143}{0.792} & \estcip{+0.085}{+0.021}{+0.152}{0.582} & \bestcip{+0.206}{+0.110}{+0.302}{0.015} \\
Anemia & \estcip{+0.021}{-0.005}{+0.047}{1.000} & \estcip{+0.017}{-0.009}{+0.043}{1.000} & \estcip{+0.044}{+0.017}{+0.071}{0.126} \\
Back Pain & \bestcip{+0.126}{+0.105}{+0.146}{0.015} & \bestcip{+0.052}{+0.038}{+0.066}{0.015} & \estcip{-0.004}{-0.017}{+0.010}{1.000} \\
Parkinson's Disease & \estcip{-0.075}{-0.215}{+0.069}{1.000} & \estcip{+0.058}{-0.075}{+0.184}{1.000} & \estcip{-0.077}{-0.208}{+0.062}{1.000} \\
Rheumatoid Arthritis & \estcip{-0.059}{-0.133}{+0.016}{1.000} & \estcip{-0.007}{-0.082}{+0.069}{1.000} & \estcip{-0.006}{-0.066}{+0.055}{1.000} \\
Psoriasis & \estcip{+0.079}{+0.015}{+0.143}{0.792} & \estcip{+0.043}{-0.004}{+0.091}{1.000} & \estcip{+0.027}{-0.020}{+0.073}{1.000} \\
Suicide Ideation / Self Harm & \estcip{+0.041}{-0.027}{+0.108}{1.000} & \estcip{+0.117}{+0.047}{+0.191}{0.090} & \estcip{+0.072}{-0.007}{+0.154}{1.000} \\
\midrule
\multicolumn{4}{l}{\textbf{64-shot (64 positive / 64 negative examples)}} \\
\midrule
Hospitalization & \bestcip{+0.023}{+0.018}{+0.028}{0.012} & \estcip{-0.002}{-0.006}{+0.002}{1.000} & \estcip{+0.002}{-0.003}{+0.007}{1.000} \\
Death & \estcip{+0.017}{-0.008}{+0.040}{1.000} & \estcip{+0.029}{+0.003}{+0.055}{1.000} & \estcip{+0.006}{-0.019}{+0.031}{1.000} \\
Hypertension & \estcip{+0.007}{-0.006}{+0.020}{1.000} & \bestcip{+0.038}{+0.026}{+0.050}{0.012} & \estcip{-0.003}{-0.016}{+0.010}{1.000} \\
Diabetes Mellitus & \bestcip{+0.068}{+0.044}{+0.092}{0.012} & \estcip{+0.031}{+0.012}{+0.051}{0.150} & \estcip{+0.008}{-0.011}{+0.027}{1.000} \\
Atrial Fibrillation & \estcip{+0.004}{-0.031}{+0.041}{1.000} & \estcip{+0.036}{+0.007}{+0.067}{0.475} & \estcip{+0.007}{-0.018}{+0.032}{1.000} \\
Pneumonia & \estcip{-0.031}{-0.065}{+0.003}{1.000} & \estcip{-0.021}{-0.056}{+0.014}{1.000} & \estcip{-0.063}{-0.099}{-0.027}{0.071} \\
COPD & \estcip{+0.029}{+0.005}{+0.054}{0.697} & \bestcip{+0.079}{+0.052}{+0.106}{0.012} & \estcip{-0.016}{-0.038}{+0.006}{1.000} \\
Chronic Kidney Disease & \estcip{+0.017}{-0.003}{+0.038}{1.000} & \estcip{+0.022}{+0.006}{+0.037}{0.202} & \estcip{-0.005}{-0.021}{+0.011}{1.000} \\
Ischemic Heart Disease & \estcip{-0.018}{-0.040}{+0.003}{1.000} & \estcip{+0.014}{-0.005}{+0.033}{1.000} & \estcip{-0.033}{-0.054}{-0.011}{0.125} \\
Myocardial Infarction & \estcip{-0.027}{-0.063}{+0.009}{1.000} & \estcip{+0.068}{+0.028}{+0.108}{0.064} & \estcip{-0.022}{-0.057}{+0.013}{1.000} \\
Cerebral Infarction & \estcip{+0.038}{-0.010}{+0.085}{1.000} & \estcip{+0.034}{-0.001}{+0.069}{1.000} & \estcip{-0.040}{-0.096}{+0.015}{1.000} \\
Heart Failure & \estcip{+0.009}{-0.020}{+0.038}{1.000} & \estcip{+0.037}{+0.008}{+0.066}{0.475} & \estcip{+0.005}{-0.029}{+0.041}{1.000} \\
Cardiac Arrest & --- & --- & --- \\
Abdominal Aortic Aneurysm & --- & --- & --- \\
Pulmonary Embolism & \estcip{-0.047}{-0.122}{+0.029}{1.000} & \estcip{+0.090}{+0.023}{+0.157}{0.360} & \estcip{+0.017}{-0.063}{+0.099}{1.000} \\
Aortic Stenosis & --- & --- & --- \\
Mitral Valve Insufficiency & \estcip{+0.053}{-0.000}{+0.107}{1.000} & \estcip{+0.079}{+0.028}{+0.132}{0.071} & \bestcip{+0.132}{+0.067}{+0.200}{0.012} \\
Endocarditis & --- & --- & --- \\
Rheumatic Fever & \estcip{+0.031}{-0.027}{+0.090}{1.000} & \estcip{-0.011}{-0.074}{+0.049}{1.000} & \estcip{+0.004}{-0.047}{+0.054}{1.000} \\
Anemia & \estcip{+0.002}{-0.023}{+0.027}{1.000} & \estcip{+0.005}{-0.021}{+0.032}{1.000} & \estcip{-0.043}{-0.067}{-0.017}{0.078} \\
Back Pain & \bestcip{+0.041}{+0.030}{+0.052}{0.012} & \estcip{-0.001}{-0.011}{+0.008}{1.000} & \estcip{-0.003}{-0.013}{+0.006}{1.000} \\
Parkinson's Disease & --- & --- & --- \\
Rheumatoid Arthritis & \estcip{+0.009}{-0.047}{+0.068}{1.000} & \estcip{+0.041}{-0.012}{+0.094}{1.000} & \estcip{-0.005}{-0.058}{+0.048}{1.000} \\
Psoriasis & \estcip{+0.042}{+0.002}{+0.082}{1.000} & \estcip{+0.033}{+0.001}{+0.064}{1.000} & \estcip{+0.010}{-0.024}{+0.044}{1.000} \\
Suicide Ideation / Self Harm & \bestcip{+0.116}{+0.052}{+0.179}{0.043} & \estcip{+0.103}{+0.040}{+0.170}{0.064} & \estcip{+0.047}{-0.005}{+0.103}{1.000} \\
\bottomrule
\end{tabular}
\end{table}

\begin{table}[]
\caption{\textbf{Per-task $\Delta$AUROC (Qwen3-Emb-8B minus baseline) on UKB (continued).}}
\centering
\footnotesize
\setlength{\tabcolsep}{2pt}
\begin{tabular}{lccc}
\toprule
\textbf{Task} & \textbf{CLMBR-T-Base} & \textbf{BioClinicalBERT} & \textbf{Count-based Model} \\
\midrule
\multicolumn{4}{l}{\textbf{All training data}} \\
\midrule
Hospitalization & \bestcip{+0.009}{+0.007}{+0.012}{0.015} & \bestcip{+0.023}{+0.021}{+0.026}{0.015} & \bestcip{-0.011}{-0.013}{-0.008}{0.015} \\
Death & \estcip{+0.009}{-0.009}{+0.029}{1.000} & \estcip{+0.032}{+0.010}{+0.056}{0.263} & \estcip{+0.016}{-0.010}{+0.041}{1.000} \\
Hypertension & \bestcip{+0.019}{+0.009}{+0.028}{0.015} & \bestcip{+0.032}{+0.024}{+0.040}{0.015} & \bestcip{-0.022}{-0.031}{-0.012}{0.015} \\
Diabetes Mellitus & \bestcip{+0.046}{+0.027}{+0.065}{0.015} & \bestcip{+0.034}{+0.017}{+0.052}{0.021} & \estcip{-0.018}{-0.037}{+0.001}{1.000} \\
Atrial Fibrillation & \estcip{+0.038}{-0.001}{+0.080}{1.000} & \estcip{+0.030}{+0.001}{+0.060}{1.000} & \bestcip{+0.358}{+0.284}{+0.432}{0.015} \\
Pneumonia & \estcip{-0.019}{-0.044}{+0.005}{1.000} & \estcip{+0.036}{+0.011}{+0.060}{0.250} & \estcip{+0.005}{-0.022}{+0.032}{1.000} \\
COPD & \bestcip{+0.045}{+0.022}{+0.068}{0.015} & \bestcip{+0.067}{+0.046}{+0.089}{0.015} & \estcip{-0.019}{-0.041}{+0.003}{1.000} \\
Chronic Kidney Disease & \estcip{+0.026}{+0.008}{+0.045}{0.270} & \bestcip{+0.041}{+0.022}{+0.059}{0.015} & \bestcip{+0.045}{+0.024}{+0.067}{0.015} \\
Ischemic Heart Disease & \estcip{-0.009}{-0.024}{+0.006}{1.000} & \bestcip{+0.026}{+0.011}{+0.040}{0.041} & \estcip{-0.014}{-0.031}{+0.003}{1.000} \\
Myocardial Infarction & \estcip{-0.006}{-0.034}{+0.021}{1.000} & \estcip{+0.037}{+0.011}{+0.063}{0.273} & \estcip{-0.014}{-0.049}{+0.020}{1.000} \\
Cerebral Infarction & \estcip{+0.007}{-0.039}{+0.051}{1.000} & \estcip{+0.034}{-0.002}{+0.071}{1.000} & \estcip{+0.073}{-0.013}{+0.159}{1.000} \\
Heart Failure & \estcip{+0.014}{-0.014}{+0.042}{1.000} & \estcip{+0.040}{+0.013}{+0.067}{0.273} & \bestcip{+0.166}{+0.112}{+0.222}{0.015} \\
Cardiac Arrest & \estcip{+0.003}{-0.088}{+0.094}{1.000} & \estcip{+0.040}{-0.039}{+0.120}{1.000} & \estcip{+0.196}{+0.068}{+0.326}{0.167} \\
Abdominal Aortic Aneurysm & \estcip{+0.067}{+0.005}{+0.126}{1.000} & \estcip{+0.039}{-0.021}{+0.100}{1.000} & \bestcip{+0.346}{+0.236}{+0.449}{0.015} \\
Pulmonary Embolism & \estcip{-0.002}{-0.073}{+0.069}{1.000} & \estcip{+0.093}{+0.027}{+0.159}{0.267} & \estcip{+0.109}{+0.031}{+0.192}{0.311} \\
Aortic Stenosis & \estcip{+0.046}{-0.007}{+0.099}{1.000} & \estcip{+0.001}{-0.070}{+0.075}{1.000} & \estcip{+0.228}{+0.097}{+0.353}{0.080} \\
Mitral Valve Insufficiency & \estcip{+0.042}{-0.009}{+0.096}{1.000} & \estcip{+0.069}{+0.018}{+0.123}{0.435} & \bestcip{+0.300}{+0.232}{+0.363}{0.015} \\
Endocarditis & \estcip{+0.038}{-0.026}{+0.101}{1.000} & \estcip{-0.040}{-0.127}{+0.041}{1.000} & \estcip{+0.155}{+0.016}{+0.301}{1.000} \\
Rheumatic Fever & \estcip{+0.044}{-0.011}{+0.100}{1.000} & \estcip{+0.010}{-0.044}{+0.063}{1.000} & \bestcip{+0.318}{+0.248}{+0.382}{0.015} \\
Anemia & \bestcip{+0.038}{+0.019}{+0.058}{0.015} & \bestcip{+0.047}{+0.028}{+0.066}{0.015} & \estcip{+0.010}{-0.010}{+0.029}{1.000} \\
Back Pain & \bestcip{+0.055}{+0.046}{+0.063}{0.015} & \estcip{+0.008}{+0.001}{+0.015}{1.000} & \bestcip{-0.016}{-0.023}{-0.009}{0.015} \\
Parkinson's Disease & \estcip{-0.051}{-0.144}{+0.039}{1.000} & \bestcip{+0.131}{+0.060}{+0.204}{0.031} & \estcip{+0.135}{-0.019}{+0.277}{1.000} \\
Rheumatoid Arthritis & \estcip{+0.020}{-0.038}{+0.081}{1.000} & \estcip{+0.057}{-0.004}{+0.120}{1.000} & \estcip{+0.072}{-0.005}{+0.152}{1.000} \\
Psoriasis & \estcip{+0.031}{+0.001}{+0.061}{1.000} & \estcip{+0.001}{-0.027}{+0.028}{1.000} & \bestcip{+0.127}{+0.074}{+0.179}{0.015} \\
Suicide Ideation / Self Harm & \estcip{+0.058}{+0.008}{+0.107}{1.000} & \estcip{+0.048}{-0.013}{+0.110}{1.000} & \bestcip{+0.300}{+0.152}{+0.445}{0.015} \\
\bottomrule
\end{tabular}
\end{table}

\begin{table}[]
\caption{\textbf{Per-task $\Delta$AUROC (Qwen3-Emb-8B restricted to CLMBR-T-Base codes minus CLMBR-T-Base) on UKB.}
Cells report the AUROC difference ($\Delta$AUROC), 95\% bootstrap confidence intervals, and Holm-adjusted $p$-values obtained from paired patient-level bootstrap tests with 10{,}000 resamples. Positive values indicate better performance of Qwen3-Emb-8B. 
Multiple testing was controlled separately for each shot setting ($k=8$, $k=64$, and all training data) using Holm’s procedure across all tasks and baseline comparisons (25/20 tests per setting). Bold indicates statistically significant differences ($p_{\text{adj}}<0.05$).}
\label{tab:ukb_significance_qwen_sensitivity}
\centering
\footnotesize
\setlength{\tabcolsep}{2pt}
\begin{tabular}{lccc}
\toprule
\textbf{Task} & \textbf{8-shot} & \textbf{64-shot} & \textbf{All training data} \\
\midrule
Hospitalization & \bestcip{+0.048}{+0.043}{+0.053}{0.005} & \estcip{-0.007}{-0.012}{-0.002}{0.116} & \estcip{-0.002}{-0.004}{+0.001}{1.000} \\
Death & \bestcip{-0.156}{-0.200}{-0.111}{0.005} & \estcip{+0.005}{-0.021}{+0.033}{1.000} & \estcip{+0.004}{-0.015}{+0.024}{1.000} \\
Hypertension & \estcip{+0.007}{-0.007}{+0.020}{1.000} & \estcip{-0.013}{-0.026}{+0.000}{0.780} & \estcip{+0.003}{-0.006}{+0.012}{1.000} \\
Diabetes Mellitus & \bestcip{-0.060}{-0.094}{-0.026}{0.008} & \estcip{+0.031}{+0.006}{+0.057}{0.266} & \bestcip{+0.035}{+0.016}{+0.054}{0.014} \\
Atrial Fibrillation & \estcip{-0.008}{-0.055}{+0.041}{1.000} & \estcip{-0.026}{-0.064}{+0.013}{1.000} & \estcip{+0.006}{-0.033}{+0.047}{1.000} \\
Pneumonia & \estcip{-0.065}{-0.109}{-0.020}{0.094} & \estcip{+0.006}{-0.029}{+0.041}{1.000} & \estcip{-0.036}{-0.064}{-0.008}{0.232} \\
COPD & \estcip{+0.022}{-0.009}{+0.055}{1.000} & \estcip{+0.003}{-0.024}{+0.028}{1.000} & \estcip{+0.024}{+0.001}{+0.046}{0.788} \\
Chronic Kidney Disease & \bestcip{+0.067}{+0.040}{+0.094}{0.005} & \estcip{+0.006}{-0.013}{+0.026}{1.000} & \estcip{+0.002}{-0.017}{+0.021}{1.000} \\
Ischemic Heart Disease & \bestcip{+0.102}{+0.077}{+0.127}{0.005} & \estcip{+0.018}{-0.003}{+0.038}{1.000} & \estcip{+0.002}{-0.013}{+0.016}{1.000} \\
Myocardial Infarction & \bestcip{+0.082}{+0.031}{+0.133}{0.040} & \estcip{+0.015}{-0.016}{+0.046}{1.000} & \estcip{+0.030}{+0.004}{+0.056}{0.502} \\
Cerebral Infarction & \estcip{+0.081}{+0.017}{+0.143}{0.194} & \estcip{+0.018}{-0.030}{+0.066}{1.000} & \estcip{+0.013}{-0.026}{+0.049}{1.000} \\
Heart Failure & \estcip{-0.037}{-0.086}{+0.013}{1.000} & \estcip{-0.005}{-0.036}{+0.024}{1.000} & \estcip{+0.001}{-0.028}{+0.029}{1.000} \\
Cardiac Arrest & \estcip{+0.013}{-0.105}{+0.124}{1.000} & --- & \estcip{-0.044}{-0.148}{+0.055}{1.000} \\
Abdominal Aortic Aneurysm & \estcip{+0.060}{+0.010}{+0.112}{0.352} & --- & \estcip{+0.083}{+0.021}{+0.140}{0.202} \\
Pulmonary Embolism & \estcip{-0.112}{-0.207}{-0.015}{0.402} & \estcip{-0.056}{-0.125}{+0.013}{1.000} & \estcip{-0.039}{-0.103}{+0.026}{1.000} \\
Aortic Stenosis & \estcip{+0.049}{-0.030}{+0.131}{1.000} & --- & \estcip{+0.065}{-0.009}{+0.137}{1.000} \\
Mitral Valve Insufficiency & \estcip{+0.077}{-0.005}{+0.155}{0.845} & \estcip{+0.031}{-0.023}{+0.085}{1.000} & \estcip{+0.016}{-0.033}{+0.065}{1.000} \\
Endocarditis & \estcip{+0.085}{+0.003}{+0.170}{0.619} & --- & \estcip{+0.075}{-0.020}{+0.164}{1.000} \\
Rheumatic Fever & \estcip{-0.060}{-0.127}{+0.008}{0.994} & \estcip{+0.022}{-0.040}{+0.084}{1.000} & \estcip{-0.030}{-0.091}{+0.031}{1.000} \\
Anemia & \estcip{-0.003}{-0.032}{+0.027}{1.000} & \estcip{+0.014}{-0.011}{+0.039}{1.000} & \estcip{+0.027}{+0.010}{+0.045}{0.070} \\
Back Pain & \estcip{-0.010}{-0.030}{+0.010}{1.000} & \bestcip{+0.024}{+0.013}{+0.035}{0.004} & \bestcip{+0.040}{+0.032}{+0.048}{0.005} \\
Parkinson's Disease & \estcip{-0.059}{-0.176}{+0.067}{1.000} & --- & \estcip{-0.029}{-0.136}{+0.072}{1.000} \\
Rheumatoid Arthritis & \estcip{-0.044}{-0.115}{+0.026}{1.000} & \estcip{-0.044}{-0.104}{+0.017}{1.000} & \estcip{+0.034}{-0.024}{+0.096}{1.000} \\
Psoriasis & \bestcip{+0.092}{+0.029}{+0.154}{0.042} & \bestcip{+0.062}{+0.024}{+0.100}{0.034} & \estcip{+0.049}{+0.019}{+0.080}{0.055} \\
Suicide Ideation / Self Harm & \estcip{-0.046}{-0.134}{+0.042}{1.000} & \bestcip{+0.109}{+0.043}{+0.178}{0.036} & \estcip{+0.052}{-0.001}{+0.105}{0.887} \\
\bottomrule
\end{tabular}
\end{table}

\clearpage
\begin{figure}[h]
    \centering
    \includegraphics[width=\linewidth]
    {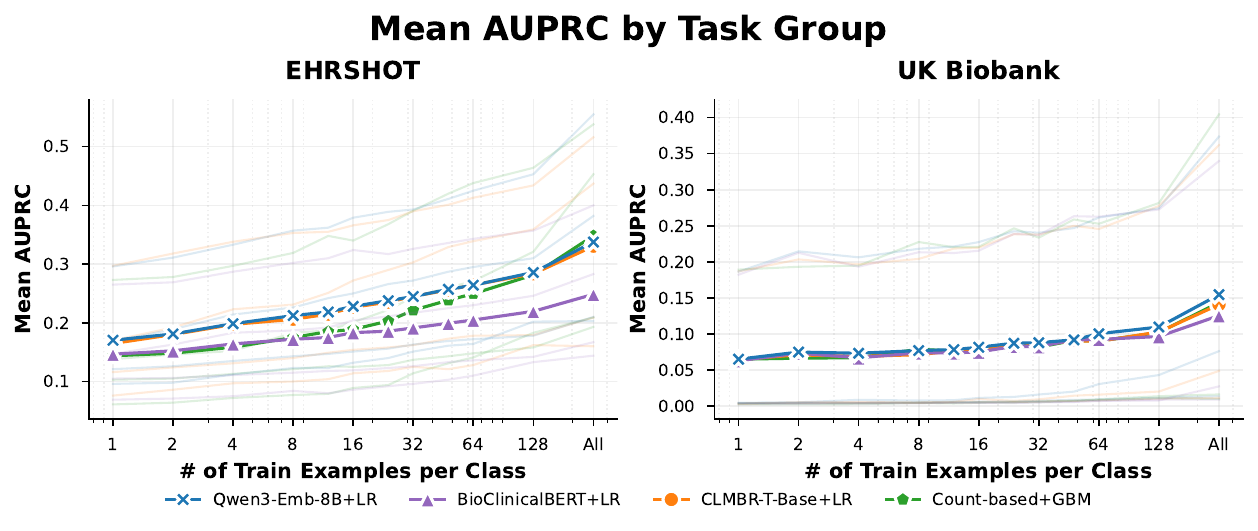}
    \vspace{-0.5cm}
    \caption{\textbf{Few-Shot Performance on EHRSHOT and UKB.} Macro-averaged \acf{auprc} performance across all subtasks of EHRSHOT (left) and UK Biobank (right). Blurred lines show averaged \ac{auprc} values for the different task groups.} 
    \label{fig:performance_EHRShot_UKB_AUPRC}
\end{figure}
\begin{figure}[h]
    \centering
    \includegraphics[width=\linewidth]{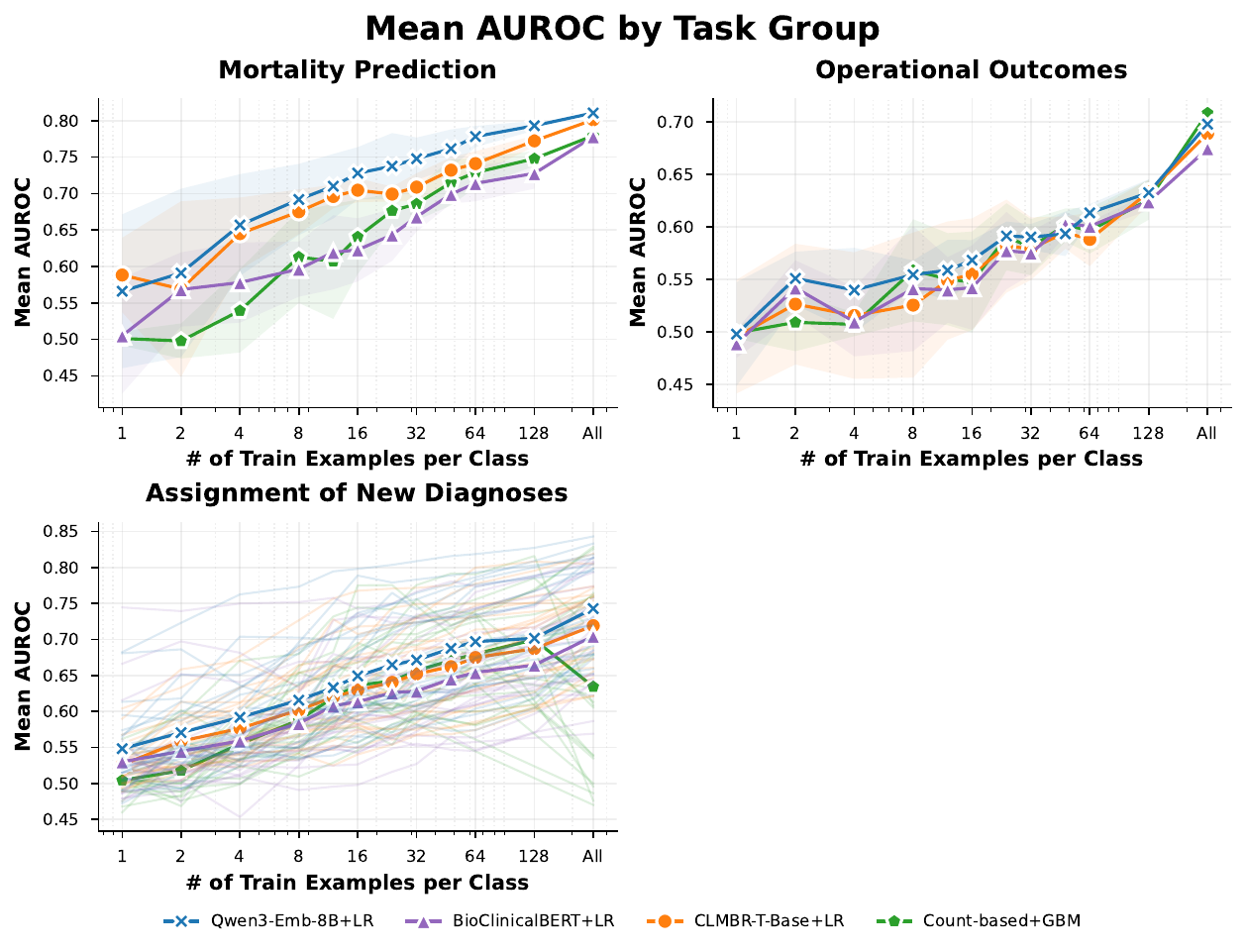}
    \caption{\textbf{Few-Shot AUROC Performance on UKB.} Mean \acf{auroc} performance across subtasks for three task groups (bold). Blurred lines show averages across five bootstrapped runs using different seeds. Shaded regions show standard deviation.}
    \label{fig:performance_auroc_UKB}
\end{figure}
\clearpage
\begin{figure}[h]
    \centering
    \includegraphics[width=\linewidth]{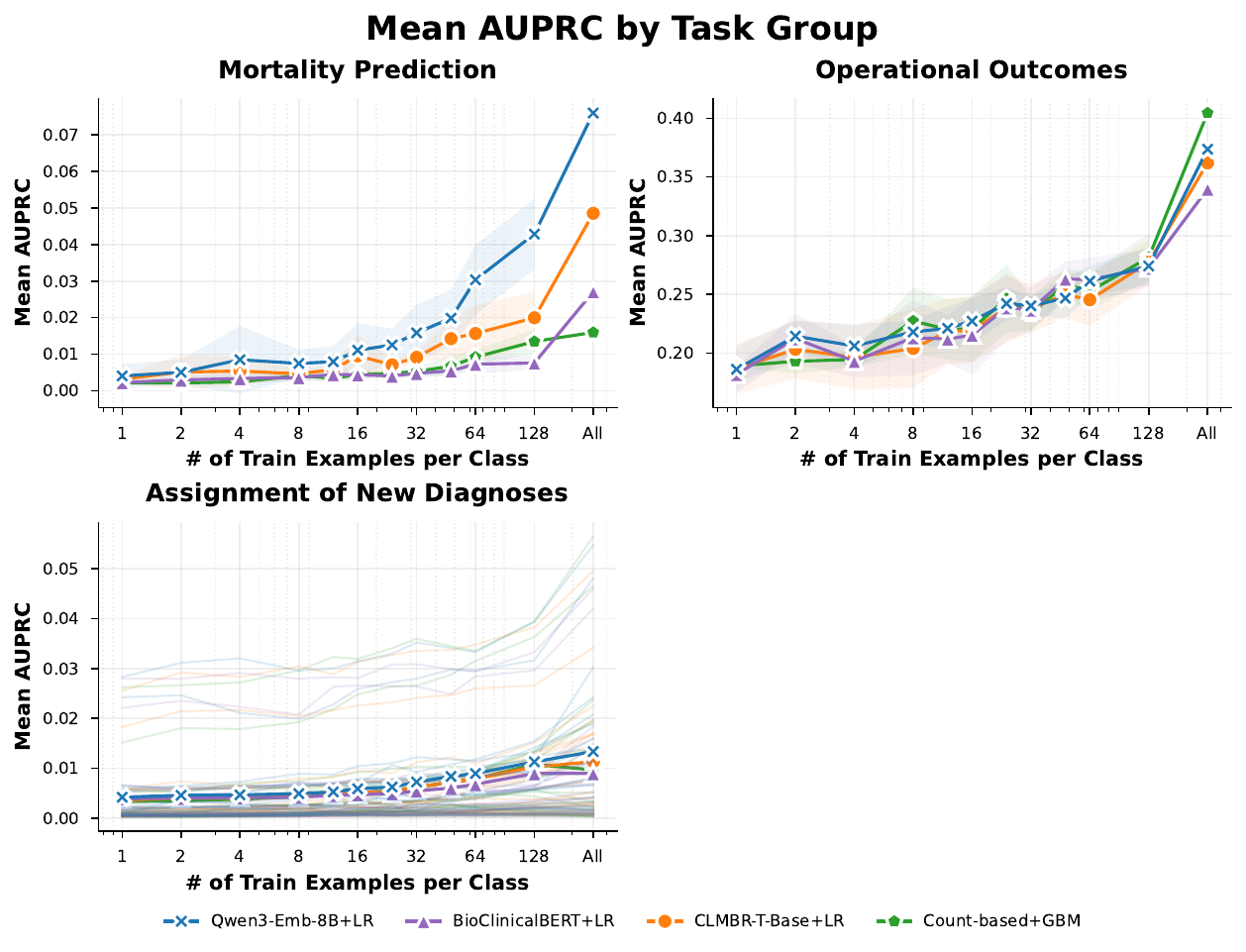}
    \caption{\textbf{Few-Shot AUPRC Performance on UKB.} Mean \acf{auprc} performance across subtasks for three task groups (bold). Blurred lines show averages across five bootstrapped runs using different seeds. Shaded regions show standard deviation.}
    \label{fig:performance_auprc_UKB}
\end{figure}
\begin{figure}[h]
    \centering
    \includegraphics[width=\linewidth]{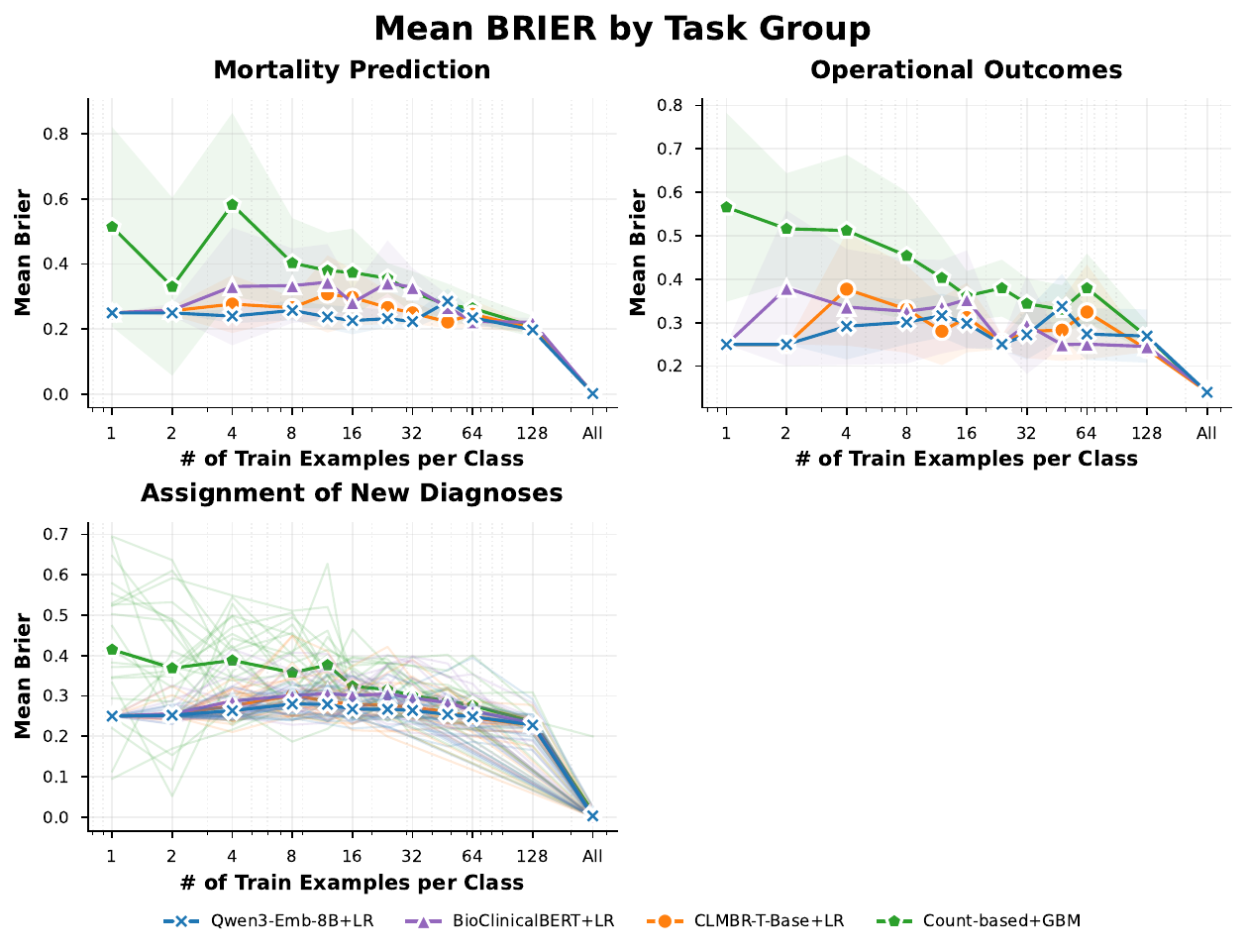}
    \caption{\textbf{Few-Shot Brier Score on UKB.} Mean Brier score across subtasks for three task groups (bold). Blurred lines show averages across five bootstrapped runs using different seeds \cite{wornow_ehrshot_2023}. Shaded regions show standard deviation.}
    \label{fig:brier_performance_in_few_shot_settings_ukb}
\end{figure}
\clearpage
\begin{figure}[h]
    \centering
    \includegraphics[width=0.95\linewidth]{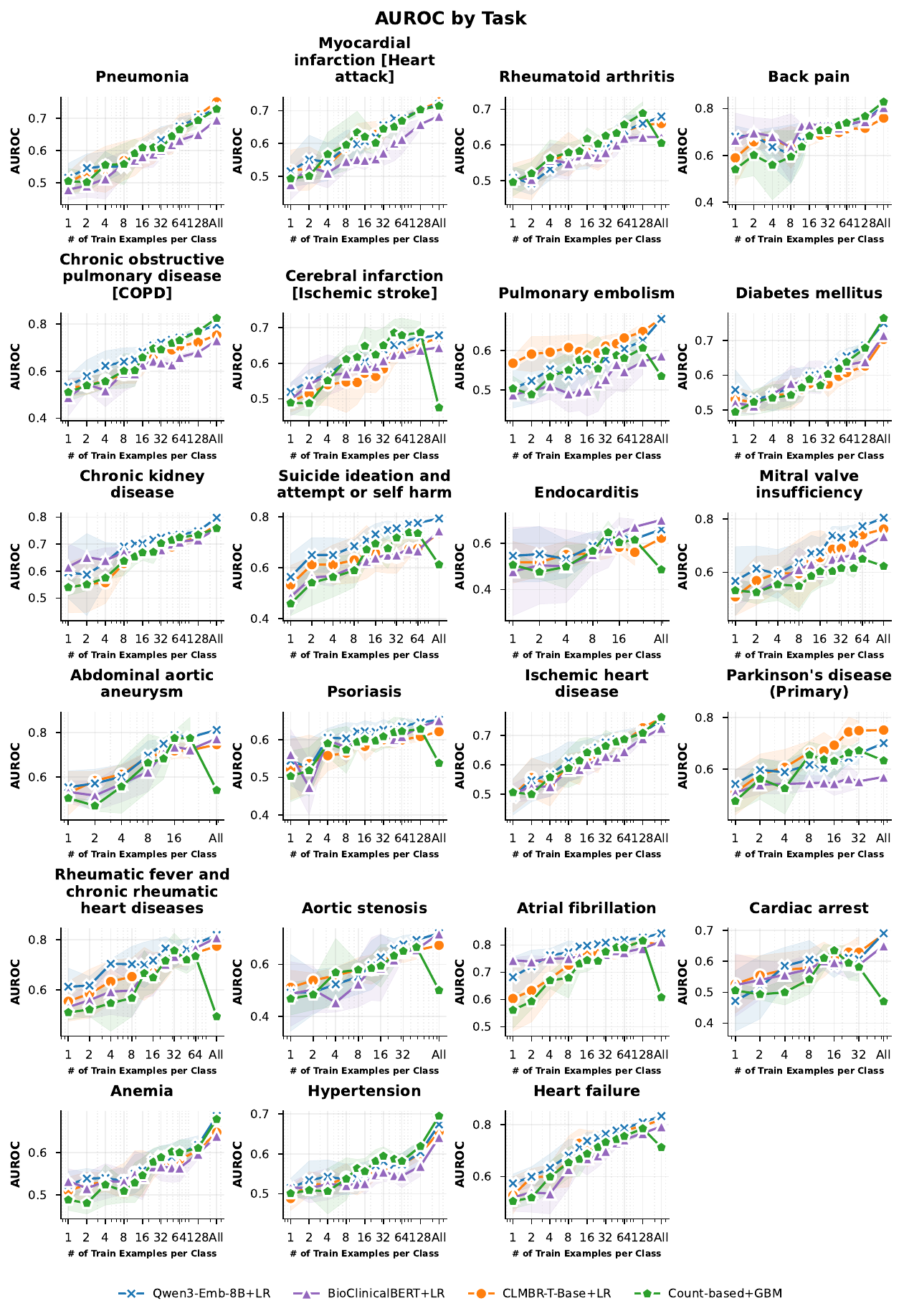}
    \caption{\textbf{Disease Onset AUROC Performance on UKB.} \Acf{auroc} performance with standard deviation across five few-shot replicates for all assignment-of-new-diagnosis tasks.}
    \label{fig:performance_tasks_auroc_UKB}
\end{figure}
\clearpage
\begin{figure}[h]
    \centering
    \includegraphics[width=0.95\linewidth]{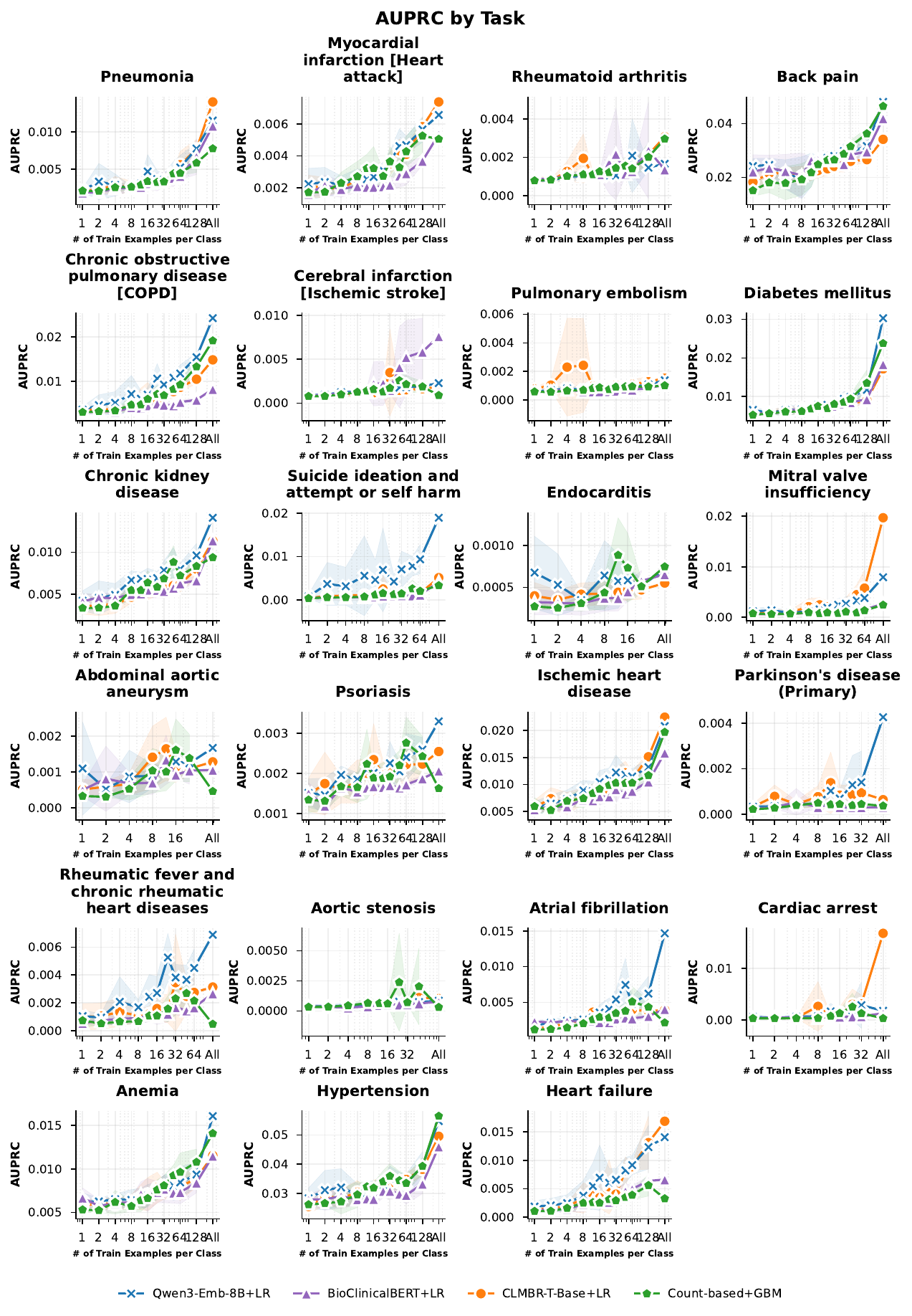}
    \caption{\textbf{Disease Onset AUPRC Performance on UKB.} \Acf{auprc} performance with standard deviation across five few-shot replicates for all assignment-of-new-diagnosis tasks.}
    \label{fig:performance_tasks_auprc_UKB}
\end{figure}
\clearpage
\begin{figure}[h]
    \centering
    \includegraphics[width=0.95\linewidth]{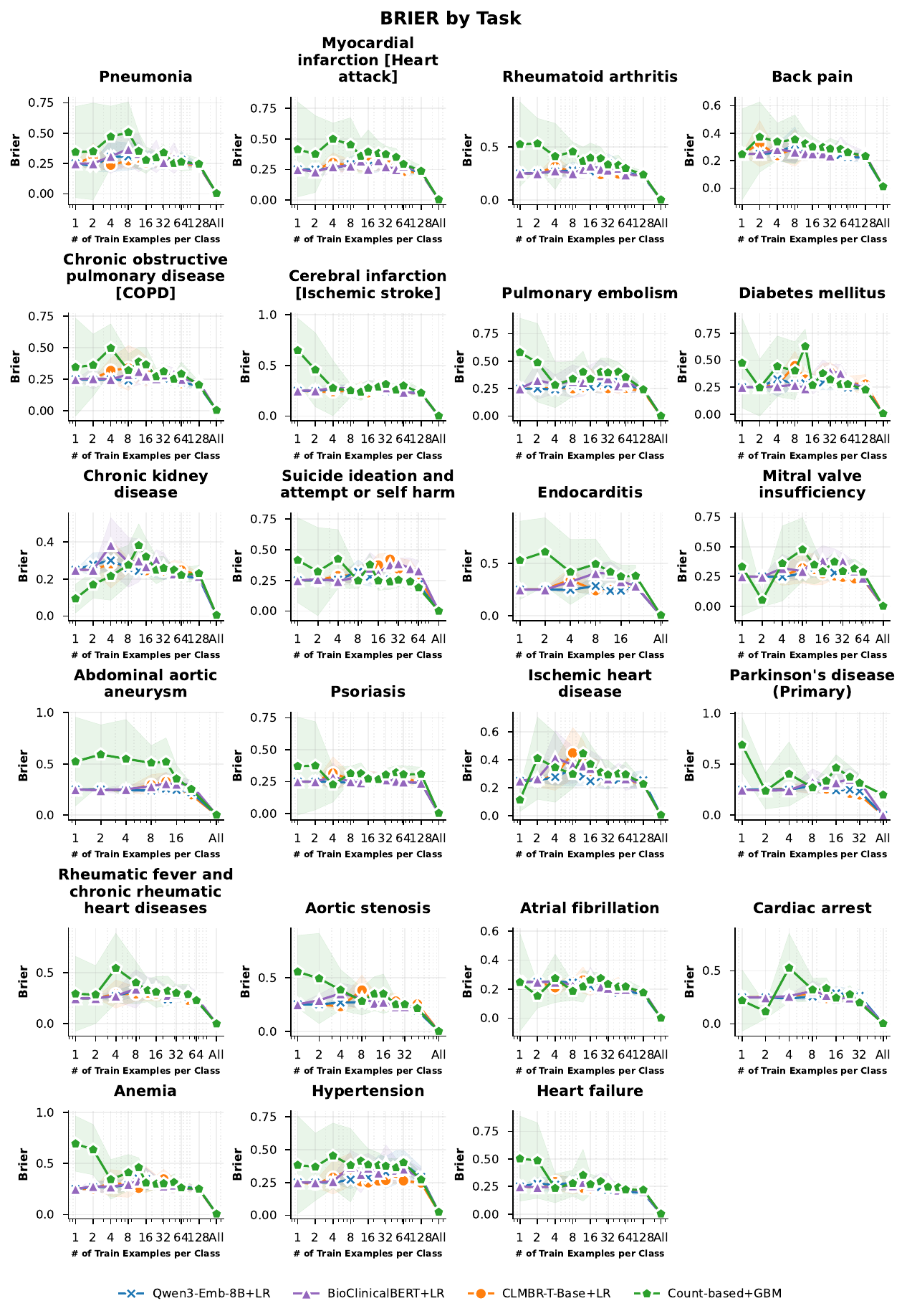}
    \caption{\textbf{Disease Onset Brier Score on UKB.} Brier score with standard deviation across five few-shot replicates for all assignment-of-new-diagnosis tasks.}
    \label{fig:task_specific_brier_performance_ukb}
\end{figure}
\clearpage
\begin{figure}[h]
    \centering
    \includegraphics[width=0.95\linewidth]{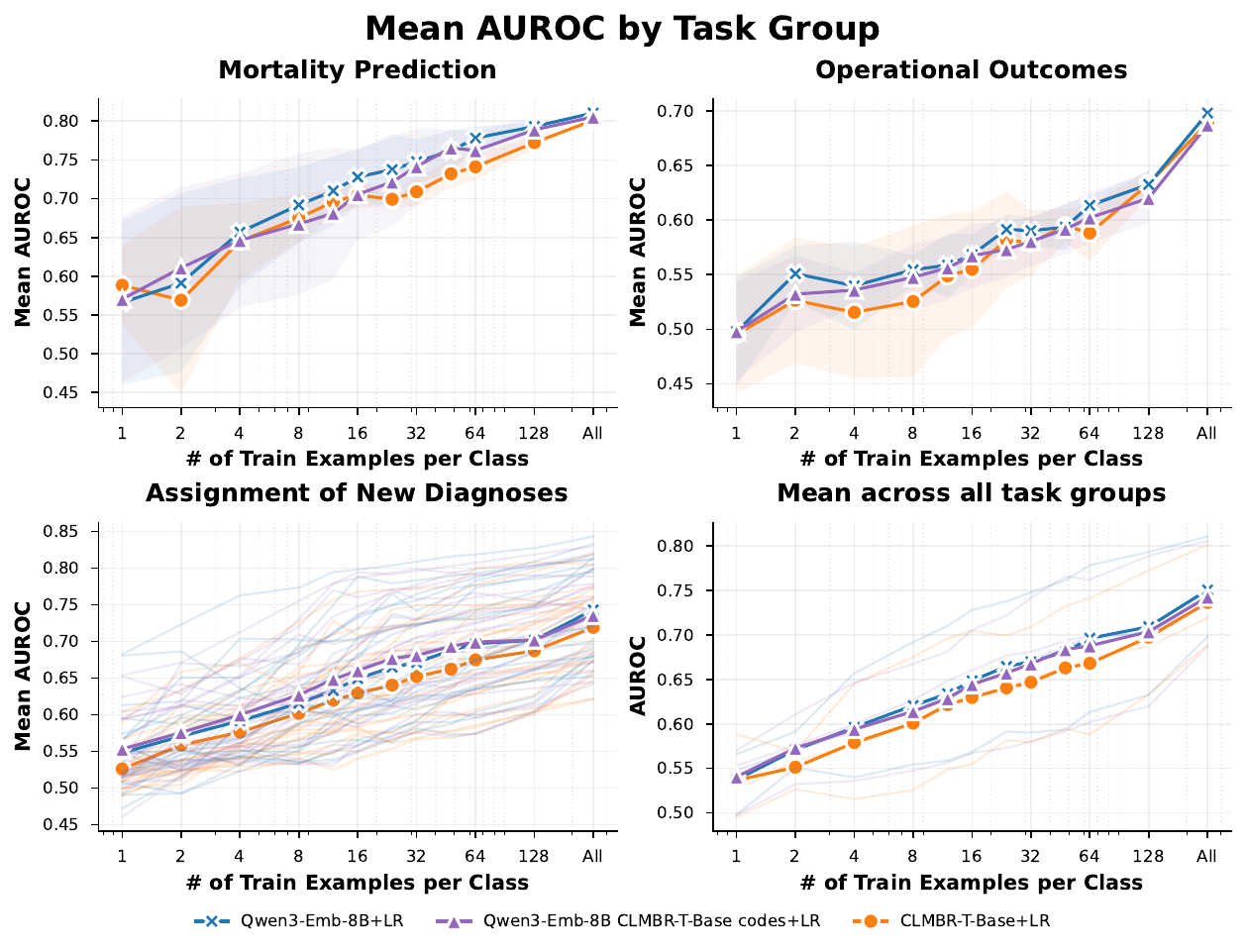}
    \caption{\textbf{Few-Shot AUROC Performance for Sensitivity Analysis on UKB.} Mean \acf{auroc} performance across subtasks for three task groups (bold) and macro-averaged AUROC performance across all subtasks on the UK Biobank. Results are reported for Qwen3-Emb-8B using all UKB codes, Qwen3-Emb-8B restricted to medical codes mappable to the \ac{ehr} foundation model CLMBR-T-Base, and the CLMBR-T-Base model. Shaded regions indicate standard deviation. Blurred lines for assignment of new diagnoses represent averages across five bootstrapped runs using different seeds. The blurred lines for the mean across all task groups represent the averaged AUROC values of the different task groups.}
    \label{fig:sensitivity_analysis_auroc_ukb}
\end{figure}
\clearpage

\subsection{Additional Performance Results for Encoder and Decoder Models}

\begin{table}[h!]
    \caption{\textbf{Performance for All Examples on EHRSHOT.} Macro-averaged AUROC performance and bootstrapped 95\% confidence intervals for the frozen Qwen3-Emb-8B encoder baseline and the LoRA-tuned Qwen encoder and decoder variants at $k=128$. Fine-tuning via LoRA did not improve over the frozen Qwen3-Emb-8B baseline. Among the tuned variants, the fine-tuned decoder slightly outperformed the fine-tuned encoder, but both remained below the frozen baseline and the decoder required substantially higher computational cost.}
    \label{tab:ehrshot_k128_encoder_decoder}
    \centering
    \footnotesize
    \setlength{\tabcolsep}{2.6pt}
    \begin{tabular}{>{\raggedright\arraybackslash}p{3.05cm}
                >{\raggedright\arraybackslash}p{1.8cm}
                >{\raggedright\arraybackslash}p{1.8cm}
                >{\raggedright\arraybackslash}p{1.8cm}
                >{\raggedright\arraybackslash}p{1.8cm}
                >{\raggedright\arraybackslash}p{1.8cm}}
    \toprule
\textbf{Model}                           & \textbf{Operational Outcomes} & \textbf{Anticipating Lab Test Results} & \textbf{Assignment of New Diagnoses} & \textbf{Anticipating Chest X-ray Findings} & \textbf{Macro Avg. Across Task Groups} \\ \midrule
\multicolumn{6}{l}{\textbf{Encoder models}} \\ \midrule
Qwen3-Emb-8B & $\ci{0.767}{.726}{.803}$ & $\ci{0.770}{.749}{.789}$ & $\ci{0.702}{.614}{.783}$ & $\ci{0.695}{.654}{.733}$ & $\ci{0.733}{.686}{.777}$ \\
Qwen3-Emb-8B (LoRA) & $\ci{0.736}{.693}{.776}$ & $\ci{0.819}{.804}{.834}$ & $\ci{0.667}{.573}{.751}$ & $\ci{0.646}{.605}{.686}$ & $\ci{0.717}{.669}{.762}$ \\
\midrule
\multicolumn{6}{l}{\textbf{Decoder models}} \\ \midrule
Qwen3-8B (LoRA) & $\ci{0.692}{.651}{.732}$ & $\ci{0.848}{.835}{.860}$ & $\ci{0.711}{.620}{.801}$ & $\ci{0.626}{.588}{.664}$ & $\ci{0.719}{.673}{.764}$ \\
    \bottomrule
    \end{tabular}
\end{table}

\begin{figure}[htbp]
    \centering
    \includegraphics[width=\linewidth]{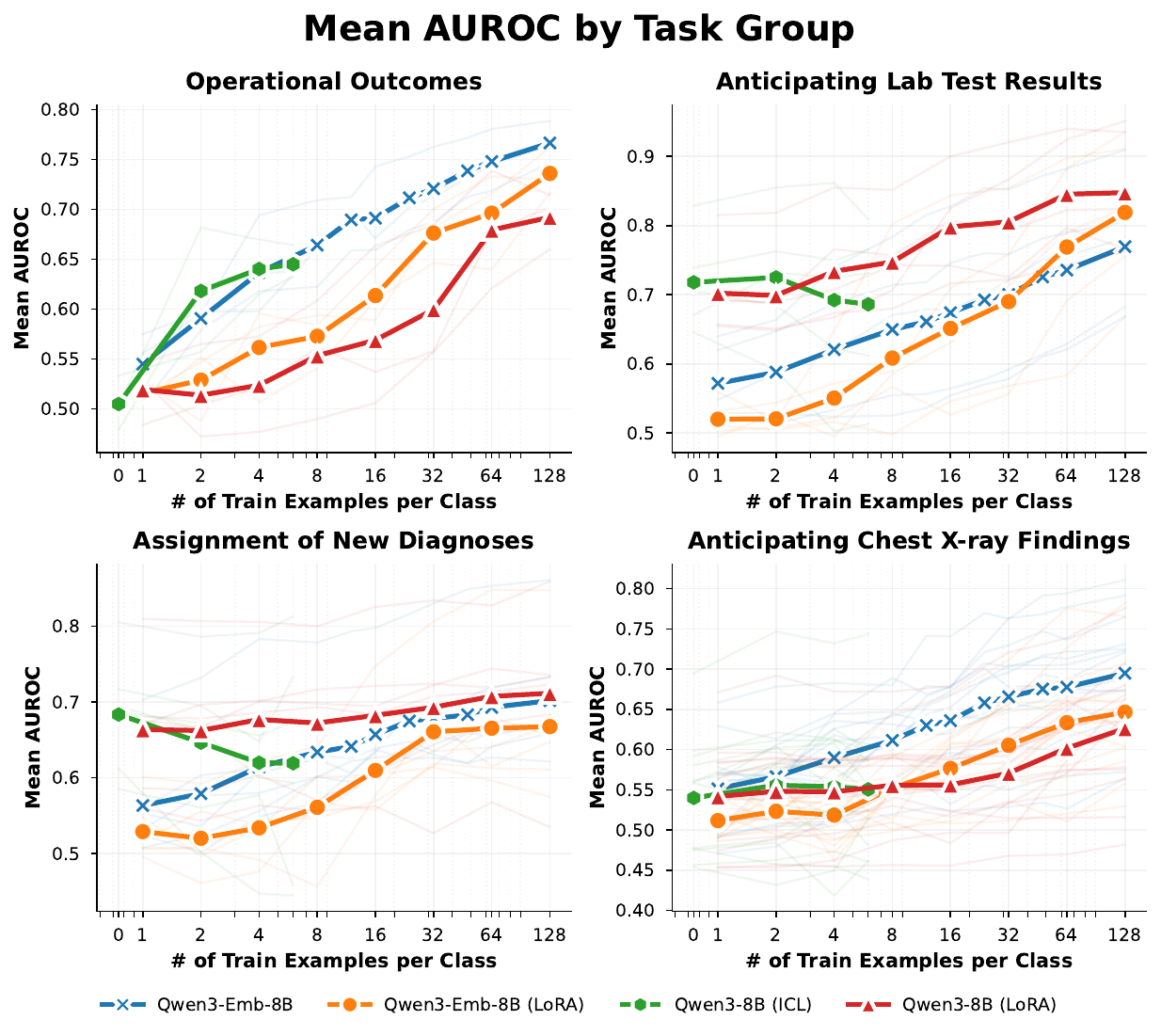}
    \caption{\textbf{Few-Shot AUROC Performance of Encoder and Decoder Models on EHRSHOT by Task Group.} Mean \acf{auroc} across subtasks for the four EHRSHOT task groups. Blurred lines show averages across five bootstrapped runs using different seeds. The decoder \ac{icl} curves are shown for \np{0}, \np{2}, \np{4}, and \np{6} shots only because larger \ac{icl} settings were not computationally feasible.}
    \label{fig:appendix_qwen3_encoder_decoder_by_group}
\end{figure}
\clearpage
\begin{figure}[htbp]
    \centering
    \includegraphics[width=\linewidth]{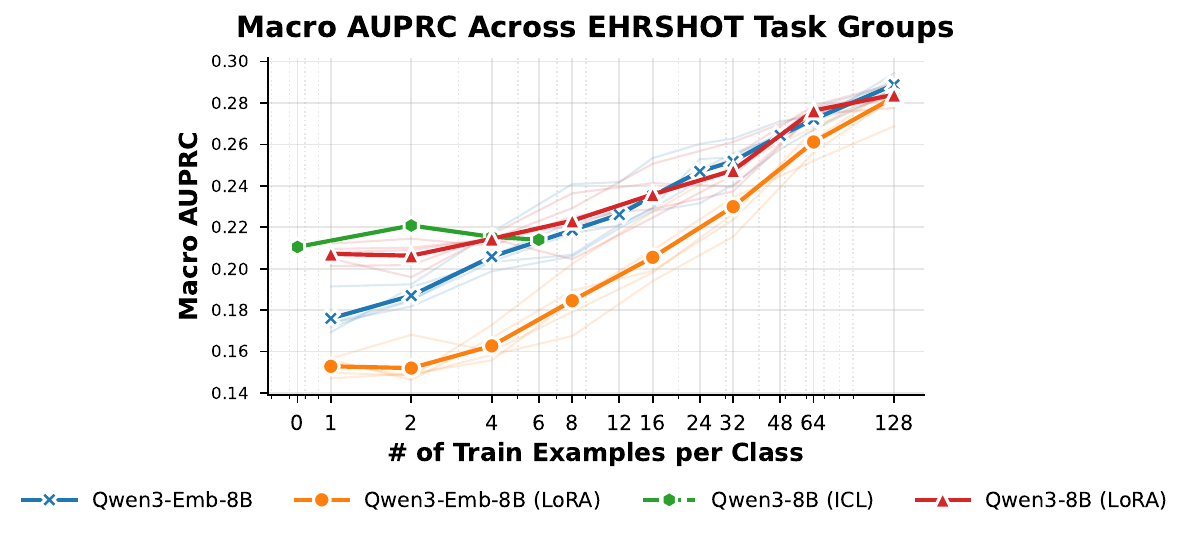}
    \caption{\textbf{Few-Shot AUPRC Performance of Encoder and Decoder Models on EHRSHOT.} Macro-averaged \acf{auprc} across all EHRSHOT subtasks for zero to \np{128} training examples per class, comparing the frozen encoder baseline, the \ac{lora}-tuned encoder, decoder \ac{icl} at \np{0}, \np{2}, \np{4}, and \np{6} shots, and the \ac{lora}-tuned decoder.}
    \label{fig:qwen3_encoder_decoder_overall_auprc}
\end{figure}
\clearpage
\begin{figure}[htbp]
    \centering
    \includegraphics[width=\linewidth]{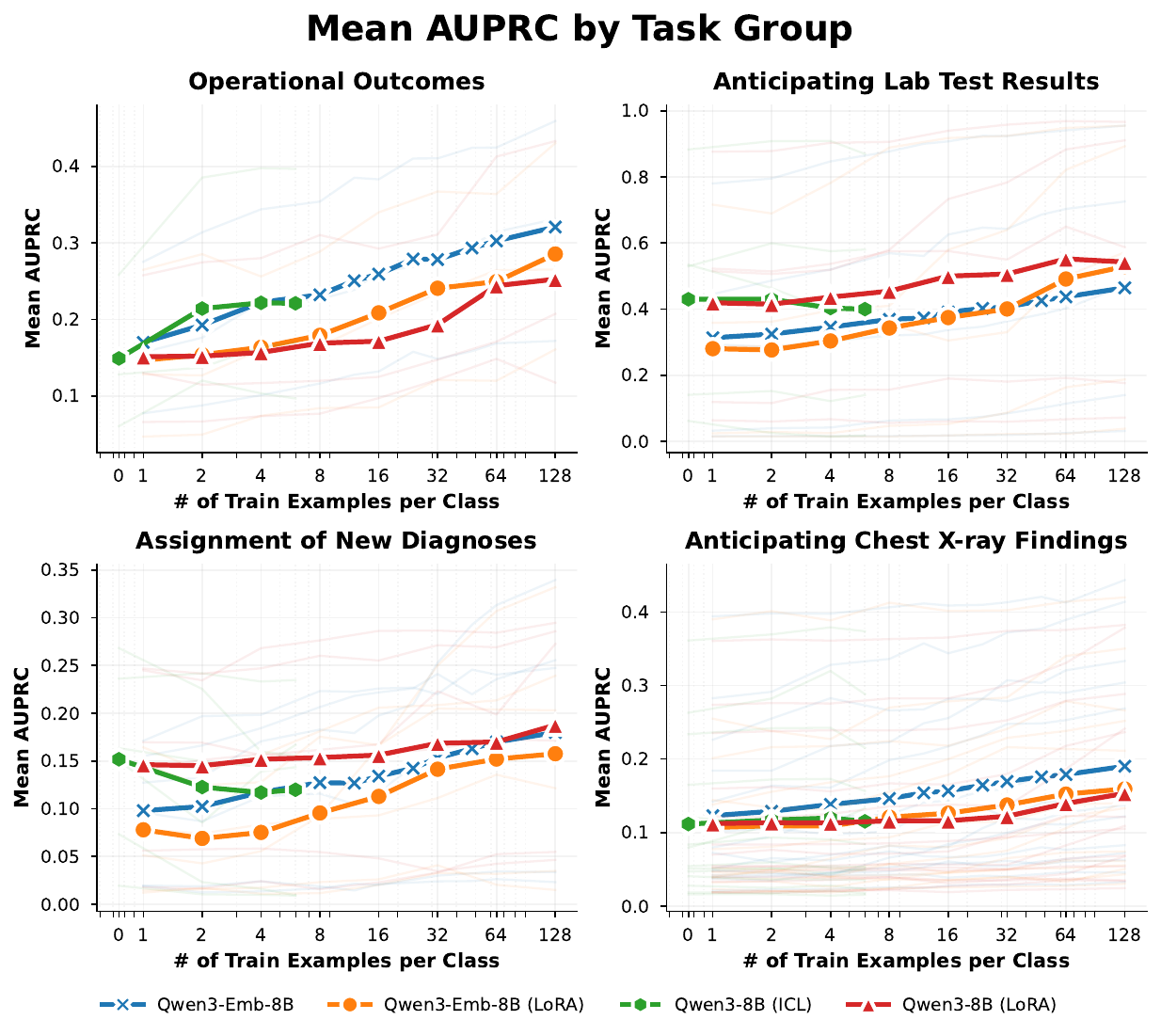}
    \caption{\textbf{Few-Shot AUPRC Performance of Encoder and Decoder Models on EHRSHOT by Task Group.} Mean \acf{auprc} across subtasks for the four EHRSHOT task groups. Blurred lines show averages across five bootstrapped runs using different seeds. Decoder \ac{icl} beyond \np{6} shots was not computationally feasible and is therefore not shown.}
    \label{fig:qwen3_encoder_decoder_by_group_auprc}
\end{figure}

\end{document}